\documentclass{article}
\usepackage{kr}

\usepackage{times}
\usepackage{soul}
\usepackage{url}
\usepackage[utf8]{inputenc}
\usepackage[small]{caption}
\usepackage{graphicx}
\usepackage{amsmath}
\usepackage{amsthm}
\usepackage{booktabs}
\usepackage{algorithmic}
\usepackage{algorithm2e} 

\usepackage{bbm}
\usepackage{misc}
\usepackage{pifont}
\usepackage{xspace}
\usepackage{boxedminipage}
\usepackage{amssymb}
\usepackage{mathtools}

\usepackage{thmtools,thm-restate}

\usepackage{stmaryrd} 

\usepackage{tikz}
\usetikzlibrary{decorations.pathreplacing}
\usetikzlibrary{matrix}
\usetikzlibrary{arrows}
\usetikzlibrary{positioning}

\newtheorem{lemma}{Lemma} 
\newtheorem{prop}{Proposition} 
\newtheorem{theorem}{Theorem} 
\newtheorem{conjecture}{Conjecture} 
\newtheorem{exmp}{Example}
\newtheorem{defn}{Definition}

\usepackage{latexsym}

\title{How to Approximate Ontology-Mediated Queries}

\author{Anneke Haga$^1$ \and Carsten Lutz$^1$ \and Leif Sabellek$^1$ \and Frank Wolter$^2$\\
\affiliations
$^1$Department of Computer Science, University of Bremen, Germany\\
$^2$Department of Computer Science, University of Liverpool, UK\\
\emails
\{anneke,clu,sabellek\}@uni-bremen.de,
wolter@liverpool.ac.uk
}

\begin{document}

\date{}

\maketitle

\begin{abstract}
  We introduce and study several notions of approximation for
  ontology-mediated queries based on the description logics
  \ALC and \ALCI. Our approximations are of two kinds: we may
  (1)~replace the ontology with one formulated in a
  tractable ontology language such as $\ELI$ or certain TGDs and
  (2)~replace the database with one from a tractable class such as
the class of databases whose treewidth is bounded by a constant.  We
  determine the computational complexity and the relative
  completeness of the resulting approximations. (Almost) all of them reduce the
  data complexity from \coNP-complete to \PTime, in some cases even to
  fixed-parameter tractable and to linear time. While approximations
  of kind~(1) also reduce the combined complexity, this tends to not
  be the case for approximations of kind~(2). In some cases, the
  combined complexity even increases.
\end{abstract}

\section{Introduction}

Ontology-mediated querying enriches database queries by an ontology,
in this way providing domain knowledge and extending the language
available for formulating queries.  For ontologies written in popular
expressive description logics (DLs) such as \ALC and \ALCI, however,
the complexity of ontology-mediated querying is prohibitively high,
{\sc coNP}-complete in data complexity \cite{Schaerf-93} and \mbox{\ExpTime-}
resp.\ \TwoExpTime-complete in combined complexity
\cite{DBLP:conf/cade/Lutz08}. As a consequence, practical
implementations often resort to approximating the answers to ontology
mediated queries (OMQs)
\cite{DBLP:conf/rr/TserendorjRKH08,DBLP:conf/esws/ThomasPR10,DBLP:journals/jair/ZhouGNKH15},
mostly using rather pragmatic approaches.  The aim of this paper is to
carry out a systematic study of OMQ approximation from a theoretical
angle, introducing several principled notions of approximation and
clarifying their computational complexity and relative completeness.
In particular, we aim to find approximations that reduce the data
complexity to \PTime or even to fixed-parameter tractability (FPT).
Preferably, they should additionally reduce the combined complexity.

We mainly consider approximation from below, that is, approximations
that are sound, but incomplete.  While we also present some first
results on approximation from above, whenever we speak of
approximation without further qualification we mean approximation from
below. An OMQ is a triple $Q=(\Omc,\Sigma,q)$ where \Omc is an
ontology, $q$ an actual query such as a conjunctive query (CQ), and
$\Sigma$ a signature for the databases \Dmc that $Q$ is evaluated
on. Our starting point is the observation that this gives us three
points of attack for approximation: we can relax the ontology \Omc,
the query $q$, and the database \Dmc. However, relaxing the query is
not useful for attaining \PTime data complexity as in the DLs
mentioned above, ontology-mediated
querying is {\sc coNP}-hard 
already for atomic
queries (AQs), that is, for very simple CQs of the form $A(x)$. We are
thus left with the ontology and the database.

For ontology relaxing approximation, we choose an ontology language \Lmc that admits
ontology-mediated querying in \PTime in data complexity. To define
approximate answers to OMQ $Q=(\Omc,\Sigma,q)$, we then consider all
\Lmc-ontologies $\Omc'$ with \mbox{$\Omc \models \Omc'$} (to
guarantees soundness), replace \Omc with $\Omc'$, and take the union
of the answers to the resulting OMQs $Q'$. Equivalently, we can use a
single $\Omc'$, namely the unique \emph{logically strongest}
\Lmc-ontology with $\Omc \models \Omc'$. Such $\Omc'$ will typically
be infinite \cite{ourijcai19,ourijcai20}, but it turns out that it
never has to be materialized by an algorithm that computes approximate
answers; the ontologies $\Omc'$ in fact only serve the purpose of defining the
semantics of approximation. As choices for \Lmc, we consider Horn
description logics such as \ELI and sets of restricted
tuple-generating dependencies (TGDs), also known as existential rules
and Datalog$^\pm$ \cite{DBLP:conf/rweb/GottlobMP15}.  A related (but
stronger) notion of OMQ approximation was proposed in
\cite{ourijcai20}.

For database relaxing approximation, we choose a class of databases
\Dmf that admits ontology-mediated querying in \PTime in data
complexity. To define approximate answers to OMQ $Q$ on input database
\Dmc, we then consider all databases $\Dmc' \in \Dmf$ such that there
is a homomorphism from $\Dmc'$ to \Dmc (to guarantee soundness) and
take the union of the answers to $Q$ on all such $\Dmc'$. As choices
for \Dmf, we consider databases of bounded treewidth 
and databases that are proper trees. 
Equivalently, we can use the unraveling of \Dmc into a structure of
bounded treewidth, resp.\ into a tree. Such an unraveling may be
infinite, but again this is unproblematic as the unraveling never has
to be materialized by an algorithm that computes approximate answers.
%

Both of these approaches to approximation can also be used to define
approximation from above.  For ontology strengthening approximation,
one requires $\Omc' \models \Omc$ and for database strengthening
approximation, one requires that there is a homomorphism from \Dmc to
$\Dmc'$, rather than the other way around. In both cases, one then
takes the intersection of the answers rather than the union. The
resulting approximations are complete, but unsound. Note that
approximation from above is particularly useful in combination with
approximation from below
\cite{DBLP:conf/rr/TserendorjRKH08,DBLP:journals/jair/ZhouGNKH15}. If
both approximations produce the same answers, one has actually
succeeded to compute the `real', non-approximate answers.

An OMQ language is a pair $(\Lmc,\Qmc)$ with \Lmc an ontology language
and \Qmc a query language. We consider the approximation of OMQ
languages $(\Lmc,\Qmc)$ where $\Lmc \in \{\ALC,\ALCI\}$ and
$\Qmc \in \{ \text{UCQ}, \text{CQ}, \text{AQ}, \text{bELIQ} \}$ with
UCQ denoting unions of CQs and bELIQ denoting the class of unary CQs
that correspond to \ELI-concepts (ELIQs) and of Boolean CQs
$\exists x \, q(x)$ with $q(x)$ an ELIQ. The exact problem that we
consider is (approximate) \emph{OMQ evaluation}, meaning to decide, given an OMQ
$Q$, a database \Dmc, and a tuple $\bar a$ of constants from $\Dmc$,
whether $\bar a$ is an (approximate) answer to $Q$ on \Dmc. We 
give an overview of our results. 

We start in Section~\ref{sect:eliu} with ontology relaxing
approximation, choosing for \Lmc the description logic $\ELI^u_\bot$,
that is, the extension of \ELI with the universal role and bottom
(logical falsity). We then prove that $\ELI^u_\bot$-ontology relaxing
OMQ evaluation is in \PTime in data complexity and \ExpTime-complete
in combined complexity for all OMQ languages $(\Lmc,\Qmc)$ mentioned
above. In contrast, non-approximate OMQ evaluation is
\TwoExpTime-complete in $(\ALCI,\Qmc)$ for $\Qmc \in \{ \text{CQ},
\text{UCQ} \}$. If we consider more restricted classes of queries, the
complexity improves further.  In fact, $\ELI^u_\bot$-ontology relaxing
OMQ evaluation is fixed-parameter tractable (FPT) with single
exponential overall running time if \Qmc is the class of all CQs or
UCQs whose treewidth is bounded by a constant; here and in what
follows, the parameter is the size of the OMQ. For $\Qmc =
\text{bELIQ}$, we even obtain linear time in data complexity.

In Section~\ref{sect:treedb}, we consider tree-database relaxing
approximation. These (almost) deliver the same answers as
$\ELI^u_\bot$-ontology relaxing approximation in
$(\ALCI,\text{bELIQs})$, but are incomparable for more expressive
query languages. They turn out to be less well-behaved regarding
combined complexity, being \TwoExpTime-complete in $(\ALCI,\Qmc)$
for $\Qmc \in \{ \text{UCQ}, \text{CQ} \}$; this in fact even holds for (U)CQs of bounded
treewidth. If $\Lmc=\ALC$ or $\Qmc \in \{ \text{AQ}, \text{bELIQ} \}$,
then they are only \ExpTime-complete, as in the non-approximate
case. On the other hand, tree-database relaxing approximation enjoys a
slight advantage in data complexity over $\ELI^u_\bot$-ontology
relaxing approximation, namely linear time for \emph{all} OMQ
languages $(\Qmc,\Lmc)$ considered in this paper. We also prove the
surprising result that tree-database relaxing OMQ evaluation is
\ExpSpace-hard in $(\ALC,\text{CQ})$ and \TwoExpTime-complete in
$(\ALC,\text{UCQ})$. This means that it is \emph{harder} in combined
complexity than non-approximate OMQ evaluation (which is only
\ExpTime-complete in these two cases), while it is easier in data
complexity.

In Section~\ref{sect:tgd}, we revisit ontology relaxing approximation,
replacing $\ELI^u_\bot$ with frontier-one TGDs whose rule bodies and
heads are of bounded treewidth. Recall that a TGD is
\emph{frontier-one} if body and head share at most a single
variable~\cite{DBLP:conf/ijcai/BagetLMS09}. For rule heads, we do not
only require bounded treewidth, but the existence of a tree
decomposition in which the bags overlap in at most a \emph{single}
element. The resulting approximations are significantly more complete
than $\ELI^u_\bot$-ontology relaxing approximations, but enjoy the
same favourable computational properties regarding both data and combined 
complexity except that we do not attain linear time. We also
observe that by increasing the treewidth of the rule bodies and heads,
we obtain infinite hierarchies of increasingly complete
approximations.

In Section~\ref{sect:btw}, we generalize tree-database relaxing
approximation into btw-database relaxing approximation, replacing tree
databases with databases of bounded treewidth. These are strictly more
complete than TGD-ontology relaxing approximations.  They enjoy the
same computational properties as tree-database relaxing approximations
both regarding data and combined complexity except that we do not
attain linear time. 

Finally, in Section~\ref{sect:fromabove} we consider approximation
from above. For database strengthening approximation, the results are
negative: we show {\sc coNP}-completeness in data complexity even if
the original OMQ is from $(\EL,\text{CQ})$, an
OMQ language that admits non-approximate OMQ evaluation in \PTime
in data complexity. Ontology strengthening approximation 
looks more promising.  We consider the fragment
$\mathcal{ELIU}_\bot$ of \ALC and show that $\ELI_\bot$-ontology
relaxing OMQ evaluation in $(\mathcal{ELIU}_\bot,\Qmc)$ is FPT with
double exponential overall running time (thus in \PTime in data
complexity) and \TwoExpTime-complete in combined complexity for $\Qmc
\in \{ \text{AQ}, \text{CQ}, \text{UCQ} \}$. Note that non-approximate
OMQ evaluation in $(\mathcal{ELIU}_\bot,\text{AQ})$ is only
\ExpTime-complete \cite{DBLP:conf/cade/Lutz08}, and thus this is
another case where approximate OMQ evaluation is harder in combined
complexity than non-approximate OMQ evaluation.
Full proofs are in the appendix.

\medskip {\bf Related work.} Several approaches achieve practically
efficient OMQ evaluation by a pragmatic translation of ontologies into
languages that enjoy \PTime data complexity such as Datalog, OWL 2 QL,
or OWL 2 EL. This includes Screech
\cite{10.1007/11574620_29,DBLP:conf/rr/TserendorjRKH08}, TrOWL
\cite{DBLP:conf/aaai/PanT07,DBLP:conf/esws/ThomasPR10}, and PAGOdA
\cite{DBLP:journals/jair/ZhouGNKH15}, see also
\cite{DBLP:conf/cade/MartinezFGHH14}. Approximations are computed both
from below and above, in the spirit of knowledge compilation
\cite{DBLP:journals/jacm/SelmanK96}.  Approximations of ontologies in
tractable languages with stronger guarantees were recently studied in
\cite{ourijcai19,ourijcai20}, but querying and data were (mostly) not
considered. In database theory, approximate querying (without
ontologies) was studied in
\cite{DBLP:conf/icdt/FinkO11,DBLP:journals/siamcomp/BarceloL014,DBLP:conf/icdt/Barcelo0Z18}. The
approximation and rewriting of OMQs in(to) datalog is studied in
\cite{DBLP:journals/tods/BienvenuCLW14,DBLP:journals/ai/KaminskiNG16,DBLP:journals/lmcs/FeierKL19}. In
the context of Horn DLs, OMQ approximation that achieves FPT was
considered in \cite{DBLP:conf/lics/BarceloFLP19}.









\section{Preliminaries}
\label{sect:prelim}

{\bf Description Logics.}
Let \NC and \NR be countably infinite sets of \emph{concept names} and
\emph{role names}. Further fix a countably infinite supply of
\emph{constants}. A \emph{role} is a role name $r$ or an \emph{inverse
  role} $r^-$, with $r$ a role name and $(r^-)^- = r$.  An
\emph{\ALCI-concept} is defined according to the syntax rule
$$C, D ::= \top \mid \bot \mid A \mid \neg C \mid C \sqcap D \mid \exists r.C $$
where $A$ ranges over concept names and $r$ over roles. We use $C
\sqcup D$ as abbreviation for $\neg (\neg C \sqcap \neg D)$, \mbox{$C
  \rightarrow D$} for $\neg C \sqcup D$, and $\forall r.C$ for $\neg
\exists r.\neg C$. An \emph{$\ELI_\bot$-concept} is an \ALCI-concept
that does not use negation ``$\neg$'' and an
\emph{$\ELI^u_\bot$}-concept is an $\ELI_\bot$-concept that may use
the \emph{universal role} $u$ in place of a role name. \ELI-concepts
and $\ELI^u$-concepts do not admit $\bot$.  Let $\Lmc \in \{ \ALCI,
\ELI, \ELI^u, \ELI_\bot, \ELI^u_\bot \}$.  An \emph{\Lmc-ontology} is
a finite set of \emph{concept inclusion (CIs)} $C \sqsubseteq D$ with
$C$ and~$D$ \Lmc-concepts.
We sometimes also consider infinite ontologies that, however, only
serve the purpose of defining a semantics and never have to be
represented explicitly. For $\ELI^u_\bot$-ontologies, we assume
w.l.o.g.\ that $\bot$ occurs only in CIs of the form
$C \sqsubseteq \bot$ where $\bot$ does not occur in $C$. A
\emph{database} is a finite set of \emph{facts} of the form $A(a)$ or
$r(a,b)$ where $A \in \NC \cup \{ \top \}$, $r \in \NR$, and $a,b$ are constants. We
use $\mn{adom}(\Dmc)$ to denote the set of constants used in database
\Dmc, also called its \emph{active domain}.

A \emph{signature} $\Sigma$ is a set of concept and role names,
uniformly referred to as \emph{symbols}. We use $\mn{sig}(X)$ to
denote the set of symbols used in any syntactic object $X$ such as a
concept or an ontology.  A \emph{$\Sigma$-database} is a database \Dmc
with $\Sig(\Dmc) \subseteq \Sigma$.  The \emph{size} of a (finite) syntactic
object $X$, denoted $||X||$, is the number of symbols needed to write
it as a word using a suitable encoding. 

The semantics is given in terms of \emph{interpretations} \Imc, which
we define to be a (possibly infinite and) non-empty set of facts. We
use $\Delta^\Imc$ to denote the set of individual names in \Imc,
define $A^\Imc = \{ a \mid A(a) \in \Imc \}$ for all $A \in \NC$, and
$r^\Imc = \{ (a,b) \mid r(a,b) \in \Imc \}$ for all $r \in \NR$.  The
extension $C^\Imc$ of \ALCI-concepts $C$ is then defined as usual
\cite{DBLP:books/daglib/0041477}. The universal role $u$ is
always interpreted as $u^\Imc = \Delta^\Imc \times \Delta^\Imc$.
This
definition of interpretation is slightly different from the usual one,
but equivalent; 
its virtue is uniformity as every database is a (finite)
interpretation.
Interpretation \Imc \emph{satisfies} CI
$C \sqsubseteq D$ if $C^\Imc \subseteq D^\Imc$, fact $A(a)$ if
$a \in A^\Imc$, and fact $r(a,b)$ if $(a,b) \in r^\Imc$. We thus make
the standard names assumption, that is, we interpret constants as
themselves. For $S \subseteq \Delta^\Imc$, we use $\Imc|_S$ to denote
the restriction of \Imc to facts that only contain constants from $S$.

Interpretation \Imc is a \emph{model} of an ontology or database if it
satisfies all inclusions or facts in it. A database $\Dmc$ is
\emph{satisfiable} w.r.t.\ an ontology \Omc if there is a model \Imc
of \Omc and~\Dmc. We write $\Omc \models \Omc'$ if every model of
ontology \Omc is also a model of ontology $\Omc'$.
We associate every interpretation \Imc with an undirected graph
$G_\Imc=(V,E)$ where $V=\Delta^\Imc$ and
$E=\{ \{d,e\} \mid (d,e) \in r^\Imc \text{ for some } r \in \NR \}$.
We say that \Imc is a \emph{tree} if $G_\Imc$ is acyclic without self
loops and multi-edges, that is, $(d,e) \in r_1^\Imc$ implies
$(d,e) \notin r_2^\Imc$ for all distinct roles $r_1,r_2$. Note that,
somewhat unusually, our trees need thus not be connected.

\smallskip

{\bf Queries.}
A \emph{conjunctive query (CQ)} is of the form
$q(\bar x) = \exists \bar y\,\varphi(\bar x,\bar y)$, where $\bar x$
and $\bar y$ are tuples of variables and $\varphi(\bar x,\bar y)$ is a
conjunction of \emph{atoms} of the form $A(x)$ and $r(x,y)$,
\mbox{$A \in \NC$}, $r \in \NR$, and $x,y$ variables from
$\bar x \cup \bar y$.  We require that all variables in $\bar x$ are
used in $\varphi$, call the variables in $\bar x$ the \emph{answer variables} of~$q$,
and use $\mn{var}(q)$ to denote $\bar x \cup \bar y$.  We take the
liberty to write $\alpha \in q$ to indicate that $\alpha$ is an atom
in $q$ and sometimes write $r^-(x,y) \in q$ in place of
\mbox{$r(y,x) \in q$}. The CQ $q$ gives rise to a database $\Dmc_q$,
often called the canonical database for $q$,
obtained by viewing the variables in $q$ as constants and the
atoms as facts.  For
$V \subseteq \mn{var}(q)$, we use $q|_V$ to denote the restriction of
$q$ to the atoms that use only  variables in~$V$.

A \emph{homomorphism} from interpretation $\Imc_1$ to interpretation 
$\Imc_2$ is a function $h :\Delta^{\Imc_1} \to \Delta^{\Imc_2}$ such 
that $d \in A^{\Imc_1}$ implies $h(d) \in A^{\Imc_2}$ and 
$(d,e) \in r^{\Imc_1}$ implies $(h(d),h(e)) \in r^{\Imc_2}$ for all 
$d,e \in \Delta^{\Imc_1}$, $A \in \NC$, and $r \in \NR$. 
A homomorphism from CQ $q$ to
interpretation \Imc is a homomorphism from $\Dmc_q$ to \Imc. A tuple
$\bar d \in (\Delta^\Imc)^{|\bar x|}$
is an \emph{answer} to $q$ on
\Imc 
if there is a homomorphism $h$
from $q$ to \Imc with $h(\bar x)=\bar d$. 
%
A \emph{contraction} of a CQ $p$ is a CQ that can be obtained from $q$
by identifying variables. The identification of two answer variables 
is not admitted and the identification of an answer variable $x$ with a 
quantified variable $y$ results in $x$. 

A {\em union of conjunctive queries (UCQ)} $q(\bar x)$ is a disjunction
of CQs that all have the same answer variables $\bar x$. A tuple
$\bar d \in (\Delta^\Imc)^{|\bar x|}$ is an \emph{answer} to $q$ on
interpretation \Imc, written $\Imc \models q(\bar d)$,
if $\bar d$ is an answer to
some CQ in $q$ on \Imc. 
We use $q(\Imc)$ to denote set of all
answers to $q$ on \Imc.  The \emph{arity} of $q$ is the length of
$\bar x$ and $q$ is \emph{Boolean} if it is of arity zero.

An \emph{\ELI-query (ELIQ)} is a unary CQ $q(x)$ such that $\Dmc_q$ is
a connected tree and a \emph{Boolean \ELI-query (BELIQ)} is a Boolean
CQ $q()$ such that $\Dmc_q$ is a connected tree. One can alternatively
define ELIQs as being of the form $C(x)$ with $C$ an \ELI-concept, and
BELIQs as being of the form $\exists u . C$ with $C$ an
$\ELI$-concept and we may thus use ELIQs as \ELI-concepts and BELIQs as
$\ELI^u$-concepts, and vice versa.  For uniformity, we use
\emph{bELIQ} to refer to a CQ that is either an ELIQ or a BELIQ.  An
\emph{atomic query (AQ)} is an ELIQ of the form $A(x)$, $A$ a concept
name.

\smallskip
{\bf Ontology-Mediated Queries.}
An \emph{ontology-mediated query (OMQ)} is a triple
$Q=(\Omc,\Sigma,q)$ with \Omc an ontology,
$\Sigma \subseteq \mn{sig}(\Omc) \cup \mn{sig}(q)$ a signature called
the \emph{data signature}, and $q$ a query such as a UCQ. We write $Q(\bar
x)$ to indicate that the answer variables of $q$ are $\bar x$.
%
The signature $\Sigma$ expresses the promise that $Q$ is only
evaluated on $\Sigma$-databases.
Let \Dmc be such a database. Then $\bar a \in
\mn{adom}(\Dmc)^{|\bar x|}$ is an \emph{answer} to $Q$ on \Dmc,
written $\Dmc \models Q(\bar a)$, if $\Imc \models q(\bar a)$ for all
models \Imc of \Omc and~\Dmc.  When more convenient, we might
alternatively write $\Dmc,\Omc \models q(\bar a)$. We further write $Q(\Dmc)$
to denote the set of all answers to $Q$ on \Dmc. For OMQs $Q_1(\bar
x)$ and $Q_2(\bar x)$, $Q_i =(\Omc_i,\Sigma,q_i)$, we say that $Q_1$
is \emph{contained} in $Q_2$ and write $Q_1 \subseteq Q_2$, if for
every $\Sigma$-database \Dmc, 
$Q_1(\Dmc) \subseteq Q_2(\Dmc)$. We say that $Q_1$ is \emph{equivalent} to $Q_2$
and write $Q_1 \equiv Q_2$, if $Q_1 \subseteq Q_2$ and $Q_2 \subseteq
Q_1$.  We use $(\Lmc,\Qmc)$ to denote the \emph{OMQ language} that
contains all OMQs~$Q$ in which $\Omc$ is formulated in DL \Lmc and $q$
in query language \Qmc, such as in $(\ALCI,\text{UCQ})$ and $(\ELI,\text{AQ})$.

\smallskip {\bf Treewidth.}  Treewidth is a widely used notion that
measures the degree of tree-like\-ness of a graph.
As common for
example in the area of constraint satisfaction problems, we are
interested in two parameters of tree decompositions instead of only
one.  A \emph{tree decomposition} of an interpretation $\Imc$ is a
triple $(V,E, (B_v)_{v \in V})$ where $(V, E)$ is an undirected tree
and $(B_v)_{v \in V} $ is a family of subsets of $\Delta^\Imc$, often
referred to as bags, such that:
\begin{enumerate} 
\item for all $d \in \Delta^\Imc$, $\{v \in V\mid d \in B_v\}$ is
  nonempty and connected in~$(V,E)$;
\item if $(d_1,d_2)\in r^\Imc$ for any role name $r$, then there is a $v \in V$ with $d_1,d_2 \in B_v$.
\end{enumerate}
We call $(V,E, (B_v)_{v \in V})$ an \emph{$(\ell,k)$-tree 
decomposition} if for all distinct $v,v' \in V$, 
%
$|B_v \cap B_{v'}| \leq \ell$ and 
$|B_v| \leq k$. 
An interpretation \Imc \emph{has treewidth} $(\ell,k)$ if it admits an $(\ell,k)$-tree decomposition. 
It \emph{has treewidth} $k$ if it has treewidth 
$(k,k+1)$.
As usual, the `+1' is used to achieve that trees have treewidth~1.

We also speak of the treewidth of a CQ
$q = \exists \bar y\,\varphi(\bar x,\bar y)$, which is that of
$\Dmc_q|_{\bar y}$, and of the treewidth of a UCQ $q$, which is the
maximum of the treewidths of the CQs in $q$.
Note that answer variables do not contribute to treewidth.
%
For $\ell,k \geq 1$ with
$\ell < k$, we use $\text{CQ}^{\text{tw}}_{\ell,k}$ (resp.\
$\text{CQ}^{\text{tw}}_{k}$)
to denote the
class of CQs of treewidth $(\ell,k)$ (resp.\ of treewidth $k$), and likewise for
$\text{UCQ}^{\text{tw}}_{\ell,k}$ (resp.\
$\text{UCQ}^{\text{tw}}_{k}$) and UCQs.

	
	

\smallskip
{\bf Tuple-Generating Dependencies.} A {\em
  tuple-generating dependency (TGD)} is a first-order
sentence $\vartheta$ of the form
$ \forall \bar x \forall \bar y \, \big(\phi(\bar x,\bar y)
\rightarrow \exists \bar z \, \psi(\bar x,\bar z)\big) $ such that
$\exists \bar y \, \phi(\bar x,\bar y)$ and
$\exists \bar z \, \psi(\bar x,\bar z)$ are CQs. 
For simplicity, we write $\vartheta$ as
$\phi(\bar x,\bar y) \rightarrow \exists \bar z \, \psi(\bar x,\bar
z)$. We call $\phi$ and $\psi$ the {\em body} and {\em head} of
$\vartheta$. 
The body may be the empty conjunction, i.e.~logical truth, then
denoted by $\top$, and the head may be logical falsity denoted by
$\bot$. TGDs with head $\bot$ are often called \emph{denial
  constraints} \cite{DBLP:series/synthesis/2012Fan}. The
\emph{frontier variables} of TGD $\vartheta$ are the variables that
occur unquantified in both body and head. We say that $\vartheta$ is
\emph{frontier one} if it has at most one frontier variable.
An interpretation $\Imc$
\emph{satisfies} $\vartheta$, denoted $\Imc \models \vartheta$, if
$q_\phi(\Imc) \subseteq q_\psi(\Imc)$.

We also consider ontologies that are sets of TGDs and, more generally,
sentences formulated in first-order logic (FO).  What we mean here is
the version of FO in which only unary and binary relation symbols are
used, which are from \NC and \NR, respectively. Function symbols, 
constants, and equality are not permitted.
An interpretation is
a \emph{model} of an FO ontology if it satisfies all sentences in
\Omc.
With an
\emph{FO-fragment} \Lmc, we mean a class of FO-sentences and an
\emph{\Lmc-ontology} is a finite set of sentences from \Lmc. As in
the DL case, we sometimes also consider infinite ontologies. We use
TGD to denote the FO-fragment that consists of all TGDs, and thus
speak of \emph{TGD-ontologies}.
It is easy to see that 
every $\ELI^u_\bot$-ontology is also a frontier one
TGD-ontology. 

A standard tool for dealing with TGD ontologies \Omc is the
\emph{chase} that constructs from \Omc and a database \Dmc a universal
model of \Omc and \Dmc, that is, a model $\mn{ch}_\Omc(\Dmc)$ of \Dmc
and \Omc that homomorphically embeds into every model of \Dmc and \Omc
and thus satisfies $\Dmc,\Omc \models q(\bar a)$ iff
$\mn{ch}_\Omc(\Dmc) \models q(\bar a)$ for all CQs $q$ and tuples
$\bar a$. Details are given in the appendix, see also
\cite{DBLP:conf/pods/JohnsonK82,DBLP:journals/jair/CaliGK13}.

\section{OMQ Approximation}
\label{sect:approxdef}

We introduce two notions of OMQ approximation from below: one where we
relax the ontology and one where we relax the database. We start with
the former. 

For an OMQ $Q(\bar x)=(\Omc,\Sigma,q)$, a $\Sigma$-database \Dmc, and
an ontology language $\Lmc'$, we use $\mn{app}_{\Lmc'}(Q,\Dmc)$ to
denote the set of tuples $\bar a \in \mn{adom}(\Dmc)^{|\bar x|}$ such
that $\bar a \in Q'(\Dmc)$ for some OMQ $Q'=(\Omc',\Sigma,q)$ where
$\Omc'$ is a (finite) $\Lmc'$-ontology with $\Omc \models \Omc'$. Note
that the ontology $\Omc'$ might contain symbols that do not occur
in~\Omc, we will see later that this in fact results in additional
answers.
%
%
%
Every choice of an OMQ language $(\Lmc, \Qmc)$ and an ontology
language $\Lmc'$ gives rise to an approximate
OMQ evaluation problem, as follows.
\begin{center}
  \fbox{\begin{tabular}{l@{\;}l}
          \multicolumn{2}{l}{\emph{$\Lmc'$-ontology relaxing OMQ evaluation in $(\Lmc,\Qmc)$}}
			\\[.5mm]{\small INPUT}: & OMQ $Q(\bar x) =
                                              (\Omc,\Sigma,q)
                                             \in (\Lmc,\Qmc)$, \\
&			$\Sigma$-database $\Dmc$,  tuple $\bar a \in \mn{adom}(\Dmc)^{|\bar x|}$
			\\[.5mm]
                {\small OUTPUT}: &  `yes' if
                                    $\bar a \in \mn{app}_{\Lmc'}(Q,\Dmc)$
                                            and `no' otherwise
	\end{tabular}}
\end{center}
 It follows from the definition that ontology relaxing approximation
 is sound, in the sense that $\mn{app}_{\Lmc'}(Q,\Dmc)\subseteq
 Q(\Dmc)$
 for all OMQs $Q=(\Omc,\Sigma,q)$ and $\Sigma$-databases \Dmc.
%
 We concentrate on the case where $\Lmc'$ is an ontology language such that
 $(\Lmc',\Qmc)$ enjoys \PTime data complexity. For
 $(\Lmc,\Qmc)=(\ALCI,\text{CQ})$, for instance, we might choose
 $\Lmc'=\ELI^u_\bot$.

 In the definition of
 $ \mn{app}_{\Lmc'}(Q,\Dmc)$, one can equivalently replace the
 infinitely many $\Lmc'$-ontologies $\Omc'$ with the
 single but infinite $\Lmc'$-ontology $\Omc^\approx_{\Lmc'}$ that
 consists of all $\Lmc'$-sentences $\vp$ with $\Omc \models \vp$.
 In fact, the following lemma is a consequence of compactness.
%
%
%
 For an OMQ $Q(\bar x)=(\Omc,\Sigma,q)$ and an ontology language $\Lmc'$,
we use $Q^\approx_{\Lmc'}$ to denote the OMQ $(\Omc^\approx_{\Lmc'},\Sigma,q)$.
%
%
\begin{lemma}
  \label{lem:infont}
  Let $Q(\bar x)=(\Omc,\Sigma,q) \in (\text{FO},\text{UCQ})$ be an OMQ and $\Lmc'$ an ontology
  language. Then for every $\Sigma$-database \Dmc, $
  \mn{app}_{\Lmc'}(Q,\Dmc) = Q^\approx_{\Lmc'}(\Dmc)$.
\end{lemma}
Note that we do not insist that the infinitely many $\Omc'$ or the
infinite $\Omc^\approx_{\Lmc'}$ is ever explicitly generated when
computing approximate answers.

 We next define a mode of approximation that is based on relaxing the
 database by replacing it with a homomorphic pre-image.  A
 \emph{pointed database} is a pair $(\Dmc,\bar b)$ with \Dmc a
 database and $\bar b$ a tuple over $\mn{adom}(\Dmc)$.  
  For an OMQ $Q(\bar x)=(\Omc,\Sigma,q)$, 
a $\Sigma$-database \Dmc, and a class of pointed databases \Dmf, we use $\mn{app}_{\Dmf}(Q,\Dmc)$ to
  denote the set of tuples $\bar a \in \mn{adom}(\Dmc)^{|\bar x|}$
  such that for some $(\Dmc',\bar b) \in \Dmf$ with $\bar b \in Q(\Dmc')$
there is a homomorphism $h$ from $\Dmc'$ to \Dmc with
  $h(\bar b)=\bar a$.  Every choice of $(\Lmc, \Qmc)$ and $\Dmf$
gives rise to an approximate OMQ evaluation problem, as follows.
\\[-6mm]
\begin{center}
	\fbox{\begin{tabular}{l@{\;}l}
			\multicolumn{2}{l}{\emph{$\Dmf$-database relaxing OMQ evaluation in $(\Lmc,\Qmc)$}}
			\\[.5mm]{\small INPUT} : & OMQ $Q(\bar x) =
                                              (\Omc,\Sigma,q)
                                             \in (\Lmc,\Qmc)$, \\
&			$\Sigma$-database $\Dmc$,  tuple $\bar a \in \mn{adom}(\Dmc)^{|\bar x|}$
			\\[.5mm]
                {\small OUTPUT} : &  `yes' if
                                    $\bar a \in \mn{app}_{\Dmf}(Q,\Dmc)$
                                            and `no' otherwise
	\end{tabular}}
\end{center}
%
Answers to any OMQ $Q=(\Omc,\Sigma,q) \in
(\text{FO},\text{UCQ})$ are preserved under homomorphisms
if \Omc does not use equality,
 that is, if $\Dmc_1,\Dmc_2$ are
 databases, $h$ is a homomorphism from $\Dmc_1$ to $\Dmc_2$, and
 $\bar a \in Q(\Dmc_1)$ for an OMQ $Q$, then $h(\bar a) \in Q(\Dmc_2)$
\cite{DBLP:journals/tods/BienvenuCLW14}.
As a consequence, database relaxing approximation
is sound.
%

We are interested in choosing \Dmf such that evaluating OMQs from
$(\Lmc,\Qmc)$ on \Dmf enjoys \PTime data complexity. An important example
are classes of databases that are of bounded
treewidth, the simplest case being the class of databases that are trees.
More precisely, we use
$\Dmf_{\!\curlywedge}$ (with
`$\curlywedge$' symbolizing a tree) to denote the class of all pointed
databases $(\Dmc,\bar
a)$ such that the restriction of \Dmc to domain $\mn{adom}(\Dmc)
\setminus \bar
a$ is a tree.  Recall that a tree does not need to be connected.
The resulting notion of approximation is closely related to 
$\ELI^u_\bot$-ontology relaxing approximation. 

In the same way in which we have rephrased ontology relaxing
approximation in terms of a single infinite ontology, we can sometimes
(depending on the choice of \Dmf) rephrase database relaxing
approximation in terms of evaluation on a single infinite database.
We illustrate this for the case
\mbox{$\Dmf=\Dmf_{\!\curlywedge}$}. 
Let $\Dmc$ be a database and $S \subseteq \mn{adom}(\Dmc)$. A
\emph{path} in \Dmc is a sequence $p=a_0r_1a_1r_2 \cdots r_na_n$,
$n \geq 0$, where $a_0,\dots,a_n \in \mn{adom}(\Dmc)$,
$r_1,\dots,r_n$ are (potentially inverse) roles,
and $r_{i+1}(a_i,a_{i+1}) \in \Dmc$ for $0 \leq i < n$.  We use
$\mn{tail}(p)$ to denote $a_n$. The \emph{tree unraveling}
$\Dmc^\approx_S$ of \Dmc at $S$ is the (potentially infinite) database
that contains the following facts: all facts from $\Dmc|_S$,
$r(p,prb)$ for every path $prb$, $A(p)$ for every path $p$ with
$A(\mn{tail}(p)) \in \Dmc$, and $r(a,p)$ for every $r(a,b) \in \Dmc$
and every path $p$ with $a \in S$ and $\mn{tail}(p)=b$ ($r$ a
potentially inverse role). Note that $\Dmc^\approx_S$ is a tree if and
only if $S=\emptyset$.  Thus, the tree unravelings
$\Dmc^\approx_{\bar a}$ in the following lemma are in general not tree
databases. 
%
\begin{restatable}{lemma}{lemunravfund}
  \label{lem:unravfund}
  Let $Q =(\Omc,\Sigma,q) \in (\text{FO},\text{UCQ})$.  Then for all $\Sigma$-databases
  \Dmc and $\bar a \in \mn{adom}(\Dmc)^{|\bar x|}$,
  $\bar a \in \mn{app}_{\Dmf_{\!\curlywedge}}(Q,\Dmc)$ iff $\bar a \in 
  Q(\Dmc^\approx_{\bar a})$. 
\end{restatable}
%

%

\section{$\ELI^u_\bot$-Ontology Relaxing Approximation}
\label{sect:eliu}

We consider $\ELI^u_\bot$-ontology relaxing evaluation of OMQs from
$(\ALCI,\text{UCQ})$, starting with an example.
\begin{exmp}
  Let $Q(x)=(\Omc,\Sigma,q) \in (\ALC,\text{CQ})$ where
  $$
  \begin{array}{rcl}
  \Omc &=& \{ \top \sqsubseteq \forall r . (B_1 \rightarrow A) \sqcup 
  \forall r . (B_2 \rightarrow A) \} \\[.5mm]
    \Sigma &=& \{r,A,B_1,B_2\} \\[.5mm]
    q &=& \exists y \, r(x,y) 
    \wedge A(y).
      \end{array}
      $$
      Further  let 
  $
    \Dmc = \{ r(a,b_1), r(a,b_2),B_1(b_1),B_2(b_2) \}. 
  $
  Clearly, $a \in Q(\Dmc)$. The ontology $\Omc^\approx_{\ELI^u_\bot}$
  contains CI $\exists r . B_1 \sqcap \exists r . B_2 \sqsubseteq 
  \exists r . A$, thus also $a \in Q^\approx_{\ELI^u_\bot}(\Dmc)$. 
\end{exmp}
We next illustrate incompleteness, 
which cannot be avoided by any notion of approximation from below that attains 
\PTime data complexity. This follows from the existence of OMQs that express non-3-colorability. 
\begin{exmp}
\label{ex:3col}
  Let $Q()=(\Omc,\Sigma,\exists x \,D(x)) \in (\ALC,\text{BELIQ})$ with 
  $$\Omc = \{ \top  \sqsubseteq  R \sqcup G \sqcup B, \
      X \sqcap \exists e . X  \sqsubseteq  D \mid  X \in 
      \{R,G,B\} \}$$
  %
  and $\Sigma = \{ e \}$. Every $\Sigma$-database \Dmc can be
  viewed as an undirected graph by `forgetting' the direction of
  $e$-edges. 
  Then $\Dmc \models Q$ iff the graph is not 3-colorable. In
  contrast,the careful chase algorithm given below can be used to
  verify that $\Dmc \not\models Q^\approx_{\ELI^u_\bot}$ for all
  $\Sigma$-databases \Dmc. It is easy to modify the example so as
to use an AQ in place of a BELIQ.
\end{exmp}
For readers who are disappointed by the extreme incompleteness
in the previous example,
we remark replacing $\ELI^u_\bot$ with 
classes of TGDs improves the situation.

We next present three observations regarding our definition of
ontology relaxing approximation. The first observation is that it
increases completeness to admit in $\Omc^\approx_{\ELI^u_\bot}$ symbols
that do not occur in \Omc.
\begin{exmp}\label{exm:xx}
  Let $Q(x)=(\Omc,\Sigma,q) \in (\ALC,\text{CQ})$ with
  $$
  \begin{array}{r@{\;}c@{\;}l}
  \Omc &=& \{ \top
  \sqsubseteq \forall r . (B_1 \rightarrow B) \sqcup \forall r . (B_2
           \rightarrow B) \} \\[1mm]
    \Sigma &=& \{r,A,B,B_1,B_2\} \quad 
    q = \exists y \,
    r(x,y) \wedge A(y) \wedge B(y).
      \end{array}
      $$
      Then $\Omc^\approx_{\ELI^u_\bot}$ contains the CI
  $$
    \exists r . (B_1 \sqcap A) \sqcap \exists r . (B_2 \sqcap A)
      \sqsubseteq \exists r . (A \sqcap B)
  $$
  despite the fact that $A$ does not occur in \Omc. Let
  $$\Dmc = \{
  r(a,b_1), r(a,b_2), B_1(b_1), B_2(b_2),A(b_1),A(b_2) \}.
  $$
  Then $a \in Q^\approx_{\ELI^u_\bot}(\Dmc)$, but we show in the
  appendix that this is no longer true when we remove from
  $\Omc^\approx_{\ELI^u_\bot}$ all CIs that use a symbol that does not
  occur in \Omc.
%
\end{exmp}
It is, however, easy to see that it suffices to admit in
$\Omc^\approx_{\ELI^u_\bot}$ the symbols that occur in \Omc or in
$\Sigma$ while additional symbols do not further increase
completeness.
%

The second observation is that $\ELI^u_\bot$-ontology relaxing
approximation is more complete than $\ELI_{\bot}$-ontology relaxing
approximation. 
  In fact, it seems to be much more challenging to
compute answers for the latter while offering no obvious benefit
compared to the former, and thus we do not consider it in this paper.
\begin{exmp}\label{exm:yy}
  Let $Q()=(\Omc,\Sigma,q) \in (\ALC,\text{CQ})$ be the Boolean OMQ with
  $$
    \Omc = \{ A \sqsubseteq B \sqcup \forall r . B \} \quad
    \Sigma=
              \{r,A,B\} \quad
    q = \exists x \, B(x).
  $$
  Let $\Dmc= \{A(a),r(a,b)\}$.
  Then $\Omc^\approx_{\ELI^u_\bot}$ contains $A \sqcap \exists r.\top 
  \sqsubseteq \exists u . B$ and thus 
  $\Dmc \models Q^\approx_{\ELI^u_\bot}$, but it is shown in the appendix that $\Dmc \not\models Q^\approx_{\ELI_{\bot}}$.
\end{exmp}
We remark that $\ELI^u_\bot$-ontology relaxing approximation is
also more complete than $\ELI^u$-ontology relaxing approximation, examples
are easy to find.

Third, we observe that defining ontology
relaxing approximation in terms of ontologies that are implied by the
original ontology 
does not necessarily result in maximum completeness. In fact, the
following example shows that it may pay off to use an
$\ELI^u_\bot$-ontology that is \emph{not} a consequence of the
original ontology. This is a very interesting effect, but we do not
investigate it further.
\begin{exmp}
We use the ontology $\Omc$ and signature $\Sigma$ from Example~\ref{exm:yy}.
Let $Q_{0}()=(\Omc,\Sigma,q_{0})$ for    
  $$
   q_{0}=\exists x \exists y \, A(x) \wedge A(y) \wedge B(y) \wedge r(x,y) \wedge r(y,x),
  $$
  and let $\Dmc_{0} = \{A(a),r(a,b),r(b,a),A(b) \}$. While  $\Dmc_{0}\models Q_{0}$,
the careful chase
  algorithm below yields $\Dmc_{0}\not\models Q^{\approx}_{\ELI^u_\bot}$. Now take $\Omc' = \{ A \sqsubseteq B \}$ and note that $\Omc \not\models \Omc'$ but $Q_{0}$ is equivalent to $(\Omc',\Sigma,q_{0})$ as in fact both OMQs are equivalent to the CQ $q_{0}$ with atom $B(y)$ dropped.
\end{exmp}
%
The following is the main result of this section. When we speak about
fixed-parameter tractability (FPT), we generally mean that the
parameter is the size of the OMQ, that is, we refer to running time
$f(||Q||) \cdot O(||\Dmc||^c)$ where $f$ is a computable function
and $c$ a constant.
\begin{theorem}
  \label{thm:mainone}
  Let $\Lmc \in \{ \ALC, \ALCI \}$. Then   $\ELI^u_\bot$-ontology relaxing OMQ evaluation
  is
  \begin{enumerate}

  \item \ExpTime-complete in combined complexity and \PTime-complete
    in data complexity in $(\Lmc,\Qmc)$,
    $\Qmc \in \{ \text{AQ}, \text{CQ}, \text{UCQ} \}$;

  \item FPT with single exponential running time
    in $(\Lmc,\Qmc)$, $\Qmc \in \{ \text{CQ}^{\text{tw}}_k, \text{UCQ}^{\text{tw}}_k
    \mid k \geq 1 \}$;

  \item in linear time in data complexity in $(\Lmc,\text{bELIQ})$
    with running time $2^{O(||Q||)} \cdot O(||\Dmc||)$.

  \end{enumerate}
\end{theorem}
To prove Theorem~\ref{thm:mainone}, we first establish the following
crucial lemma that relates $\ELI^u_\bot$-ontology relaxing OMQ
evaluation to tree unravelings.
\begin{restatable}{lemma}{lemcrucial}  
  \label{lem:crucial}
  Let $Q(\bar x) = (\Omc,\Sigma,q) \in (\ALCI,\text{bELIQ})$, 
  \Dmc be a $\Sigma$-database, and $\bar a \in \mn{adom}(\Dmc)^{|\bar 
    x|}$. Then 
  \begin{enumerate}

  \item 
  $\bar a \in Q^\approx_{\ELI^u_\bot}(\Dmc)$ iff 
  $\bar a \in Q(\Dmc^\approx_{\emptyset})$; 

  \item \Dmc is satisfiable w.r.t.\ $\Omc^\approx_{\ELI^u_\bot}$
    iff $\Dmc^\approx_{\emptyset}$ is satisfiable w.r.t.\ \Omc. 

  \end{enumerate}
\end{restatable}
We first prove Point~3 of Theorem~\ref{thm:mainone}. We are thus given
an OMQ $Q(\bar x)=(\Omc,\Sigma,q) \in (\ALCI,\text{bELIQ})$, a
$\Sigma$-database \Dmc, and a tuple
$\bar a \in \mn{adom}(\Dmc)^{|\bar x|}$ and have to decide whether
$\bar a \in Q^\approx_{\ELI^u_\bot}(\Dmc)$. By Point~1 of
Lemma~\ref{lem:crucial}, it suffices to decide whether
$\bar a \in Q(\Dmc^\approx_{\emptyset})$. This is much more convenient
as we are back to the original ontology instead of having to deal
directly with $\Omc^\approx_{\ELI^u_\bot}$. In the appendix, we show
that deciding $\bar a \in Q(\Dmc^\approx_{\emptyset})$ can be reduced
in linear time (data complexity) to the unsatisfiability of
propositional Horn formulas, which is well-known to be in linear time
\cite{DBLP:journals/jlp/DowlingG84}.

Regarding the upper bounds in Points~1 and~2, we first observe that we
can concentrate on CQs rather than UCQs. This follows from the fact
that $\ELI^u_\bot$-ontologies have universal models, even if infinite.
\begin{lemma}
\label{lem:UCQtoCQ}
Let $Q(\bar x)=(\Omc,\Sigma,q) \in (\ALCI,\text{UCQ})$ with
$q=p_1 \vee \cdots \vee p_n$, and let \Dmc be a $\Sigma$-database.
Then $Q^\approx_{\ELI^u_\bot}(\Dmc)=Q_1(\Dmc) \cup \cdots \cup
Q_n(\Dmc)$, $Q_i=(\Omc^\approx_{\ELI^u_\bot},\Sigma,p_i)$ for
$1 \leq i \leq n$.
\end{lemma}
%
%
We now describe an algorithm that establishes the upper bounds in Points~1
and~2 of Theorem~\ref{thm:mainone} and that we refer to as a
\emph{careful chase}, see also \cite{DBLP:conf/ijcai/BienvenuOSX13}.

Assume that we are given an OMQ
$Q(\bar x)=(\Omc,\Sigma,q) \in (\ALCI,\text{CQ})$, a $\Sigma$-database
\Dmc, and a tuple $\bar a \in \mn{adom}(\Dmc)^{|\bar x|}$. We use
$\mn{trees}(q)$ to denote the set of bELIQs that can be obtained from
CQ $q$ by first quantifying all variables, then taking a contraction,
then an induced subquery, and then choosing at most one variable as
the answer variable. In addition, $\mn{trees}(q)$ contains all AQs
$A(x)$ with $A$ a concept name used in~\Omc.

The algorithm first extends \Dmc to a database $\Dmc'$ as follows:
\begin{itemize}

\item whenever $\Dmc, \Omc^\approx_{\ELI^u_\bot} \models p()$ with
  $p() \in \mn{trees}(q)$ a BELIQ, then take a disjoint copy of
  $\Dmc_p$ and add it to~\Dmc; 

\item whenever $\Dmc, \Omc^\approx_{\ELI^u_\bot} \models p(a)$ with
  $p(x) \in \mn{trees}(q)$ an ELIQ, then take a disjoint copy of
  $\Dmc_p$ and add it to~\Dmc, glueing the root $x$ of $\Dmc_p$
  to $a$. 

\end{itemize}
$\Dmc, \Omc^\approx_{\ELI^u_\bot} \models p()$ and
$\Dmc, \Omc^\approx_{\ELI^u_\bot} \models p(a)$ can be decided in time
$2^{O(||Q||)} \cdot O(||\Dmc||)$ by Point~3 of
Theorem~\ref{thm:mainone}. Note that $\Dmc'$ is a subdatabase of 
$\mn{ch}_{\Omc^\approx_{\ELI^u_\bot}}(\Dmc)$, which is why we speak of 
a careful chase. The algorithm then carries out the following steps:
\begin{description}

\item[\textnormal{(i)}] If $\Dmc$ is unsatisfiable w.r.t.\ $\Omc^\approx_{\ELI^u_\bot}$,
  then returns `yes';

\item[\textnormal{(ii)}] check whether $\bar a \in q(\Dmc')$ and return
  the result.

\end{description}
By Point~ 2 of Lemma~\ref{lem:crucial}, the unsatisfiability check in
(i) is equivalent to checking whether $\Dmc^\approx_{\emptyset}$ is
unsatisfiable w.r.t.\ \Omc. This is the case if and only if
$\Dmc^{\approx}_{\emptyset},\Omc\models \exists x \, A(x)$ with $A$ a
fresh concept name, which can be decided in time
$2^{O(||Q||)} \cdot O(||\Dmc||)$ by Point~3 of
Theorem~\ref{thm:mainone}.

Checking $\bar a \in q(\Dmc')$ in (ii) can be implemented using brute
force to attain \ExpTime combined complexity and \PTime data
complexity or using as a blackbox an algorithm that runs within the
time requirements of fixed-parameter tractability to attain FPT when
$q$ is of bounded treewidth. We prove in the appendix that the
algorithm is correct and achieves the upper bounds stated in
Theorem~\ref{thm:mainone}.

The \ExpTime lower bounds in Theorem~\ref{thm:mainone} are proved by a
straightforward reduction from the subsumption of concept names in
\ALC~\cite{DBLP:books/daglib/0041477}: $A$ is subsumed by $B$ w.r.t.\
\Omc iff $\{A(a)\},\Omc \models B(a)$ iff
$\{ A(a)\},\Omc^\approx_{\ELI^u_\bot} \models B(a)$. This trivial
reduction also shows that using fragments of $\ELI^u_\bot$ such as
$\EL_\bot$ or $\EL^u_\bot$ as a target for ontology approximation
cannot improve combined complexity. The \PTime lower bound in Point~1
is inherited from OMQ evaluation in $(\EL,\text{AQ})$
\cite{DBLP:conf/ijcai/CalvaneseGLLR15}.

\section{Tree-Database Relaxing Approximation}
\label{sect:treedb}

We study $\Dmf_{\!\curlywedge}$-database relaxing approximation that
for the sake of readability we from now on refer to as
tree-database relaxing approximation.
%
%
%
  We start with observing that tree-database
  relaxing approximation is incomparable to $\ELI^u_\bot$-ontology
  relaxing approximation.
\begin{exmp}
  \label{ex:first}
  Let $Q(x)=(\Omc,\Sigma,A(x)) \in (\ALCI,\text{AQ})$ where
  $$\Omc = \{ P \sqcap \exists r . P \sqsubseteq A,\ \neg P \sqcap
  \exists r . \neg P \sqsubseteq A \} \quad\Sigma = \{P,r,A\}.
 $$
  Consider $\Dmc= \{r(a,a)\}$. Then
  $a \notin Q^\approx_{\ELI^u_\bot}(\Dmc)$, but
  $a \in Q(\Dmc_{\{a\}}^\approx)$ since $\Dmc^\approx_{\{a\}}=\Dmc$.
  %

  \smallskip
  
  Conversely, let $Q(x)=(\emptyset,\{r\},q) \in (\ALCI,\text{CQ})$ where
  $$q(x) = \exists y_1 \exists y_2 \exists y_3 \, r(x,y_1) \wedge
  r(y_1,y_2) \wedge r(y_2,y_3) \wedge r(y_3,y_1),$$
  and $\Dmc = \{ r(a,b_1), r(b_1,b_2), r(b_2,b_3),r(b_3,b_1) \}$. Then
  $Q^\approx_{\ELI^u_\bot}(\Dmc) = \{a\}$, but
  $Q(\Dmc^\approx_{\{a\}}) = \emptyset$.
\end{exmp}
%
%
%
Note that the OMQs in Example~\ref{ex:first} are based on CQs that are
not bELIQs. This is no coincidence, as the following is a consequence
of Lemma~\ref{lem:crucial} and the fact that
$\Dmc^\approx_\emptyset \subseteq \Dmc^\approx_S$ for all databases
\Dmc and $S \subseteq \mn{adom}(\Dmc)$.
\begin{prop}
\label{prop:eliqcoincide}
In $(\ALCI,\text{bELIQ})$, tree-database relaxing OMQ evaluation
is at least as complete as $\ELI^u_\bot$-ontology relaxing OMQ evaluation.
\end{prop}
The converse of 
Proposition~\ref{prop:eliqcoincide} fails, as per the
first part of Example~\ref{ex:first}.
The first main result of this section follows. 
\begin{restatable}{theorem}{thmtreeunravmain}
\label{thm:treeunravmain}
Tree-database relaxing OMQ evaluation is
\begin{enumerate}

\item \TwoExpTime-complete in combined complexity and in linear time
  in data complexity (thus FPT) with running
  time $2^{2^{O(||Q||)}} \cdot O(||\Dmc||)$ in $(\ALCI,\Qmc)$, $\Qmc
  \in \{ \text{CQ}, \text{UCQ}, \text{CQ}^{\text{tw}}_k,
  \text{UCQ}^{\text{tw}}_k \mid k \geq 1 \}$;


\item \ExpTime-complete in combined complexity and in linear time in
  data complexity (thus FPT) with running time
  $2^{O(||Q||)} \cdot O(||\Dmc||)$ in $(\ALC,\Qmc)$ and
  $(\ALCI,\Qmc)$, $\Qmc \in \{ \text{AQ}, \text{bELIQ} \}$.
\end{enumerate}
\end{restatable}
The upper bound in Point~2 of
Theorem~\ref{thm:treeunravmain} is proved similarly to Point~3
of Theorem~\ref{thm:mainone}, by reduction to the unsatisfiability of
propositional Horn formulas.  To prove the upper bound in Point~1, we
first show that one can construct from an OMQ
$Q=(\Omc,\Sigma,q) \in (\ALCI,\text{UCQ})$ and $\Sigma$-database \Dmc
an OMQ $Q'=(\Omc,\Sigma',q')$ and a $\Sigma'$-database $\Dmc'$
such that $Q(\Dmc)=Q(\Dmc')$ and $q'$ is a conjunction of disjunctions
of BELIQs. This again enables a reduction to the
unsatisfiability of propositional Horn formulas. Note that a
counterpart of Lemma~\ref{lem:UCQtoCQ} does not hold for tree-database
relaxing approximation and thus we treat UCQs directly.  The lower
bounds are trivial as (non-approximate) evaluation in all mentioned
OMQ languages is hard already on databases of the form
$\Dmc=\{A(a)\}$, which satisfy $\Dmc^\approx_{\{a\}}=\Dmc$, and
for CQs of bounded treewidth \cite{DBLP:conf/cade/Lutz08}.

In contrast to $\ELI^u_\bot$-ontology relaxing approximation, we
achieve no improvement in combined complexity in the \ALCI case, but
we achieve linear time data complexity also for (U)CQs that are not of
bounded treewidth. Informally, this is because database relaxing OMQ
evaluation also approximates answers to the actual query (if it is a
(U)CQ) while ontology relaxing evaluation only approximates the impact
of the ontology. For comparison we recall that without ontologies,
evaluating CQs of unbounded treewidth is $W[1]$-hard, thus most
likely not linear time in data complexity and not even FPT~\cite{DBLP:journals/jacm/Grohe07}. 

We next prove the second main result of this section. Recall that
non-approximate OMQ evaluation in $(\ALC,\text{CQ})$ and
$(\ALC,\text{UCQ})$ is \ExpTime-complete in combined complexity and
{\sc coNP}-complete in data complexity. Suprisingly, tree-database
relaxing evaluation in these OMQ languages is \emph{harder} in
combined complexity than non-approximate evaluation while it is
simpler in data complexity (the latter by Theorem~\ref{thm:treeunravmain}).
\begin{restatable}{theorem}{thmtwoexpwithouti}
  \label{thm:twoexpwithouti}
  Tree-database relaxing OMQ evaluation is \ExpSpace-hard in
  $(\ALC,\text{CQ})$ and \TwoExpTime-hard in   $(\ALC,\text{UCQ})$.
\end{restatable}
The proof of Theorem~\ref{thm:twoexpwithouti} adapts hardness proofs
from \cite{DBLP:conf/cade/Lutz08} for non-approximate OMQ evaluation in
$(\ALCI,\text{CQ})$, simulating inverse roles by making use of the
unraveled database.  

\section{TGD-Ontology Relaxing Approximation}
\label{sect:tgd}

We generalize ontology relaxing approximation from $\ELI^u_\bot$
to TGDs that are frontier-one and have bodies and heads of bounded
treewidth. This yields an infinite hierarchy of increasingly
more complete approximations.

A TGD
$\vartheta=\phi(\bar x,\bar y) \rightarrow \exists \bar z \, \psi(\bar
x,\bar z)$ is an \emph{$\ell,k,\ell',k'$-TGD}, $\ell < k$ and
$\ell' < k'$, if its body has treewidth $(\ell,k)$ and its head has
treewidth $(\ell',k')$ when viewed as CQs in which all variables are
quantified.  Every $\ELI^u_\bot$-CI can be written as a frontier-one
$1,2,1,2$-TGD, but there are frontier-one $1,2,1,2$-TGDs that are not
$\ELI^u_\bot$-CIs, such as $r(x,x) \rightarrow s(x,x)$. From now on,
when speaking about $\ell,k,\ell',k'$-TGDs, we generally mean
frontier-one TGDs.  An \emph{$\ell,k,\ell',k'$-TGD-ontology} is an
FO-ontology that consists only of (frontier-one)
$\ell,k,\ell',k'$-TGDs. If any of $\ell,k,\ell',k'$ is not bounded, we
assign to it value $\omega$.

We study $\ell,k,\ell',k'$-TGD-ontology relaxing OMQ evaluation.
Recall that, by Lemma~\ref{lem:infont}, computing such approximations
for an OMQ $Q \in (\text{FO},\text{UCQ})$ corresponds to evaluating
the OMQ $Q^\approx_{\ell,k,\ell',k'\text{-TGD}}$ which is based on ontology
$\Omc^\approx_{\ell,k,\ell',k'\text{-TGD}}$. For brevity, we
drop the suffix `-TGD' and for instance speak about
$\ell,k,\ell',k'$-ontology relaxing OMQ evaluation and write
$Q^\approx_{\ell,k,\ell',k'}$.

We first observe that restricting the treewidth of the body is
necessary to attain \PTime data complexity and that we cannot hope to
attain the favourable combined complexity enjoyed by
$\ELI^u_\bot$-ontology relaxing approximation for every value
of $\ell'$ and $k'$. 
%
The following is a consequence of
Example~\ref{ex:threecolagain} below and of hardness proofs
in~\cite{DBLP:conf/cade/Lutz08}.
\begin{restatable}{prop}{prophowwechoose}
  \label{prop:howwechoose}
  $\omega,\omega,1,2$-ontology relaxing OMQ evaluation in
  $(\ALC,\text{AQ})$ is {\sc coNP}-hard in data complexity and there are
  $\ell',k'$ such that $1,2,\ell',k'$-ontology relaxing OMQ evaluation in
  $(\ALCI,\text{CQ})$ is \TwoExpTime-hard in combined complexity.
\end{restatable}
To tame the combined complexity of TGD-ontology-relaxing
approximation, we concentrate on the case where $\ell'=1$. We next
consider the choice of values for $k$ and $k'$, the treewidth
of rule bodies and heads. We first show that $k$ gives rise to an
infinite hierarchy of increasingly more complete
approximations.
%
\begin{exmp}
\label{ex:threecolagain}  
Consider the OMQ $Q=(\Omc,\{e\},\exists x \, D(x))$ from
Example~\ref{ex:3col} that expresses non-3-colorability. For every
$\{e\}$-database \Dmc, there is a $k \geq 1$ such that
$\Dmc \models Q$ iff $\Dmc \models Q^\approx_{1,k,1,2}$: for
$k=|\mn{adom}(\Dmc)|$, $\Omc^\approx_{1,k,1,2}$ contains
$q_\Dmc \rightarrow \exists x\, D(x)$ if \Dmc is not 3-colorable,
where $q_\Dmc$ is \Dmc viewed as a CQ. Unless
$\PTime =\NPclass$, there are thus no $\ell,k,\ell',k'$ such that
$Q^\approx_{\ell,k,\ell',k'} \supseteq Q^\approx_{1,k+p,1,2}$ for all
$p > 0$. In fact, the existence of such $\ell,k,\ell',k'$ implies that
3-colorability is in \PTime by Theorem~\ref{thm:tgdmain} below.
  %
  
\end{exmp}
Example~\ref{ex:threecolagain} also shows that 
TGD-ontology relaxing approximations are more complete than 
$\ELI^u_\bot$-ontology relaxing approximations, c.f.\
Example~\ref{ex:3col}. 

\begin{figure}
	\begin{boxedminipage}{\columnwidth}
      \centering
\begin{tikzpicture}[->,>=stealth',level distance = 1cm,font=\sffamily\small]
\tikzstyle{node}=[inner sep=1.0pt]
\node [node] (01) at (0,0) {$x_{0,1}$};
\node [node] (A01) at (-0.45,0.275) {$A_{0,1}$};
\node [node] (11) at (1.5,0) {$x_{1,1}$};
\node [node] (A11) at (1.895,0.275) {$A_{1,1}$};
\node [node] (21) at (3,0) {$x_{2,1}$};
\node [node] (A21) at (3.45,0.275) {$A_{2,1}$};

\node [node] (02) at (0,1.1) {$x_{0,2}$};
\node [node] (A02) at (-0.45,1.375) {$A_{0,2}$};
\node [node] (12) at (1.5,1.1) {$x_{1,2}$};
\node [node] (A12) at (1.95,1.375) {$A_{1,2}$};
\node [node] (22) at (3,1.1) {$x_{2,2}$};
\node [node] (A22) at (3.45,1.375) {$A_{2,2}$};

\node [node] (00) at (0,-1.1) {$x_{0,0}$};
\node [node] (A00) at (-0.45,-0.825) {$A_{0,0}$};
\node [node] (10) at (1.5,-1.1) {$x_{1,0}$};
\node [node] (A10) at (1.895,-0.825) {$A_{1,0}$};
\node [node] (20) at (3,-1.1) {$x_{2,0}$};
\node [node] (A20) at (3.45,-0.85) {$A_{2,0}$};

\draw [->] (11) -- (01) node[midway, below] {$r$};
\draw [->] (11) -- (21) node[midway, below] {$r$};
\draw [->] (11) -- (12) node[midway, left] {$r$};
\draw [->] (11) -- (10) node[midway, left] {$r$};

\draw [->] (02) -- (01) node[midway, right] {$r$};
\draw [->] (02) -- (12) node[midway, below] {$r$};

\draw [->] (22) -- (21) node[midway, left] {$r$};
\draw [->] (22) -- (12) node[midway, below] {$r$};

\draw [->] (00) -- (01) node[midway, right] {$r$};
\draw [->] (00) -- (10) node[midway, below] {$r$};

\draw [->] (20) -- (21) node[midway, left] {$r$};
\draw [->] (20) -- (10) node[midway, below] {$r$};

\end{tikzpicture}
	\end{boxedminipage}
\caption{CQ from Example~\ref{ex:leif} for the case $m=2$.}
\label{fig:picEx8}
\end{figure}
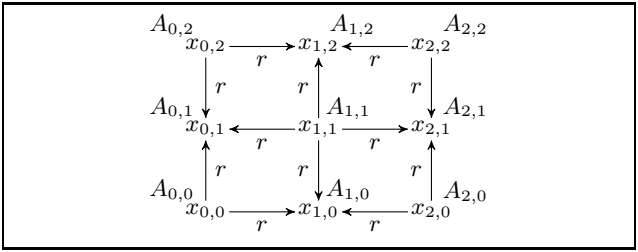

For $k'$, we make the weaker observation that there is no maximum
value to be used uniformly for all OMQs.
\begin{exmp}
  \label{ex:leif}
  For $m \geq 2$, consider the Boolean CQ
\[q_m = \bigwedge_{\substack{i,j,i',j' \in \{1,\ldots,m\} \\ i+j \text{ is even} \\ |i-i'| + |j-j'| = 1}} r(x_{i,j},x_{i',j'}) \wedge \bigwedge_{i,j \in \{1,\ldots,m\}} A_{i,j}(x_{i,j})
\]
which 
takes the form of an $m \times m$ grid in which the direction of the
roles alternate and every point in the grid is labeled with a unique
concept name, see Figure~\ref{fig:picEx8}.  Consider the following two
ways of contracting $q_m$ into a path: identify $x_{i,j}$ with
$x_{i',j'}$ if $i+j=i'+j'$ or identify $x_{i,j}$ with $x_{i',j'}$ if
$i-j=i'-j'$. Let $C_1$ and $C_2$ be the two paths obtained, viewed as
$\ELI$-concepts.

Set $\Omc_m = \{A \sqsubseteq C_1 \sqcup C_2\}$ and
$Q_m() = (\Omc_m, \{A\}, q_m)$, and let $\Dmc = \{A(a)\}$.  Then $(\Omc_m)_{1,2,1,m^2}^\approx$ contains the TGD
$A \rightarrow q_m$, so $\Dmc \models (Q_m)_{1, 2, 1, m^2}^\approx$.
In contrast, we argue in the appendix that
$\Dmc \not\models (Q_m)_{\omega, \omega, 1, m^2-1}^\approx$.
\end{exmp}
For every fixed OMQ, however, there is a maximum useful value for
$k'$. The next proposition is established analyzing the 
algorithm in the proof of Theorem~\ref{thm:tgdmain}
below.
\begin{restatable}{prop}{propmaxk}
  Let $\ell,k \geq 1$ with $\ell < k$.  
  For every $Q(\bar x)=(\Omc,\Sigma,q) \in (\ALCI,\text{UCQ})$
  and  $k' \geq |\mn{var}(q)|$, 
    $\Qmc^\approx_{\ell,k,1,|\mn{var}(q)|} \equiv
  \Qmc^\approx_{\ell,k,1,k'}$.
\end{restatable}

The main result of this section is as follows.
\begin{theorem}
  \label{thm:tgdmain}
  Let $\Lmc \in \{ \ALC, \ALCI \}$ and $\ell,
  k,k' \geq 1$ with $\ell < k$.
  Then 
    $\ell,k,1,k'$-ontology relaxing OMQ evaluation is
  \begin{enumerate}

  \item \ExpTime-complete in combined complexity and \PTime-complete
    in data complexity in $(\Lmc,\Qmc)$,
    $\Qmc \in \{ \text{AQ}, \text{CQ}, \text{UCQ} \}$;

  \item FPT 
    in $(\Lmc,\Qmc)$, $\Qmc \in \{ \text{CQ}^{\text{tw}}_p, \text{UCQ}^{\text{tw}}_p
    \mid p \geq 1 \}$.

  \end{enumerate}

\end{theorem}
So TGD-ontology relaxing approximation inherits the good
computational properties of $\ELI^u_\bot$-ontology relaxing
approximation except for linear time for bELIQs, while being 
significantly more complete.

The lower bounds are proved exactly as for $\ELI^u_\bot$-ontology
relaxing approximation, see Section~\ref{sect:eliu}.  For the upper
bounds, we treat the CQs in a UCQ independently and use a careful
chase algorithm that essentially follows the lines of the careful
chase presented in Section~\ref{sect:eliu}.  An important difference
is that a counterpart of Lemma~\ref{lem:crucial} in which
$Q^\approx_{\ELI^u_\bot}$ is replaced with $Q^\approx_{\ell,k,1,k'}$
and $\Dmc^\approx_\emptyset$ with an unraveling of \Dmc into a
database of treewidth $\ell,k$ fails to hold if $k' > k$ (the `only
if' direction of Point~1 fails). We resort to
Theorem~\ref{thm:eliqsgen} below, which is a central
ingredient to the proof. 
For a database \Dmc and $\ell,k \geq 1$, with
  $\ell < k$, $\Dmc^\approx_{\ell,k}$ denotes the unraveling
  of $\Dmc$ into a database of treewidth $(\ell,k)$, defined 
  in the
  appendix. While
  \mbox{$\mn{adom}(\Dmc) \cap \mn{adom}(\Dmc^\approx_{\ell,k})=\emptyset$},
  $\Dmc^\approx_{\ell,k}$ contains constants that
  are `copies' of each $a \in \mn{adom}(\Dmc)$. We use
  $\langle \Dmc^\approx_{\ell,k},a \rangle$ to denote a database
  obtained from $\Dmc^\approx_{\ell,k}$ by choosing a copy of $a$ in
  $\Dmc^\approx_{\ell,k}$ and renaming it back to $a$. With
  $\langle \Dmc^\approx_{\ell,k},() \rangle$, we mean~$\Dmc^\approx_{\ell,k}$.  
\begin{restatable}{theorem}{thmeliqsgen} 
\label{thm:eliqsgen} 
Let $\ell,k,k' \geq 1$ with $\ell < k$.  Given an OMQ
$Q(\bar x)=(\Omc,\Sigma,q) \in (\ALCI,\text{CQ})$ of arity at most one
and with $\Dmc_q$ of treewidth $(1,k')$, a $\Sigma$-database
\Dmc, and $\bar a \in \mn{adom}(\Dmc)^{|\bar x|}$, deciding whether
$\bar a \in Q(\langle \Dmc^{\approx}_{\ell,k},\bar a\rangle)$ is in \ExpTime in
combined complexity and FPT.\footnote{Note that $\Dmc_q$ being of
treewidth $(1,k')$ is a stricter condition than $q$ being of treewidth
$(1,k')$.} 
\end{restatable}
The proof of Theorem~\ref{thm:eliqsgen} uses
alternating tree automata. 

\smallskip

We close this section with commenting on our use of frontier one
TGDs. Using arguments similar to those in
Example~\ref{ex:threecolagain}, it is easy to see that approximating
an OMQ $Q(\bar x)=(\Omc,\Sigma,q) \in (\ALCI,\text{CQ})$ in terms of
unrestricted TGDs does actually not result in any approximation at
all, that is, $Q(\Dmc)=Q^\approx_{\text{TGD}}(\Dmc)$ for all
$\Sigma$-databases \Dmc. We conjecture
%
that the results 
in this section 
generalize 
to frontier-guarded TGDs
\cite{DBLP:conf/kr/BagetLM10}. The gain in completeness
appears
to be modest. 

\section{BTW-Database Relaxing Approximation}
\label{sect:btw}

We study database relaxing approximation based on databases of
bounded treewidth. For 
$\ell,k \geq 1$ with $\ell < k$, let $\Dmf_{\ell,k}$ denote the class
of pointed databases $(\Dmc,\bar a)$ such that the restriction of \Dmc
to domain $\mn{adom}(\Dmc)\setminus \bar a$ has treewidth
$\ell,k$. For readability, we speak of $\ell,k$-database relaxing
approximation in place of $\Dmf_{\ell,k}$-database relaxing
approximation. As for TGD-ontology relaxing approximations,
the parameter $k$ gives rise to an infinite hierarchy of increasingly
more complete approximations. 

%
%
We first observe a counterpart of Lemma~\ref{lem:unravfund}. Let \Dmc
be a database, $\bar a$ a tuple over $\mn{adom}(\Dmc)$, and $\ell,k
\geq 1$ with $\ell < k$.  With $\Dmc^\approx_{\bar a,\ell,k}$, we
denote the database obtained by unraveling $\Dmc$ such that
$(\Dmc^\approx_{\bar a,\ell,k},\bar a) \in \Dmf_{\ell,k}$. Details are in the appendix.

\begin{restatable}{lemma}{lemunravfundtwo}
  \label{lem:unravfundtwo}
  Let $Q =(\Omc,\Sigma,q) \in (\text{FO},\text{UCQ})$ and
  \mbox{$\ell,k \geq 1$} with $\ell < k$.  Then for all $\Sigma$-databases
  \Dmc and $\bar a \in \mn{adom}(\Dmc)^{|\bar x|}$,
  $\bar a \in \mn{app}_{\Dmf_{\ell,k}}(Q,\Dmc)$ iff $\bar a \in 
  Q(\Dmc^\approx_{\bar a,\ell,k})$. 
\end{restatable}

We next
relate $\ell,k,\ell',k'$-ontology relaxing approximation to 
$\ell,k$-database relaxing approximation. 
\begin{exmp}
  \label{ex:foobarstretch}
  Let $Q_n()=(\emptyset,\{r\},q_n) \in (\ALCI,\text{CQ})$ where
  $$q_n =
  \exists x_1 \cdots \exists x_n \bigwedge_{1 \leq i,j \leq n}
  r(x_i,x_j),$$ and let $\Dmc_n = \{ r(a_i,a_j) \mid 1 \leq i,j \leq n
  \}$. Then $\Dmc_n \models (Q_n)^\approx_{1,2,1,2}$, 
  but
  $(\Dmc_n)^\approx_{\emptyset,n-2,n-1} \not\models Q_n$.

  \smallskip

  Conversely, take $Q(x_1,x_2)=(\Omc,\Sigma,q) \in (\ALCI,\text{CQ})$ where
  $$
  \begin{array}{rcl}
  \Omc &=& \{ A \sqsubseteq \forall s . B \sqcup \forall s . \forall s . B
     \} \\[0,5mm]
  \Sigma &=& \{A,A_1,A_2,A_3,B,r,s\} 
  \end{array}
  $$
  $$
  \begin{array}{rcl}
      q &=& \exists y_1\cdots \exists y_5 
            r(x_1,y_1) \wedge \bigwedge_{1 \leq i \leq 4}
            r(y_i,y_{i+1}) \,
            \wedge \\[0.5mm]
    && r(y_5,x_2) \wedge B(y_3)
  \end{array}
  $$
  and let 
  $$
  \begin{array}{r@{\;}c@{\;}l}
    \Dmc &=& \{ A(a_1), r(a_1,a_2),\dots,r(a_3,a_4), \\[1mm]
    && \;\; s(a_1,a_{2}), s(a_2,a_{3}), r(a_2,a_2), r(a_4,a_4)
  \}.
  \end{array}
  $$
  %
  The algorithms underlying the theorems in this and the previous
  section can be used to show that $(a_1,a_5) \in
  Q(\Dmc^\approx_{\emptyset,1,2})$, but
  $Q^\approx_{\omega,\omega,\omega,\omega}(\Dmc)=\emptyset$.
\end{exmp}
A straightforward variation of Example~\ref{ex:threecolagain} shows
that the parameter $k$ indeed gives rise to an infinite hierarchy of
increasingly more complete approximations.
%
%
\begin{restatable}{prop}{propeliqbtw}
\label{prop:eliqbtw}
Let $\ell,k,\ell',k' \geq 1$ with $\ell < k$ and $\ell' < k'$.
\begin{enumerate}

\item In
   $(\ALCI,\text{UCQ}^{\mn{tw}}_{\ell,k})$, 
%
  $\ell,k$-database relaxing 
  OMQ evaluation is at least as complete as 
  $\ell,k,\ell',k'$-ontology 
  relaxing OMQ evaluation. 

\item For OMQ from $(\ALCI,\text{CQ}^{\mn{tw}}_{\ell',k'})$ of arity
  at most $r \leq 1$, \mbox{$\ell+r,k+r,\ell',k'$}-ontology 
  relaxing OMQ evaluation is at least as complete as 
  $\ell,k$-database relaxing 
  OMQ evaluation.

\end{enumerate}
\end{restatable}
%
%
We remark that Point~2 of Proposition~\ref{prop:eliqbtw} no longer holds if CQs
are replaced by UCQs.
We now formulate the main result of this section.
\begin{restatable}{theorem}{thmbtwunravmain}
  \label{thm:btwunravmain}
  Let $\ell, k \geq 1$,
  $\ell < k$. Then $\ell,k$-database relaxing OMQ evaluation is
\begin{enumerate}

\item \TwoExpTime-complete in combined
  complexity and fixed-parameter
  tractable with
  double exponential running time in $(\ALCI,\Qmc)$, $\Qmc  \in \{
  \text{CQ}, \text{UCQ} , \text{CQ}^{\text{tw}}_p,
  \text{UCQ}^{\text{tw}}_p \mid p \geq 1\}$;


\item \ExpTime-complete in combined complexity  and fixed-parameter
  tractable with
  single exponential running time in $(\ALC,\Qmc)$ and $(\ALCI,\Qmc)$,
  $\Qmc \in \{ \text{AQ}, \text{bELIQ} \}$.
\end{enumerate}
\end{restatable}
We thus achieve FPT even for (U)CQs of unbounded treewidth, as
for tree-database relaxing approximations,
but 
not linear time. The
lower bounds are proved exactly as for
Theorem~\ref{thm:treeunravmain}. The upper bounds are shown by an
elimination approach. We remark that $\ell,k$-database relaxing
approximation coincides with the answers given by canonical $(\ell,k)$-Datalog
programs, see 
\cite{DBLP:journals/lmcs/FeierKL19}. 

%
%
%
%
%
%
%

Recall that for tree-database unraveling approximation, the combined
complexity \emph{increases} for OMQ languages based on \ALC and
(U)CQs. It seems clear that the same is true for BTW-database
relaxing approximation and that it can be proved by adapting the
proof of Theorem~\ref{thm:twoexpwithouti}. A central idea is to
replace single constants in databases by cliques. 
\begin{conjecture}
  Let $\ell, k \geq 1$, $\ell < k$. Then  $\ell,k$-database
  relaxing OMQ evaluation in $(\ALC,\text{UCQ})$ is \TwoExpTime-hard.
\end{conjecture}
%
%
%

\section{Approximation from Above}
\label{sect:fromabove}

  \begin{figure*}

    \centering 
\begin{tabular}{ |c|c|c||c|c| }
 \hline  
 & \multicolumn{2}{|c||}{Ontology relaxing} & \multicolumn{2}{|c|}{Database relaxing}\\
 \hline  
 & $\ELI^u_\bot$ & TGD & trees & bounded treewidth\\
  \hline  
  \hline  
  $(\ALC(\Imc),\text{bELIQ})$ & \textsc{lin} / \textsc{ExpTime} & FPT / \textsc{ExpTime} & \textsc{lin} / \textsc{ExpTime} & FPT / \textsc{ExpTime}\\
 \hline  
 $(\ALC,\text{CQ})$ & \multicolumn{2}{|c||}{\PTime~/ \textsc{ExpTime}} & FPT / \textsc{ExpSpace}--\textsc{2ExpTime} & FPT / in \textsc{2ExpTime}\\
 \hline  
  $(\ALC,\text{UCQ})$ & \multicolumn{2}{|c||}{\PTime~/ \textsc{ExpTime}} & FPT / \textsc{2ExpTime} & FPT / in \textsc{2ExpTime}\\
 \hline  
   $(\ALCI,\text{(U)CQ}^{\text{tw}}_k)$ & \multicolumn{2}{|c||}{FPT / \textsc{ExpTime}} & \multicolumn{2}{|c|}{FPT / \textsc{2ExpTime}}\\
 \hline  
    $(\ALCI,\text{(U)CQ})$ & \multicolumn{2}{|c||}{PTime / \textsc{ExpTime}} & \multicolumn{2}{|c|}{FPT / \textsc{2ExpTime}}\\
 \hline  
\end{tabular}

\caption{Results for approximate OMQ evaluation, data and  
  parametric complexity / combined complexity. }
\label{fig:results}
\vspace*{-3mm}
\end{figure*}
The approximations studied so far are from below, thus sound but
incomplete. We define dual approximations from above that are complete
but unsound: ontology strengthening approximation and database
strengthening approximation. It turns out that
these are computationally less well-behaved. While
the former may increase combined complexity,
the latter does not even enjoy \PTime data
complexity.

We start with ontology strengthening approximation. 
For an OMQ $Q(\bar x)=(\Omc,\Sigma,q)$, a
$\Sigma$-database \Dmc, and an ontology language $\Lmc'$,
we use $\mn{app}^\uparrow_{\Lmc'}(Q,\Dmc)$ to
denote the set of tuples $\bar a \in \mn{adom}(\Dmc)^{|\bar x|}$ such
that $\bar a \in Q'(\Dmc)$ for all OMQs $Q'=(\Omc',\Sigma,q)$ where $\Omc'$
is an $\Lmc'$-ontology with $\Omc' \models \Omc$.  Every choice of
$(\Lmc, \Qmc)$ and $\Lmc'$ gives rise to an approximate OMQ evaluation
problem.
\begin{center}
	\fbox{\begin{tabular}{@{}l@{\;}l}
                \multicolumn{2}{@{}l}{
                $\Lmc'$-ontology strengthening evaluation in $(\Lmc,\Qmc)$}
			\\[.5mm]{\small INPUT} : & OMQ $Q(\bar x) =
                                              (\Omc,\Sigma,q)
                                             \in (\Lmc,\Qmc)$,  \\
&			$\Sigma$-database $\Dmc$,  tuple $\bar a \in \mn{adom}(\Dmc)^{|\bar x|}$
			\\[.5mm]
                {\small OUTPUT} : &  `yes' if
                                    $\bar a \in \mn{app}_{\Lmc'}^\uparrow(Q,\Dmc)$
                                            and `no' otherwise
	\end{tabular}}
\end{center}
%

We consider $\ELI_\bot$-ontology strengthening OMQ evaluation in
$(\ELIU_\bot, \text{UCQ})$ where $\ELIU_\bot$ is the extension of
$\ELI_\bot$ with disjunction.  Note that we can find an implying
$\ELI_\bot$-ontology for every $\ELIU_\bot$-ontology, namely
$\{ \top \sqsubseteq \bot \}$.  In contrast to ontology relaxing
approximation, it does not seem beneficial to use the
universal role.  The following example illustrates unsoundness.
\begin{exmp}
  \label{ex:unsoundtwo}
  Take $Q(x)=(\Omc,\Sigma,q)$ where
  $$
  \begin{array}{r@{\;}c@{\;}l}
    \Omc &=& \{ A \sqsubseteq A_1 \sqcup A_2,\\[0.5mm]
    &&\phantom{\{}\exists r .(A_i \sqcap
  B_1) \sqcap \exists r. (A_i \sqcap B_2) \sqsubseteq B \mid i \in
             \{1,2\} \} \\[0.5mm]
    \Sigma&=&\{A,A_1,A_2,B,B_1,B_2,r\} \\[0.5mm]
    q&=&B(x).
      \end{array}
$$
  Let $\Dmc=\{r(a,b_1),r(a,b_2),A(b_1),B_1(b_1),A(b_2),B_2(b_2)\}$. Then
  $a \in \mn{app}^\uparrow_{\ELI_\bot}(\Dmc)$ as every
  $\ELI_\bot$-ontology
  $\Omc'$ with $\Omc' \models \Omc$ implies 
  $A \sqsubseteq A_1$ or   $A \sqsubseteq A_2$.
  But
  $a \notin Q(\Dmc)$.
\end{exmp}
Let $\Omc$ be an $\ELIU_\bot$-ontology.
A set $\Mmc$ of $\ELI_\bot$-ontologies is an \emph{exhaustive
$\ELI_\bot$-approximation set for~$\Omc$} if 
$\widehat \Omc \models \Omc$ for every $\widehat \Omc \in \Mmc$, and for every
$\ELI_\bot$-ontology $\Omc'$ with $\Omc' \models \Omc$, there is an
$\widehat \Omc \in \Mmc$ such that $\Omc' \models \widehat \Omc$.
Such sets \Mmc are interesting because for all OMQs
$Q=(\Omc,\Sigma,q)$ with $q$ a UCQ and all $\Sigma$-databases~\Dmc, 
$\mn{app}^\uparrow_{\ELI_\bot}(Q,\Dmc) = \bigcap_{\widehat \Omc \in \Mmc}
Q_{\widehat\Omc}(\Dmc)$ where
$Q_{\widehat\Omc}=(\widehat\Omc,\Sigma,q)$. Consider for instance $\Omc=\{\top \sqsubseteq A_1 \sqcup A_2\}$, for which $\{ \Omc_1,\Omc_2\}$
is an exhaustive $\ELI_\bot$-approximation set where
$\Omc_i=\{\top \sqsubseteq A_i\}$.  In the appendix, we show how to
construct a finite exhaustive $\ELI_\bot$-approximation set for any
given 
$\ELIU_\bot$-ontology 
and use this to prove the upper bounds in the following result.
\begin{restatable}{theorem}{thmontabove}
  \label{thm:ontabove}
  Let $\Lmc \in \{ \text{AQ},\text{CQ},\text{UCQ} \}$.
$\ELI_\bot$-ontology strengthening OMQ evaluation in
$(\ELIU_\bot,\Lmc)$ is $\TwoExpTime$-complete in combined
complexity and FPT with double exponential running time.
\end{restatable}
The lower bound is proved by a reduction from the word problem for a
suitable kind of alternating Turing machine. We consider the lower
bound for $(\ELIU_\bot,\text{AQ})$ surprising as non-approximate OMQ
evaluation is only \ExpTime-complete
\cite{DBLP:conf/cade/Lutz08}. Thus, approximate OMQ evaluation is
significantly harder, the only result of this kind in the current
paper that applies to AQs. It is amusing to note that the lower bound
depends only on disjunction on the \emph{left} hand side of concept
inclusions, which are syntactic sugar, but not on the seemingly much
more `dangerous' disjunctions on the right hand side. It is in fact a
byproduct of our proofs that, without disjunctions on the left,
$\ELI_\bot$-ontology strengthening OMQ evaluation in
$(\ELIU_\bot,\text{UCQ})$ is only $\ExpTime$-complete.
\ALCI-ontologies can be rewritten in polynomial time into a
`nesting-free' normal form that is often used by reasoners and that
has sometimes been presupposed for approximation
\cite{DBLP:journals/jair/ZhouGNKH15}. The rewriting is not equivalence
preserving, but only yields a conservative extension. \ALCI-ontologies
in this form can in turn be rewritten into an equivalent
$\ELIU_\bot$-ontology without disjunction on the left.  The following
example shows that \ALC-ontologies that are not in normal form behave
differently in that they may have only infinite exhaustive
$\ELI_\bot$-approximation sets.
\begin{exmp}
  \label{ex:infiniteapproxsets}
  Let $\Omc = \{\exists r.\top \sqcap \forall r . A \sqsubseteq B_1 \sqcup B_2 \}$. Then for
  each $n \geq 1$, the $\ELI_\bot$-ontology
  $$
  \Omc_n = \{ \exists r.A \sqsubseteq \exists r^n . X, \ \exists r . (A
  \sqcap \exists r^{n-1} .X) \sqsubseteq B_1\}
  $$
  is such that $\Omc_n \models \Omc$. It is easy to see that
  $\Omc_n \not \models \Omc_m$ when $n \neq m$ and any $\ELI_\bot$-ontology $\Omc'_n$
  with $\Omc_n \models \Omc'_n \models \Omc$ is equivalent to $\Omc_{n}$.
\end{exmp}
In the appendix, we give another example which shows that
the effect pointed out in Example~\ref{ex:infiniteapproxsets} also
affects answers to OMQs. We leave the decidability and complexity of
$\ELI_\bot$-ontology relaxing approximation in $(\ALCI,\text{UCQ})$
(without assuming normal form) as an open problem.

We next turn to database strengthening approximation. 
For
an OMQ $Q(\bar x)=(\Omc,\Sigma,q)$, a
$\Sigma$-database \Dmc, and a class  \Dmf of pointed databases,
we use $\mn{app}^\uparrow_\Dmf(Q,\Dmc)$ to denote the
set of tuples $\bar a \in \mn{adom}(\Dmc)^{|\bar x|}$ such that for all
$\langle \Dmc', \bar b \rangle \in \Dmf$
and all homomorphisms $h$ from $\Dmc$ to $\Dmc'$ with $h(\bar a) = \bar b$, $\bar b \in Q(\Dmc')$.
Every choice of $(\Lmc, \Qmc)$ and $\Dmf$ gives rise to an approximate
OMQ evaluation problem.

\begin{center}
	\fbox{\begin{tabular}{@{}l@{\;}l}
			\multicolumn{2}{@{}l}{$\Dmf$-database
                strengthening evaluation in $(\Lmc,\Qmc)$}
			\\[.5mm]{\small INPUT} : & OMQ $Q(\bar x) =
                                              (\Omc,\Sigma,q)
                                             \in (\Lmc,\Qmc)$,\\
&			$\Sigma$-database $\Dmc$,  tuple $\bar a \in \mn{adom}(\Dmc)^{|\bar x|}$
			\\[.5mm]
                {\small OUTPUT} : &  `yes' if
                                    $\bar a \in \mn{app}^\uparrow_{\Dmf}(Q,\Dmc)$
                                            and `no' otherwise
              \end{tabular}}
\vspace*{.1mm}            
\end{center}
%
%
%
A natural choice for $\Dmf$ are classes of databases of
bounded treewidth. 
We only consider here the class $\Dmf_{1}$ of pointed databases
$\langle \Dmc, \bar a \rangle$ where $\mn{adom}(\Dmc) \setminus \bar a$
has treewidth $(1,2)$, i.e.\ it is a tree with multi-edge and self-loops admitted.
%
\begin{exmp}
  \label{ex:unsoundone}
  Take $Q(x)=(\Omc,\Sigma,q)$ where
  $$
  \begin{array}{r@{\;}c@{\;}l}
    \Omc &=& \{ A_i \sqcap A_j \sqsubseteq B \mid 1 \leq i < j \leq 3
             \} \\[0.5mm]
    \Sigma&=&\{A_1,A_2,A_3,B,r\}\\[0.5mm]
    q&=&\exists x \, B(x).
  \end{array}
  $$
  Let
\vspace*{-1mm}
  $$
  \begin{array}{r@{\;}c@{\;}l}
    \Dmc &=& \{ r(a_1,a_2), r(a_2, a_3), r(a_3,a_1),\\[0.5mm]
         && \phantom{\{} A_1(a_1), A_2(a_2), A_3(a_3)\}.
  \end{array}
  $$
  Then $() \in \mn{app}^\uparrow_{\Dmf_1}(\Dmc)$, but
  $() \notin Q(\Dmc)$.
\end{exmp}
Examples~\ref{ex:unsoundtwo} and~\ref{ex:unsoundone} also show that
$\Dmc_1$-database strengthening approximation and $\ELI_\bot$-ontology
strengthening approximation are incomparable as
$a \notin \mn{app}^\uparrow_{\Dmc_1}(\Dmc)$ in
Example~\ref{ex:unsoundtwo} and
$() \notin \mn{app}^\uparrow_{\ELI_\bot}(\Dmc)$ in
Example~\ref{ex:unsoundone}. It turns out that $\Dmf_1$-database
strengthening approximation does not even enjoy \PTime data
complexity, and this holds already when the original ontology is
formulated in a tractable language.
\begin{theorem}
  \label{thm:dbfromabovesucks}
  $\Dmf_1$-database strengthening approximation is
  $\coNP$-complete in data complexity in $(\ALCI,\text{UCQ})$.  The
  lower bound already holds when the ontology is empty. It also holds
  in $(\EL,\text{CQ})$.
\end{theorem}
%
The interesting part is the lower bound, proved by a non-trivial
reduction from the validity of propositional formulas.

\section{Conclusion}

We have introduced and studied various kinds of OMQ approximations.
Our results on approximation from below are summarized in
Figure~\ref{fig:results} where all entries for combined complexity
mean completeness results, unless stated otherwise.
We believe that the results in this paper show that ontology relaxing
approximation is preferable to database-relaxing approximation in most
aspects.
 First, they are conservative in the sense that they are
complete when the ontology is empty. Second, they have lower combined
complexity. And third, also for the dual notion of
ontology-strengthening
approximation one attains
\PTime data complexity. 
It would be interesting to generalize our approach to extensions of
\ALCI with, for instance, role hierarchies, transitive roles, and
functional roles. 
Moreover, ontology strengthening approximation deserves further study.

\medskip
\noindent

\section*{Acknowledgements}
Haga and Lutz were supported by DFG CRC 1320 Ease.
Wolter was supported by EPSRC grant EP/S032207/1.

\cleardoublepage

\bibliographystyle{kr}
\bibliography{refs}

\cleardoublepage

\appendix

\section{Further Preliminaries}

\subsection{The Chase}

Let $\vartheta = \phi(\bar x,\bar y) \rightarrow \exists \bar z \, \psi(\bar x,\bar z)$ be a TGD and $\Dmc$ a database. If $h$ is a homomorphism from 
$\phi(\bar x,\bar y)$ to $\Dmc$, then we say that $\vartheta$ is \emph{applicable to $\Dmc$ via $h$}. 
Now assume that $\vartheta$ is applicable to $\Dmc$ via $h$. Let
$h'(\psi(\bar x,\bar z))$ be obtained from $\psi(\bar x,\bar z)$ by
replacing the variables $x$ in $\bar x$ by $h(x)$ and replacing the variables $z$ in $\bar z$ with fresh constants. Then \emph{the result of applying $\vartheta$ to $\Dmc$ via $h$} is defined as the database $\Dmc'= \Dmc \cup h'(\psi(\bar x,\bar z))$ and we write $\Dmc \rightarrow^{h,\vartheta} \Dmc'$.

Let $\Dmc$ be a database and $\Omc$ be a set of TGDs, both potentially
infinite, but at most countable. A \emph{chase sequence of $\Dmc$
  w.r.t.~$\Omc$} is a potentially infinite sequence
$\Dmc_{0},\Dmc_{1},\ldots $ such that $\Dmc_{0}=\Dmc$ and for every
$i > 0$, there are a homomorphism $h_{i}$ and a TGD
$\vartheta_{i}\in \Omc$ with
$\Dmc_{i-1} \rightarrow^{h_{i},\vartheta_{i}} \Dmc_{i}$. The chase
sequence $\Dmc_{0},\Dmc_{1},\ldots $ is \emph{fair} if for every
$i\geq 0$ and $\vartheta \in \Omc$ that is applicable to $\Dmc_{i}$
via some homomorphism $h$, there is a $j> i$ such that $\Dmc_{j+1}$ is
the result of applying $\vartheta$ to $\Dmc_{j}$ via $h$.  We then
denote by $\text{ch}_{\Omc}(\Dmc)$ the union of all $\Dmc_{i}$ for
some fair chase sequence of $\Dmc$ w.r.t.~$\Omc$ and call it the
\emph{result of chasing $\Dmc$ with $\Omc$}. Note that our chase is
oblivious, that is, it may apply a TGD via a homomorphism $h$ despite
the fact that $h$ can be extended to a homomorphism from the head to
the database. As a consequence, the result of the chase is unique up
to
isomorphism.

\begin{lemma}
\label{lem:unimod1}
Let $\Omc$ be a (potentially infinite) TGD-ontology and $\Dmc$ a
(potentially infinite) $\Sigma$-database that is satisfiable w.r.t.\ \Omc. Then the following holds:
\begin{enumerate}
	\item $\text{ch}_{\Omc}(\Dmc)$ is a model of $\Dmc$ and $\Omc$;
	\item for every model $\Imc$ of $\Dmc$ and $\Omc$ there exists a homomomorphism $h$ from $\text{ch}_{\Omc}(\Dmc)$ to $\Imc$ with $h(a)=a$ for all $a\in
	 \mn{adom}(\Dmc)$;
	\item for $Q=(\Omc,\Sigma,q)$ with $q$ a UCQ and all $\bar a \in \mn{adom}(\Dmc)^{|\bar x|}$, $\bar a \in Q(\Dmc)$ iff $\bar a
	\in q(\mn{ch}_\Omc(\Dmc))$.   
\end{enumerate}
\end{lemma}
\noindent
\begin{proof} \
	We provide a sketch only.
	Condition~1 holds by definition of $\text{ch}_{\Omc}(\Dmc)$.
	For Condition 2, let the chase sequence that constructs $\text{ch}_{\Omc}(\Dmc)$ be $\Dmc_{0},\Dmc_{1},\ldots$ and let $\Imc$ be a model of $\Dmc$ and $\Omc$.  One can construct homomorphisms $h_{i}$ from $\Dmc_{i}$ to $\Imc$, for all $i \geq 0$, by induction on $i$ and starting with the identity mapping $h_{0}$ from $\Dmc$ to $\Imc$ in a straightforward way. Note that it does not play a role here that $\Omc$
	might be infinite. The only relevant point is that we use countable sets of symbols, which guarantees that $\Omc$ is countable as well and thus a fair chase sequence always exists.
	Condition~3 is a consequence of 1 and 2.
\end{proof}
Since every $\ELI_\bot$-ontology can be viewed as a TGD-ontology (see
Section~\ref{sect:prelim}), we may also apply the chase directly to 
$\ELI_\bot$-ontologies.

\subsection{Unraveling Databases}

We define the unraveling of a database \Dmc into a database of bounded
treewidth.  Let $1\leq \ell <k$, and $S\subseteq \mn{adom}(\Dmc)$. The
\emph{$\ell,k$-unraveling $\Dmc^\approx_{S,\ell,k}$ of \Dmc up to $S$}
is defined as follows.  An $\ell,k$-\emph{sequence} takes the form
$$v=S_0,O_{0},S_1,O_{1},S_2,\dots,O_{n-1},S_n,$$
$n \geq 0$, where $S \subseteq S_i \subseteq \mn{adom}(\Dmc)$,
$O_i \subseteq\mn{adom}(\Dmc)$,
and the following conditions are
satisfied:
%
\begin{itemize}
  
\item 
$|S_i \setminus S| \leq k$ for $0 \leq i \leq n$;

\item $S\subseteq O_{i} \subseteq S_i \cap S_{i+1}$ for $0 \leq i < n$; 

\item $|O_i \setminus S| \leq \ell$ for $0\leq i < n$.
\end{itemize}
For every
$a \in \mn{adom}(\Dmc)$, reserve a countably infinite set of fresh
constants that we refer to as \emph{copies} of $a$. For brevity, we
also consider $a$ to be a copy of itself. A \emph{bag} is a finite set
of copies of constants in $\mn{adom}(\Dmc)$.

Now let $(V,E)$ be the infinite directed tree with $V$ the set of all
$\ell,k$-sequences and $E$ the prefix order on $V$. We 
proceed inductively on $n$ to choose, for every
$v=S_{0}\cdots S_{n}\in V$, a bag $B_{v}$ that contains a copy
$a_{v}$ of every $a\in S_{n}$ and no other constants such that

\begin{enumerate}

\item if $a \in S$, then the copy $a_v$ of $a$ is $a$ itself;
  
\item if $n> 0$ and $a \in O_{n-1}S$, then the copy $a_{v}$ of $a$ is identical to the copy $a_{S_{0}\cdots S_{n-1}}$ of~$a$;
  
\item if $n=0$ and $a\notin S$ or $n> 0$ and $a\notin O_{n-1}$, then the copy $a_v$ of $a$ is fresh.
\end{enumerate}

We then define $\Dmc^{\approx}_{S,\ell,k}$ as the (unique and infinite)
database with active domain
$\bigcup_{v\in V}B_{v}$ such that 
\begin{enumerate}

\item $(V,E,(B_{v})_{v\in V})$ is
a tree decomposition of $\Dmc^{\approx}_{S,\ell,k}$ and

\item 
the `uncopying' map $a_{v}\mapsto a$ is an isomorphism from
$(\Dmc^{\approx}_{S,\ell,k})_{|B_{v}}$ onto $\Dmc_{|S_{n}}$, for every
$v=S_{0}\cdots S_{n}\in V$. 

\end{enumerate}
It is easy to see that the `uncopying'
map is a
homomorphism from $\Dmc^{\approx}_{S,\ell,k}$ onto $\Dmc$.  If
$S=\emptyset$, then we drop $S$ and simply speak of the
\emph{$\ell,k$-unraveling $\Dmc^\approx_{\ell,k}$ of \Dmc}.

\medskip

We next give some properties of unraveled databases that are
fundamental to the remainder of this paper. We first observe
that homomorphisms from databases of treewidth $(\ell,k)$ 
into a database $\Dmc$
can be lifted to homomorphisms into the $\ell,k$-unraveling of $\Dmc$.  
\begin{lemma}\label{lem:homintree}
	Let $(\Dmc, \bar a)$ be a pointed database and $1 \leq \ell < k$.
	
        \begin{enumerate}
        \item Let $(\Dmc',\bar b)$ be a pointed database of treewidth
          $(\ell,k)$. If there is a homomorphism $h$ from
          $\Dmc'$ to $\Dmc$ with $h(\bar b)=\bar a$, then there is a
          homomorphism $g$ from $\Dmc'$ to $\Dmc_{\bar
            a,\ell,k}^{\approx}$ with $g(\bar b)=\bar a$.

    \item Let $\Dmc'$ be a database of treewidth 
    $(\ell,k)$. If there is a homomorphism $h$ from $\Dmc'$ to $\Dmc$
    with $h(b)=a$, then for any copy $a'$ of $a$ in $\Dmc_{\bar
      a,\ell,k}^{\approx}$ there is a homomorphism $g$ from $\Dmc'$ to
    $\Dmc_{\bar a,\ell,k}^{\approx}$ with $g(b)=a'$.
        \end{enumerate}
  \end{lemma}
  \noindent
\begin{proof} \
(1) Assume that $(\Dmc',\bar b)$ is a pointed database of treewidth $(\ell,k)$.
Then the restriction of $\Dmc'$ 
to $\mn{adom}(\Dmc')\setminus \bar b$ has treewidth 
$(\ell,k)$. Assume there is a homomorphism
$h$ from $\Dmc'$ to \Dmc such that $h(\bar b)=\bar a$. To define $g$, let $(V,E,(B_{v})_{v\in V})$ be an $(\ell,k)$-tree decomposition of the restriction of $\Dmc'$ 
to $\mn{adom}(\Dmc')\setminus \bar b$. We may assume that $(V,E)$ is connected
and start defining $g$ by setting $g(\bar b)= h(\bar b)=\bar a$.
Now pick a first $v\in V$. We aim to define $g(b)$ for $b\in B_{v}$.
Let $(V',E')$ be the underlying tree of the tree decomposition of
$\Dmc_{\bar a,\ell,k}^{\approx}$ introduced in the definition of $\Dmc_{\bar a,\ell,k}^{\approx}$ and recall that $V'$ consists of $\ell,k$-sequences. To guide the construction, we also define a homomorphism $g'$ from $(V,E)$ and $(V',E')$ and start by setting $g'(v) = h(B_{v})\cup \bar a$. Define $g(b)$ as the copy of $h(b)$ in $B_{g'(v)}$, for all $b\in B_{v}$. Next assume that $g'$ has been defined on $U\subseteq V$ and $g$ on $\bigcup_{v\in U}B_{v}$ with $U$ connected and that $(v,v')\in E$ with $v\in U$ but $v'\not\in U$. Then $g'(v)$ is an $\ell,k$-sequence and we
can expand $g'$ by setting $g'(v')= g'(v)(h(B_{v} \cap B_{v'})\cup \bar a)(h(B_{v'})\cup \bar a)$. The definition of $g$ on $B_{v'}$ is as expected by setting $g(b)=h(b)'$ for the copy $h(b)'$ of $h(b)$ in $B_{g'(v')}$. Then $g$ restricted to $B_{v'}$ is clearly a homomorphism. Overall, it follows that the restriction   
of $g$ to $\mn{adom}(\Dmc')\setminus \bar b$ is a homomorphism to $\Dmc_{\bar a,\ell,k}^\approx$. That $g$ is a homomorphism from $\Dmc'$ to $\Dmc_{\bar a,\ell,k}^\approx$ is now a consequence of the fact that $h$ is homomorphism
and Point~2 of the definition of $\Dmc_{\bar a,\ell,k}^\approx$. 	

(2) The proof is similar to the proof Claim~(1). In this case we have to start
the definition of $g$ and $g'$ from some fixed $g(b)=a'$ with $a'$ a copy of $a=h(b)$. We use the notation from (1) and assume that $b\in B_{v}$. Consider
a shortest $\ell,k$-sequence $w=S_{0}\cdots S_{n}$ such that $a'\in B_{w}$.
Then we set $g'(v)= wO_{n}(h(B_{v})\cup \bar a)$, where $O_{n}= \{a'\} \cup \bar a$. Define $g(b')$ as the copy of $h(b')$ in $B_{g'(v)}$, for all $b'\in B_{v}$.
Observe that then $g(b)=a'$, as required. The function $g$ can now be extended to
a homomorphism from $\Dmc'$ to $\Dmc_{\bar a,\ell,k}^\approx$ in exactly the same way as above.
\end{proof}
The subsequent two lemmas are related to the notion of unraveling
tolerance considered in \cite{DBLP:journals/lmcs/LutzW17} where it is
observed that when a database \Dmc is unraveled into a proper tree
$\Dmc_{\emptyset}^\approx$ and $a\in \mn{adom}(\Dmc)$, then for any
OMQ $Q$ from $(\ELI_\bot,\text{ELIQ})$, $a$ is an answer to $Q$ on
\Dmc iff it is an answer to $Q$ on $\Dmc_{\emptyset}^\approx$, see
Lemma~\ref{lem:unravtol} below. In our case, ontologies are sets of TGDs,
and unraveling only provides databases of bounded treewidth. This makes
the formulation more subtle as we have to make sure that all
parameters fit together.
\begin{lemma}\label{lem:untol}
  Let $\Dmc$ be a database, $S \subseteq \mn{adom}(\Dmc)$, and $\Omc$
  a possibly infinite $\ell,k,\ell',k'$-ontology, $1\leq\ell< k$ and
  $1\leq\ell'< k'$. Then $\Dmc$ is satisfiable w.r.t.~$\Omc$ iff
  $\Dmc_{S,\ell,k}^{\approx}$ is satisfiable w.r.t.~$\Omc$.
  Moreover, if $\Dmc$ is satisfiable w.r.t.~$\Omc$ and $q(\bar x)$ is
  a Boolean or unary CQ of treewidth $(\ell',k')$ and
  $\bar a\in \mn{adom}(\Dmc)^{|\bar x|}$, then for all copies
  $\bar a'$ of~$\bar a$ in $\Dmc_{S,\ell,k}^{\approx}$:
  \begin{enumerate}
        \item if $\Dmc^{\approx}_{S,\ell,k}, 
              \Omc\models q(\bar a')$,
              then $\Dmc,\Omc\models q(\bar a)$;
		\item if $\Dmc,\Omc\models q(\bar a)$, $\ell'\leq \ell$, and $k'\leq k$,
                  then $\Dmc^{\approx}_{S, \ell,k},\Omc\models
                  q(\bar a')$;
		\item if $\Dmc,\Omc\models q(\bar a)$ is witnessed by
                  a homomorphism
                  $h$ from $q(\bar x)$ to $\text{ch}_{\Omc}(\Dmc)$ that
                  maps every variable distinct from $\bar x$ to
                  $\mn{adom}(\text{ch}_{\Omc}(\Dmc))\setminus
                  \text{adom}(\Dmc)$, then
                  $\Dmc^{\approx}_{S,\ell,k},\Omc\models q(\bar a')$.
	\end{enumerate}
      \end{lemma}
      \noindent
      \begin{proof} \ We first consider Points 1., 2., and 3., and
        then return to satisfiability.  Point~1 follows from the
        observation made above that there is a homomorphism from
        $\Dmc^{\approx}_{S,\ell,k}$ to $\Dmc$ that maps $\bar a'$ to
        $\bar a$.  For Points~2. and~3., we first observe the
        following. Let
        $\vartheta = \phi(\bar x,\bar y) \rightarrow \exists \bar z \,
        \psi(\bar x,\bar z)$ be any $\ell,k,\ell',k'$-TGD in $\Omc$ and let $b\in  \mn{adom}(\Dmc)^{|\bar x|}$. Then, by
        Lemma~\ref{lem:homintree} (2), there is a homomorphism from
        the body $\phi(\bar x,\bar y)$ of $\vartheta$ to $\Dmc$ that
        maps $\bar x$ to $\bar b$ iff there is a homomorphism from
        $\phi(\bar x,\bar y)$ to $\Dmc^{\approx}_{S,\ell,k}$ that maps
        $\bar x$ to $\bar b'$ for every (equivalently: any) copy
        $\bar b'$ of $\bar b$. In terms of the construction of chase sequences it follows that $\vartheta$ is applicable to $\Dmc$ via a homomorphism that maps $\bar x$ to $\bar b$ iff it is applicable 
        to $\Dmc^{\approx}_{S,\ell,k}$ via a homomorphism that maps $\bar x$ to $\bar b'$ for every (equivalently: any) copy $\bar b'$ of $\bar b$.
        It can thus be proved by induction over the construction of chase sequences that the following holds for the universal models
        $\text{ch}_{\Omc}(\Dmc)$ and
        $\text{ch}_{\Omc}(\Dmc^{\approx}_{S, \ell,k})$: for
        every $b\in \text{adom}(\Dmc)$ and any copy $b'$ of $b$ in
        $\Dmc^{\approx}_{S, \ell,k}$, the interpretation
        $\Imc_{b}$ attached in $\text{ch}_{\Omc}(\Dmc)$ to $\Dmc$ at
        $b$ is isomorphic to the interpretation
        $\Imc_{b'}$ attached in
        $\text{ch}_{\Omc}(\Dmc^{\approx}_{S,\ell,k})$ to
        $\Dmc^{\approx}_{S,\ell,k}$ at $b'$. Observe that the interpretations
        $\Imc_{b}, \Imc_{b'}$ have treewidth 
        $(\ell',k')$. Claim~3 follows directly from this
        observation. For Claim~2, use Lemma~\ref{lem:homintree} (2).

        It remains to show that $\Dmc$ is satisfiable w.r.t.~$\Omc$
        iff $\Dmc_{S,\ell,k}^{\approx}$ is satisfiable
        w.r.t.~$\Omc$. The direction from left to right follows from
        the existence of a homomorphism from
        $\Dmc_{S,\ell,k}^{\approx}$ to~$\Dmc$.  Conversely,
        assume that $\Dmc_{S,\ell,k}^{\approx}$ is satisfiable
        w.r.t.~$\Omc$. Obtain an ontology $\Omc'$ from $\Omc$ by
        replacing all occurrences of $\bot$ in $\Omc$ by
        $\exists x A(x)$, for a single fresh concept name $A$. Then
        $\Dmc$ is satisfiable w.r.t.~$\Omc'$ (as any database is
        satisfiable w.r.t.~$\Omc'$) and $\Dmc$ is satisfiable
        w.r.t.~$\Omc$ iff $\Dmc,\Omc'\not\models \exists x A(x)$. By
        Point~2 above, the latter is equivalent to
        $\Dmc_{S,\ell,k}^{\approx},\Omc'\not\models \exists x
        A(x)$ which holds since $\Dmc_{S,\ell,k}^{\approx}$ is
        satisfiable w.r.t~$\Omc$.
\end{proof}
Lemma~\ref{lem:untol} also justifies the notation
$\langle \Dmc^\approx_{\ell,k},a \rangle$ denoting ``the'' database
obtained from $\Dmc^\approx_{\ell,k}$ by choosing a copy of
$a$ in $\Dmc^\approx_{\ell,k}$ and renaming it back to $a$. In fact, one can now easily show that $\Dmc^\approx_{\ell,k},\Omc \models q(a')$ iff $\Dmc^\approx_{\ell,k},\Omc \models q(a'')$ for any possibly infinite set $\Omc$ 
of TGDs, CQ $q(x)$, and any two copies $a'$ and $a''$ of a constant $a$ in $\Dmc$.
\begin{lemma}\label{lem:untol2} \
  Let $Q(\bar x) = (\Omc,\Sigma,q)$ be an OMQ with $\Omc$ a
  possibly infinite $\ell,k,\ell',k'$-ontology, $1\leq \ell< k$
  and $1 \leq \ell' < k'$, and $q$ a CQ of treewidth $(\ell,k)$. Then
  for all $\Sigma$-databases $\Dmc$ and all $\bar a \in
  \mn{adom}(\Dmc)^{|\bar x|}$, $\bar a\in Q(\Dmc)$ iff $\bar a \in
  Q(\Dmc^{\approx}_{\bar a,\ell,k})$.
      \end{lemma}
      \noindent
\begin{proof} \
By Lemma~\ref{lem:untol}, $\Dmc$ is satisfiable w.r.t.~$\Omc$ iff $\Dmc^{\approx}_{\bar a,\ell,k}$ is satisfiable w.r.t.~$\Omc$. Thus, we may assume
that both $\Dmc$ and $\Dmc^{\approx}_{\bar a,\ell,k}$ are satisfiable w.r.t.~$\Omc$.
If $\bar a\in Q(\Dmc^{\approx}_{\bar a,\ell,k})$,
then $\bar a\in Q(\Dmc)$ follows from the existence of a homomorphism from $\Dmc^{\approx}_{\bar a,\ell,k}$ to $\Dmc$ mapping $\bar a$ to $\bar a$. For the converse direction, let $\bar a \in Q(\Dmc)$. Let $h$ be a homomorphism from $q$ to $\text{ch}_{\Omc}(\Dmc)$ mapping $\bar x$ to $\bar a$. Consider the restriction of $h$ to the set of variables $V$ in $q(x)$ that are mapped into $\mn{adom}(\Dmc)$. Then, by Lemma~\ref{lem:homintree} (1), we find a homomorphism $g$ from $V$ to the restriction of $\text{ch}_{\Omc}(\Dmc^{\approx}_{\bar a,\ell,k})$ to $\mn{adom}(\Dmc^{\approx}_{\bar a,\ell,k})$ mapping $\bar x$ to $\bar a$ and such that $g(y)$ is a copy of $h(y)$, for all variables $y\in V$ (the latter condition is a consequence of the proof of Lemma~\ref{lem:homintree} (1)). But then
the remaining variables of $q$ that are not in $V$ can be mapped to the interpretations $\Imc_{g(y)}$ attached to $g(y)$ in $\text{ch}_{\Omc}(\Dmc^{\approx}_{\bar a,\ell,k})$ in the same way they can be mapped to the interpretations $\Imc_{h(y)}$ attached to $h(y)$ in $\text{ch}_{\Omc}(\Dmc)$. The resulting homomorphism shows that $\bar a\in Q(\Dmc^{\approx}_{\bar a,\ell,k})$.
\end{proof}

\section{Proofs for Section~\ref{sect:approxdef}}

\lemunravfund*

\noindent
\begin{proof}\ For the ``if'' direction, observe that for every finite
  subset $\Dmc'$ of $\Dmc^\approx_{\bar a}$ containing $\bar a$ the
  pointed database $(\Dmc',\bar a)$ is a member of
  $\Dmf_{\!\curlywedge}$. Moreover, there is a homomorphism from
  $\Dmc'$ to $\Dmc$ mapping $\bar a$ to $\bar a$. Thus, this direction
  is a consequence of compactness and the preservation of OMQ answers
  under homomorphic images of databases
  \cite{DBLP:journals/tods/BienvenuCLW14}. For the ``only if''
  direction, assume that there is a pointed tree database $(\Dmc',\bar
  b)$ with $\bar b \in Q(\Dmc')$, and a homomorphism $h$ from $\Dmc'$
  to \Dmc such that $h(\bar b)=\bar a$. To show that $\bar a \in
  Q(\Dmc_{\bar a}^\approx)$, it suffices to construct a homomorphism
  $g$ from $\Dmc'$ to $\Dmc_{\bar a}^\approx$ with $g(\bar b)=\bar a$
  and to again use the preservation of OMQ answers under homomorphic
  images of databases. To define $g$, start with setting $g(\bar b):=
  h(\bar b) = \bar a$ and choose for any maximally connected component
  of the restriction of $\Dmc'$ to $\mn{adom}(\Dmc') \setminus \bar b$
  an arbitrary constant $c$ in that component and set $g(c)=
  h(c)$. Next, assume that $r(c_1,c_2) \in \Dmc'$ with
  $c_{1},c_{2}\not\in \bar b$, $r$ a potentially inverse role, and
  $g(c_1)$ is already defined, but $g(c_2)$ is not. Then
  $g(c_1)rh(c_2)$ is a path in \Dmc and we set
  $g(c_2)=g(c_1)rh(c_2)$. The restriction of $g$ to $\mn{adom}(\Dmc')
  \setminus \bar b$ is a homomorphism by definition of $\Dmc_{\bar
    a}^\approx$ and since, after removal of $\bar b$, $\Dmc'$ does not
  contain reflexive loops or multi-edges. To show that $g$ is a
  homomorphism on $\Dmc'$ assume that $r(c_{1},c_{2}) \in \Dmc'$ with
  $c_{1}\in \bar b$ and $c_{2}\not\in \bar b$, $r$ a potentially
  inverse role. Then $\mn{tail}(g(c_{2}))= h(c_{2})$ and so
  $(g(c_{1}),g(c_{2}))\in \Dmc^\approx_{\bar a}$ since
  $r(h(c_{1}),h(c_{2})) \in \Dmc$, as required.
\end{proof}

\section{Proofs for Section~\ref{sect:eliu}}
\label{app:eliu}

We begin by providing proofs that the examples given in this section are correct. 

\medskip
\noindent {\bf Example~\ref{exm:xx}}. 
Recall that in this example, $Q(x)=(\Omc,\Sigma,q) \in (\ALC,\text{CQ})$ with $\Omc = \{ \top
	\sqsubseteq \forall r . (B_1 \rightarrow B) \sqcup \forall r . (B_2
	\rightarrow B) \}$, $\Sigma = \{r,A,B,B_1,B_2\}$, and $q=\exists y \,
	r(x,y) \wedge A(y) \wedge B(y)$.  Also,
	$$\Dmc = \{
	r(a,b_1), r(a,b_2), B_1(b_1), B_2(b_2),A(b_1),A(b_2) \}.
	$$
        Let $Q^-_{\ELI^u_\bot} =(\Omc^-_{\ELI^u_\bot},\Sigma,q)$ for
        $\Omc^-_{\ELI^u_\bot}$ the result of removing from
        $\Omc^\approx_{\ELI^u_\bot}$ all CIs that use a symbol that
        does not occur in \Omc.
	We claim that $a \notin Q^-_{\ELI^u_\bot}(\Dmc)$
	where $Q^-_{\ELI^u_\bot} =(\Omc^-_{\ELI^u_\bot},\Sigma,q)$ for $\Omc^-_{\ELI^u_\bot}$
	the result of removing from
	$\Omc^\approx_{\ELI^u_\bot}$ all CIs that use a symbol that does not
	occur in \Omc. To see this consider the universal model 
		$\mn{ch}_{\Omc^-_{\ELI^u_\bot}}(\Dmc)$. Then $A(c)\in \mn{ch}_{\Omc^-_{\ELI^u_\bot}}(\Dmc)$ iff $c\in \{b_{1},b_{2}\}$, but 
		neither $B(b_{1})\in \mn{ch}_{\Omc^-_{\ELI^u_\bot}}(\Dmc)$ nor $B(b_{2}) \in \mn{ch}_{\Omc^-_{\ELI^u_\bot}}(\Dmc)$ from which the claim follows.
		Observe that the latter holds also in $\mn{ch}_{\Omc^{\approx}_{\ELI^u_\bot}}(\Dmc)$, but in $\mn{ch}_{\Omc^{\approx}_{\ELI^u_\bot}}(\Dmc)$ there is an additional $r$-successor $c$ of $a$ with $A(c),B(c) \in \mn{ch}_{\Omc^{\approx}_{\ELI^u_\bot}}(\Dmc)$.  

\medskip
\noindent
{\bf Example~\ref{exm:yy}}. Recall that in this example
$Q=(\Omc,\Sigma,q) \in (\ALC,\text{CQ})$ is the Boolean OMQ with
	$\Omc = \{ A \sqsubseteq B \sqcup \forall r . B \}$, $\Sigma=
	\{r,A,B\}$, and $q = \exists x B(x)$. Assume $\Dmc= \{A(a),r(a,b)\}$.
    We sketch the proof that $\Dmc \not\models Q^\approx_{\ELI_{\bot}}$.
    First observe that there does not exist any sequence of roles $\rho= r_{1},\ldots,r_{n}$ with $r_{i}\in \{r,r^{-}\}$ for all $1\leq i \leq n$ such that 
   $a\in Q'(\Dmc)$ for the query $Q'=(\Omc,\Sigma,q_{\rho})$, where
    $$
    q_{\rho}(x) = \exists x_{1}\cdots \exists x_{n} r_{1}(x,x_{1}) \wedge \cdots \wedge r_{n}(x_{n-1},x_{n}) \wedge B(x_{n}).
    $$
    Thus, we find models $\Imc_{\rho}$ of $\Dmc$ and $\Omc$ such that $\Imc_{\rho}\not\models q_{\rho}(a)$. Consider the product  $\Imc=\prod_{\rho}\Imc_{\rho}$. Then $\Imc$ is a model of $\Omc^{\approx}_{\ELI_\bot}$, as $\ELI_{\bot}$-ontologies are preserved under products \cite{DBLP:journals/tocl/HernichLPW20}, and $\Imc\not\models q_{\rho}(a)$ for any $\rho$ (we identify $a$ with the sequence $(a,a,\ldots)$ in $\Imc$).
   Consider the subinterpretation $\Imc'$ of $\Imc$ induced by all nodes in $\Imc$ that can be reached from $a$ by any path $\rho$, that is, any path using $r^{\Imc}$ and $(r^{-})^{\Imc}$. Then $\Imc'$ is a model of $\Dmc$ and $\Omc$ as $\Omc$ does not use the universal role and $\Imc'\not\models \exists x B(x)$. It follows that $\Dmc \not\models Q^\approx_{\ELI_{\bot}}$, as required.     



   \medskip
   
The following lemma is related to the notion of unraveling
tolerance, introduced in \cite{DBLP:journals/lmcs/LutzW17}. In the
language of that paper, Point~1 states that the DL $\ELI^u_\bot$ is
unraveling tolerant, even with infinite ontologies.

Note that for every database \Dmc, $\mn{adom}(\Dmc) \subseteq
\Dmc^\approx_\emptyset$ by definition of tree unravelings. In contrast
to unravelings into bounded treewidth, we thus need no `renaming back'
of those constants to their original name, c.f.\ the $\langle
\Dmc^\approx_{\ell,k},\bar a \rangle$ notation from
Section~\ref{sect:tgd}.
\begin{lemma}
	\label{lem:unravtol}  
	Let $Q(\bar x) = (\Omc,\Sigma,q) \in (\ELI^u_\bot,\text{bELIQ})$,
        \Dmc a $\Sigma$-database, and
        $a \in \mn{adom}(\Dmc)^{|\bar x|}$. Then
        \begin{enumerate}

        \item $\bar a \in Q(\Dmc)$ iff $\bar a \in Q(\Dmc^\approx_{\emptyset})$;

        \item \Dmc is satisfiable w.r.t.\ \Omc iff $\Dmc^\approx_{\emptyset}$ is
          satisfiable w.r.t.\ \Omc.
          
        \end{enumerate}
\end{lemma}
\noindent
\begin{proof}\ Point~1 is proved in \cite{DBLP:journals/lmcs/LutzW17}
  for OMQs based on finite ontologies. The result easily lifts to
  infinite
  ontologies through compactness. Point~2 is a consequence of
  Point~1. To see the latter, let $\Omc'$ be obtained from \Omc by
  replacing every CI $C \sqsubseteq \bot$ with $C \sqsubseteq A_\bot$,
  $A_\bot$ a fresh concept name. Then \Dmc is satisfiable w.r.t.\ \Omc
  iff $\Dmc,\Omc' \models \exists u . A_\bot$ iff
  $\Dmc^\approx_{\emptyset},\Omc' \models \exists u . A_\bot$ (by
  Point~1) iff $\Dmc^\approx_{\emptyset}$ is satisfiable w.r.t.~\Omc.
\end{proof}
The next lemma identifies a crucial property that relates
$\ELI^u_\bot$-ontology relaxing approximation to tree unravelings
of databases. 
\lemcrucial*
\noindent
\begin{proof}\ We start with Point~1.  First assume that
    $\bar a \in Q^\approx_{\ELI^u_\bot}(\Dmc)$. By
    Lemma~\ref{lem:unravtol}, $\bar a \in
    Q^\approx_{\ELI^u_\bot}(\Dmc^\approx_{\emptyset})$. From $\Omc \models
    \Omc^\approx_{\ELI^u_\bot}$, it follows that $\bar a \in
    Q(\Dmc^\approx_{\emptyset})$.
           
    Now assume that $\bar a \in Q(\Dmc^\approx_{\emptyset})$. Then
    compactness yields a finite subset $\Fmc$ of $\Dmc^\approx_{\emptyset}$ with
    $\bar a \in Q(\Fmc)$.  We can view \Fmc as an $\ELI$-concept $F$.
    Then $\Omc \models F \sqsubseteq A$ and thus $F \sqsubseteq A$ is
    a CI in $\Omc^\approx_{\ELI^u_\bot}$. If $q$ is an ELIQ and
    $\bar a = a_0$, then we further have
    $\Dmc^\approx_{\emptyset} \models F(a_0)$, thus
    $\Dmc, \Omc^\approx_{\ELI^u_\bot} \models F(a_0)$ by
    Lemma~\ref{lem:unravtol} which yields
    $\Dmc, \Omc^\approx_{\ELI^u_\bot} \models A(a_0)$ as required. If
    $q$ is a BELIQ $\exists u . C$, then there is an
    $a_0 \in \mn{ind}(\Fmc)$ with $\Fmc,\Omc \models C(a_0)$. Let $F'$
    be \Fmc viewed as a BELIQ with root $a_0$. Then
    $\Omc \models F' \sqsubseteq C$, thus $F' \sqsubseteq C$ is a CI in
    $\Omc^\approx_{\ELI^u_\bot}$. From $\Dmc^\approx_{\emptyset} \models F'(a_0)$,
    we
    obtain
    $\Dmc, \Omc^\approx_{\ELI^u_\bot} \models F'(a_0)$ by
    Lemma~\ref{lem:unravtol} which yields
    $\Dmc, \Omc^\approx_{\ELI^u_\bot} \models \exists u . C$ as required. 

    Now for Point~2. First assume that \Dmc is
  unsatisfiable w.r.t.\ $\Omc^\approx_{\ELI^u_\bot}$.  By
  Lemma~\ref{lem:unravtol}, $\Dmc^\approx_{\emptyset}$ is unsatisfiable w.r.t.\
  $\Omc^\approx_{\ELI^u_\bot}$. From $\Omc \models
  \Omc^\approx_{\ELI^u_\bot}$, it thus follows that $\Dmc^\approx_{\emptyset}$
  is unsatisfiable w.r.t.\ \Omc.

  Conversely, assume that \Dmc is satifiable w.r.t.\
  $\Omc^\approx_{\ELI^u_\bot}$.  By Lemma~\ref{lem:unravtol},
  $\Dmc^\approx_{\emptyset}$ is satisfiable w.r.t.\
  $\Omc^\approx_{\ELI^u_\bot}$. Further assume to the contrary
  of what is to be shown that $D^\approx_{\emptyset}$ is unsatisfiable w.r.t.\
  \Omc.
  Compactness yields a finite subset $\Fmc$ of
    $\Dmc^\approx_{\emptyset}$ that is unsatisfiable w.r.t.\ \Omc. We can view
    \Fmc as an $\ELI$-concept $F$. Then $\Omc \models F \sqsubseteq
    \bot$ and thus $F \sqsubseteq \bot$ is a CI in
    $\Omc^\approx_{\ELI^u_\bot}$, in contradiction of $\Dmc^\approx_{\emptyset}$
    being satisfiable w.r.t.\ $\Omc^\approx_{\ELI^u_\bot}$.
 \end{proof}

 We next prove Point~3 of Theorem~\ref{thm:mainone}. To prepare for
 the proof of Theorem~\ref{thm:treeunravmain}  in the subsequent
 section, we actually establish a stronger result.  By Point~1 of 
 Lemma~\ref{lem:crucial}, we can decide whether
 $\bar a \in Q(\Dmc^\approx_{\emptyset})$ in place of
 $\bar a \in \Omc^\approx_{\ELI^u_\bot}(\Dmc)$. Here, consider the
case $\bar a \in Q(\Dmc^\approx_{S})$ with
 $S \subseteq \mn{adom}(\Dmc)$ such that $S$ is given as an additional
 input with $|S| \leq |\bar x|$; we instantiate $S$ with $\emptyset$
 for the proof of Theorem~\ref{thm:mainone} and with $\bar a$ for the
 proof of Theorem~\ref{thm:treeunravmain}. Also, we consider ELIQs and
 \emph{disjunctions} of BELIQs, a class of UCQs that we denote with
 bELIQ$^\vee$.
\begin{theorem}
\label{thm:eliqs} 
Given $Q(\bar x)=(\Omc,\Sigma,q) \in (\ALCI,\text{bELIQ}^\vee)$,
a $\Sigma$-database \Dmc, $\bar a \in \mn{adom}(\Dmc)^{|x|}$,
and $S \subseteq \mn{adom}(\Dmc)$, 
it is decidable in time $2^{O(||Q||)} \cdot O(||\Dmc||)$
whether $\bar a \in 
 Q(\Dmc^\approx_{S})$.
\end{theorem}
We prove Theorem~\ref{thm:eliqs} by linear time reduction to the unsatisfiability
of propositional Horn formulas, which is in linear time
\cite{DBLP:journals/jlp/DowlingG84}. Let $Q(\bar x)=(\Omc,\Sigma,p)$, \Dmc,
$\bar a$, and $S$ be as in the theorem.

It is convenient to view $q$ as a concept, namely as an \ELI-concept
if $q$ is an ELIQ and as a disjunction of $\ELI^u$-concepts if $q$ is
a disjunction of BELIQs. We use $\Cl(\Omc,q)$ to denote the smallest
set that contains all concepts in \Omc and the concept $q$ and is
closed under subconcepts and single negation (thus $\Cl(\Omc,q)$
contains \ALCI-concepts and possibly negated $\ELI^u$-concepts).  A
\emph{type} for \Omc and $q$ is a maximal set
$t \subseteq \Cl(\Omc,q)$ such that $\bigsqcap t$ is satisfiable
w.r.t.~$\Omc$. We remark that this condition can be checked in time
$2^{O(||\Omc||)}$. In fact, standard algorithms for the satisfiability
of \ALCI-concepts w.r.t.\ \ALCI-ontologies such as type elimination
\cite{DBLP:books/daglib/0041477}
exhibit this running time.
We use $\TP(\Omc,q)$ to denote the set of all types
for~\Omc and~$q$. The type \emph{realized} at $d \in \Delta^\Imc$ in a
model \Imc of \Omc is
$$\mn{tp}_\Imc(d) := \{ C \in \TP(\Omc,q) \mid a \in C^\Imc \}.$$
For $t, t' \in \TP(\Omc,q)$ and roles $r$, we write
$t \rightsquigarrow_r t'$ if
\begin{description}
\item[($\mathsf{c1}$)] $D \in t'$ and $\exists r.D \in \Cl(\Omc,q)$ implies $\exists r.D \in t$ and
\item[($\mathsf{c2}$)]$D \in t$ and $\exists r^-.D \in \Cl(\Omc,q)$ implies $\exists r^-.D \in t'$.
\end{description}
For a set $S' \subseteq \mn{adom}(\Dmc)$, an \emph{$S'$-assignment} is
a function $\mu:S' \rightarrow \TP(\Omc,q)$ such that the
following conditions are satisfied for all $a,a_1,a_2 \in S'$:
\begin{description}

\item[($\mathsf{c3}$)] $A(a) \in \Dmc$ implies $A \in
  \mu(a)$;

\item[($\mathsf{c4}$)] $r(a_1,a_2) \in \Dmc$ implies
  $\mu(a_1) \rightsquigarrow_r \mu(a_2)$;

\item[($\mathsf{c5}$)] if $q$ is an ELIQ and $\bar a = a$, then $q
  \notin \mu(a)$;

\item[($\mathsf{c6}$)] if
  $q$ 
  is a
  disjunction of BELIQs,
  then $q \notin \mu(a)$.

\end{description}
%
We shall primarily be interested in $S$-assignments and in $S
\cup \{ a \}$-assignments for constants $a \in \mn{adom}(\Dmc)
\setminus S$. Note that there are at most
$2^{O(||Q||)}$ $S$-assignments and $S \cup
\{a\}$-assignments for each $a \in
\mn{adom}(\Dmc)$ and that we can compute the set of
$S$-assignments and  $S \cup
\{a\}$-assignments, $a \in
\mn{adom}(\Dmc)$, in time $2^{O(||Q||)} \cdot ||\Dmc||$.

We introduce a propositional variable $p_{\mu,a}$ for every
$a \in \mn{adom}(\Dmc) \setminus S$ and every
$S \cup \{a\}$-assignment~$\mu$, as well as a propositional variable
$p_{\mu}$ for every $S$-assignment $\mu$. Informally, $p_{\mu,a}$
being true means that it is not possible to simultaneously realize the
type $\mu(a)$ at $a$ (equivalently: at a path $p$ with
$\mn{tail}(p)=a$) and the type $\mu(b)$ at $b$ for every $b \in S$ in
a model of $\Dmc^\approx_S$ and~\Omc, and likewise for
propositional variables $p_\mu$. Now the propositional Horn formula
$\varphi$ is the conjunction of the following:
\begin{enumerate}

\item   $\bigwedge_{\mu \in W} p_{\mu,a} \rightarrow p_{\mu',b}$ for all
  $r(a,b) \in \Dmc$ with $r$ a potentially inverse role and $a,b
  \notin S$ and all
  $S \cup \{b\}$-assignments 
  $\mu'$, where $W$ consists of all $S \cup \{a\}$-assignments
  $\mu$ such that $\mu(a) \rightsquigarrow_r \mu'(b)$
  and $\mu(c)=\mu'(c)$ for all $c \in S$;

\item $\bigwedge_{\mu \in W} p_{\mu,a} \rightarrow p_{\mu'}$
  for all $a \in \mn{adom}(\Dmc) \setminus S$ and $S$-assignments 
  $\mu'$, where $W$ consists of all $S \cup \{a\}$-assignments
  $\mu$ such that $\mu(c)=\mu'(c)$ for all $c \in S$;

\item $p_{\mu} \rightarrow p_{\mu',a}$, for all
  $a \in \mn{adom}(\Dmc) \setminus S$, $S$-assignments $\mu$, and
  $S \cup \{a\}$-assignments $\mu'$ such that $\mu(c)=\mu'(c)$ for all
  $c \in S$;

  


\item $\bigwedge_{\mu \in W} p_{\mu,a} \rightarrow \bot$ for all $a
  \in \mn{adom}(\Dmc)\setminus S$ where $W$ is the set of all $S\cup \{a\}$-assignments.

\end{enumerate}
It is clear that $||\varphi|| \in 2^{O(||Q||)} \cdot ||\Dmc||$ and
that $\varphi$ can be constructed in
$O(2^{O(||Q||)} \cdot ||\Dmc||)$.  
Now, Theorem~\ref{thm:eliqs} is an immediate consequence of the
following.
\begin{lemma}
  \label{lem:horncorr}
  $\bar a \in Q(\Dmc^\approx_{S})$ iff $\varphi$ is unsatisfiable.
\end{lemma}
\noindent
\begin{proof}\
The (contrapositive) of the ``if'' direction 
is proved as follows. Assume that $\bar a \notin Q(\Dmc^\approx_{S})$.
Then there is a model \Imc of $\Dmc^\approx_{S}$ and \Omc with
$\bar a \notin q(\Imc)$. For each $a \in \mn{adom}(\Dmc) \setminus S$,
let $\mu_a$ denote the $S \cup \{a\}$-assignment that sets
$\mu_a(b)= \mn{tp}_\Imc(b)$ for all $b \in S \cup \{ a \}$.  Let
$\mu_S$ be the $S$-assignment defined in the same way.  It is readily
checked that the $\mu_a$ and $\mu_S$ are indeed assignments and in
particular, conditions ($\mathsf{c5}$) and~($\mathsf{c6}$) are satisfied since $\bar a \notin q(\Imc)$.

%
Define a valuation $V$ for the variables in $\varphi$ by setting,
\begin{itemize}

\item $V(p_{\mu,a})=0$ iff $\mu =\mu_a$,
  for all $a \in \mn{adom}(\Dmc) \setminus S$;

\item $V(p_\mu)=0$ iff $\mu=\mu_S$.

\end{itemize}
It is readily checked that $V$ is a model of $\varphi$. 
\smallskip

For the (contrapositive of the) ``if'' direction, assume that
$\varphi$ is satisfied by some valuation $V$. Choose some
$a_0 \in \mn{adom}(\Dmc) \setminus S$ and some $\mu_0$ with
$V(p_{\mu_{0},a_0})=0$, which exist due to the conjunct of $\vp$  in
Point~4. For every $a \in \mn{adom}(\Dmc) \setminus S$, let $T_a$ denote the set
of types from $\TP(\Omc,q)$ such that there is an
$S \cup \{a\}$-assignment $\mu$ such that $V(p_{\mu,a})=0$ and
$\mu(b)=\mu_{0}(b)$ for all $b \in S$. Due to the conjuncts in
Points~2 and~3, $T_a$ is non-empty for all $a \in \mn{adom}(\Dmc)$.

We construct a model $\Imc$ of $\Dmc^\approx_{S}$ and \Omc such that
$\mn{tp}_\Imc(a) \in T_a(a)$ for all $a \in \mn{adom}(\Dmc)$. By
definition of $T_a$ and due to Conditions ($\mathsf{c5}$)
and~($\mathsf{c6}$) of assignments, this implies
$\bar a \notin q(\Imc)$ and thus it follows that $\bar a \notin
Q(\Dmc^\approx_{S})$, as required.
We first assign to each $p \in \mn{adom}(\Dmc^\approx_{S})$ a type
$t_p \in T_{\mn{tail}(p)}$. Start with 
\begin{itemize}

\item setting $t_a = \mu_{0}(a)$ for all $a \in S$ and

\item and choosing $t_a \in T_a$ arbitrarily for all $a \in
  \mn{adom}(\Dmc) \setminus S$.

\end{itemize}
Now let $r(p_1,p_2) \in \Dmc^\approx_{S}$, $r$ a possibly inverse
role, and assume that $t_{p_1}$ is already defined, but $t_{p_2}$ is
not. By definition of $\Dmc^\approx_{S}$, $p_2=p_1 r a$ for some
(possibly inverse) role $r$ and some $a \in \mn{adom}(\Dmc)$.
Consequently, $r(\mn{tail}(p_1),\mn{tail}(p_2)) \in \Dmc$. We know
that $t_{p_1} \in T_{\mn{tail}(p_1)}$.
By the conjunct in
Point~1, there is thus a $t_{p_2} \in T_{\mn{tail}(p_2)}$ such that
$t_{p_1} \rightsquigarrow_r t_{p_2}$.

We now construct an interpretation $\Imc$ such that
$\mn{tp}_\Imc(p) = t_p$ for all $p \in
\mn{adom}(\Dmc^\approx_{S})$. Start
with setting
$$
\begin{array}{r@{\;}c@{\;}l}
  \Delta^\Imc &=& \mn{adom}(\Dmc^\approx_{S}) \\[1mm]
  A^{\Imc} &=& \{p \in \mn{adom}(\Dmc^\approx_{S}) \mid A \in t_p \} \\[1mm]
r^{\Imc} &=& \{(p_1,p_2) \mid r(p_1,p_2) \in
\Dmc^\approx_{S}\} 
\end{array}
$$
for all concept names $A$ and role names $r$. We then extend $\Imc$ by
adding, for all $p \in \mn{adom}(\Dmc^\approx_{S})$, a tree model
$\Imc_p$ of \Omc that satisfies type $t_p$ at its root, with disjoint
domain and identifying its root with~$p$. It can be proved by
induction on the structure of concepts $D$ that for all concepts
$D \in \mn{cl}(\Omc, q)$,
\begin{enumerate}
\item 
$D^\Imc \cap \mn{adom}(\Dmc^\approx_{S}) = \{p \in \mn{adom}(\Dmc^\approx_{S})
\mid D \in t_p \}$ and

\item $D^\Imc \cap \Delta^{\Imc_p} = D^{\Imc_p}\cap \Delta^{\Imc_p}$
  for all $p \in \mn{adom}(\Dmc^\approx_{S})$.
  
\end{enumerate}
From this, it easily follows that $\Imc$ is a model of
$\Omc$ and~$\Dmc^\approx_{S}$.
\end{proof}


\medskip

We now prove correctness of the careful chase algorithm.
\begin{restatable}{lemma}{lemhorn}
  \label{lem:horn}
  $\bar a \in Q^\approx_{\ELI^u_\bot}(\Dmc)$ iff $\bar a \in q(\Dmc')$.
\end{restatable}
\noindent
\begin{proof}\ We start with the following observation.
  \\[2mm]
  {\bf Claim.} The restriction of
  $\mn{ch}_{\Omc^\approx_{\ELI^u_\bot}}(\Dmc)$ to $\mn{adom(\Dmc)}$ is
  identical to the restriction of $\Dmc'$ to $\mn{adom(\Dmc)}$.  
  \\[2mm]
  For a proof, it suffices to show that for all $a,b \in
  \mn{adom}(\Dmc)$, the following conditions are satisfied:
  \begin{itemize}

  \item $r(a,b) \in \mn{ch}_{\Omc^\approx_{\ELI^u_\bot}}(\Dmc)$ iff $r(a,b)
    \in \Dmc'$.

    Immediate by definition of the chase and construction of~$\Dmc'$.
    In fact, $r(a,b) \in \mn{ch}_{\Omc^\approx_{\ELI^u_\bot}}(\Dmc)$ iff
    $r(a,b) \in \Dmc$ iff $r(a,b) \in \Dmc'$.

  \item $A(a) \in \mn{ch}_{\Omc^\approx_{\ELI^u_\bot}}(\Dmc)$ iff
    $A(a) \in \Dmc'$. 

    From $A(a) \in \mn{ch}_{\Omc^\approx_{\ELI^u_\bot}}(\Dmc)$, it
    follows that $\Dmc,\Omc^\approx_{\ELI^u_\bot} \models A(a)$ by
    Lemma~\ref{lem:unimod1} and thus $A(a) \in \Dmc'$ by construction
    of $\Dmc'$.

    Conversely, let $A(a) \in \Dmc'$. If $A(a) \in \Dmc$, then $A(a)
    \in \mn{ch}_{\Omc^\approx_{\ELI^u_\bot}}(\Dmc)$ and we are
    done. Otherwise, $\Dmc,
    \Omc^\approx_{\ELI^u_\bot} \models A(a)$ by construction
    of $\Dmc'$. Since $\Dmc$ is
    satisfiable
    w.r.t.\ $\Omc^\approx_{\ELI^u_\bot}$, we obtain $A(a) \in
    \mn{ch}_{\Omc^\approx_{\ELI^u_\bot}}(\Dmc)$.


  \end{itemize}
  Now for the proof of the lemma.  For the `only if' direction, assume that
  $\bar a \in Q^\approx_{\ELI^u_\bot}(\Dmc)$. By
  Lemma~\ref{lem:unimod1}
  and since \Dmc is satisfiable w.r.t. $\Omc^\approx_{\ELI^u_\bot}$,
  there is a homomorphism $h$ from $q$ to
  $\mn{ch}_{\Omc^\approx_{\ELI^u_\bot}}(\Dmc)$ with $h(\bar x)=\bar a$. It
  suffices to identify a homomorphism $g$ from $q$ to $\Dmc'$ with
  $g(\bar x) = \bar a$.

  By the claim, we can start the definition of $g$ by setting
  $g(x)=h(x)$ whenever $h(x) \in \mn{adom}(\Dmc)$.  For completing the
  definition, let $q'$ be obtained from $q$ in the following way:
  \begin{itemize}

  \item quantify all variables and then contract by identifying
    variables $x,y$ whenever $h(x)=h(y)$;

  \item restricting to those atoms that contain at least one variable
    $x$ with $h(x) \notin \mn{adom}(\Dmc)$;

  \item for all remaining variables $x$ with
    $h(x) \in \mn{adom}(\Dmc)$: view $x$ as an answer variable (drop
    quantification, if present), and re-add any atoms $A(x)$ and
    $r(x,x)$ from $a$.

\end{itemize}
    It is clear that
  $q'$ is a collection of ELIQs and BELIQs from
  $\mn{trees}(q)$. It suffices to extend $g$ to all these bELIQs.

  Let $p(\bar x)$ be a bELIQ in $q'$. Then
  $\Dmc,\Omc^\approx_{\ELI^u_\bot} \models p(h(\bar x))$ is witnessed by
  $h$.  By construction of
  $\Dmc'$, a copy of $p$ has been added to $\Dmc'$ with its root glued
  to $h(\bar x)$ in case that $p$ is an ELIQ and thus we can extend
  $g$ to all variables in $p(\bar x)$. If
  variables $x_1$ and $x_2$ have been identifyied during the
  construction of $p$ with the resulting variable being~$x_1$, then
  additionally set $g(x_2)=g(x_1)$.

  It is readily verified that $g$ is indeed a homomorphism from 
  $q$ to $\Dmc'$ with $g(\bar x) = \bar a$.

\medskip

For the `if' direction, assume that $\bar a \in q(\Dmc')$. Then there
is a homomorphism $h$ from $q$ to $\Dmc'$ with $h(\bar x) = \bar
a$. It suffices to show that there is a homomorphism $g$ from $\Dmc'$
to $\mn{ch}_{\Omc^\approx_{\ELI^u_\bot}}(\Dmc)$ with
$g(\bar a)=\bar a$ because then the composition of $h$ with $g$ yields
a homomorphism from $q$ to
$\mn{ch}_{\Omc^\approx_{\ELI^u_\bot}}(\Dmc)$ that witnesses
$\bar a \in Q^\approx_{\ELI^u_\bot}(\Dmc)$ by Lemma~\ref{lem:unimod1},
as required.

By the claim, we can start by setting $g(a)=h(a)$ for all
$a \in \mn{adom}(\Dmc)$.  Now assume that a copy of a bELIQ
$p(\bar x) \in \mn{trees}(q)$ has been added during the construction
of~$\Dmc'$, glueing the root to some $a \in \mn{adom}(\Dmc)$ if $p$ is
an ELIQ. Let $\bar a = a$ if $p$ is an ELIQ and $\bar a = ()$
otherwise. Then 
$\Dmc,\Omc^\approx_{\ELI^u_\bot} \models p(\bar a)$.  By
Lemma~\ref{lem:unimod1} and since \Dmc is satisfiable w.r.t.\
$\Omc^\approx_{\ELI^u_\bot}$, we find a homomorphism $h_p$ from
$p(\bar x)$ to $\mn{ch}_{\Omc^\approx_{\ELI^u_\bot}}(\Dmc)$ with
$h_p(\bar x)=\bar a$. We can extend $g$ to the added copy of
$p(\bar x)$ based on $h_p$.
\end{proof}
It remains to analyze the running time of the careful chase
algorithm. The computation of $\Dmc'$ needs
time $2^{O(||q||^2)} \cdot |\mn{adom}(\Dmc)| \cdot 2^{O(||Q||)} \cdot
O(||\Dmc||)$ while the size of $\Dmc'$ is bounded by $||\Dmc||+|\mn{adom}(\Dmc)|
\cdot 2^{O(||q||^2)}$. The unsatisfiability check in (i) runs in time
$2^{O(||Q||)} \cdot O(||\Dmc||)$.
Checking $\bar a \in q(\Dmc')$ in (ii) by brute force
means to test $||\Dmc'||^{||q||}$ candidate homomorphisms, that is,
single exponentially many in the case of combined complexity and
polynomially many in the case of data complexity, which gives the
\ExpTime and \PTime upper bounds in Point~1.  If $q \in
\text{UCQ}^{\text{tw}}_k$, we can check $\bar a \in q(\Dmc')$ using a
blackbox procedure that runs in \PTime combined
complexity~\cite{DBLP:journals/ai/DechterP89,DBLP:conf/aaai/Freuder90},
which yields the FPT upper bound in Point~2.

\section{Proofs for Section~\ref{sect:treedb}}
\label{app:treedb}


We prove the upper bounds in Theorem~\ref{thm:treeunravmain}.
\thmtreeunravmain*

\noindent
Let $Q(\bar x)=(\Omc,\Sigma,q) \in (\ALCI,\text{UCQ})$, \Dmc be a
$\Sigma$-database, and $\bar a \in \mn{adom}(\Dmc)^{|\bar x|}$.
%
Let $q^c$ be obtained from $q$ by replacing each CQ
$p$ with the UCQ that consists of all contractions $p'$ of
$p$ such that the restriction of
$p'$ to the quantified variables in it is a tree (recall that, for us,
trees need not be connected). The number of CQs in
$q^c$ is bounded by
$2^{||q||^2}$ and the size of each CQ is bounded by
$||q||$. Let $Q^c=(\Omc,\Sigma,q^c)$.
\begin{lemma}
\label{lem:contractiontotrees}
  $\bar a \in Q(\Dmc^\approx_{\bar a})$ iff 
  $\bar a \in Q^c(\Dmc^\approx_{\bar a})$. 
\end{lemma}
\noindent
\begin{proof}\ The ``if'' direction is immediate. For ``only if'',
  assume that $\bar a \notin Q^c(\Dmc^\approx_{\bar
    a})$. Then there is a forest model \Imc of $\Dmc^\approx_{\bar
    a}$ and \Omc such that $\bar a \notin
  q^c(\Imc)$, that is,
  $G_\Imc$ is obtained from $G_{\Dmc^\approx_{\bar
      a}}$ by attaching (possibly infinite) trees at constants from
  $\mn{adom}(\Dmc^\approx_{\bar
    a})$ \cite{DBLP:conf/cade/Lutz08}. It suffices to show that $\bar a \notin
  q(\Imc)$. Assume to the contrary that this is the case. Then there
  is a homomorphism $h$ from a CQ $p$ in $q$ to \Imc with $h(\bar
  x)=\bar a$.  Let $p'$ be obtained from $p$ by identifying variables
  $x$ and $y$ if $h(x)=h(y)$ and at least one of $x,y$ is not an
  answer variable. Using the definition of $\Dmc^\approx_{\bar
    a}$ and of $q^c$ and the fact that \Imc is a forest model,
  it is straightforward to show that $p'$ is a CQ in $p^c$.
  Moreover, $h$ is a homomorphism from $p'$ to \Imc, in
  contradiction to $\bar a \notin
  q^c(\Imc)$.
\end{proof}
We can thus work with $Q^c$ in place of $Q$. 

We next decorate \Dmc and decorate as well as simplify $q^c$.  Let
$\bar x = x_0 \cdots x_{n-1}$ and $\bar a = a_0 \cdots a_{n-1}$.  For
each (possibly inverted) role name $r$ used in $q$ and all $i < n$,
introduce a fresh concept name $A_{r,a_i}$ that, informally, expresses
the existence of an $r$-edge to $a_i$. Extend \Dmc to a database
$\Dmc^d$ by adding $A_{r,a_i}(b)$ for all $r(b,a_i) \in \Dmc$. This is
clearly possible in time $O(||\Dmc|| \cdot ||q||^2)$. Further, let $q^d$ be
obtained from $q^c$ by doing the following for each CQ $p$ in $q^c$:
\begin{enumerate}

\item replace each atom $r(y,x_i)$, $y$ a quantified variable, with
  $A_{r,a_i}(y)$;

\item for each atom $r(x_{i_1},x_{i_2})$, $0 \leq i_1,i_2 < n$, check
  whether $r(a_{i_1},a_{i_2}) \in \Dmc$; if the check fails, remove
  $p$ from the UCQ; if it succeeds, remove atom $r(x_{i_1},x_{i_2})$
  from $p$;

\item for each atom $A(x_i)$, $0 \leq i < n$, check whether
  $\Dmc^\approx_{\bar a},\Omc\models A(a_i)$ using
  Theorem~\ref{thm:eliqs} with $S=\bar a$; if the check fails, remove $p$ from the
  UCQ; if it succeeds, remove atom $A(x_{i})$ from $p$.
    
\end{enumerate}
If some of the CQs in $q^c$ becomes empty in the process (that is, all
of its atoms have been dropped), then we know that
$\bar a \in Q^c(\Dmc^\approx_{\bar a})$ and return `yes'. If all CQs
have been dropped, then we know that
$\bar a \notin Q^c(\Dmc^\approx_{\bar a})$ and return `no'. Clearly
all of the above can be done in time $2^{O(||Q||^2)} \cdot O(||\Dmc||)$.
Note that $q^d$ contains no answer variables as all atoms that
mention them have been dropped.
It is not hard to show the following using some basic manipulations of
homomorphisms that witness query answers.
\begin{lemma}
  $\bar a \in Q^c(\Dmc^\approx_{\bar a})$ iff
  $(\Dmc^d)^\approx_{\bar a},\Omc \models q^d$.
\end{lemma}
It thus remains to decide whether
$(\Dmc^d)^\approx_{\bar a},\Omc \models q^d$. To ease notation, from
now on we write $\Dmc$ instead of $\Dmc^d$.  By construction, $q^d$ is
a UCQ in which each CQ is a disjoint union of BELIQs. In other words,
it is a disjunction of conjunctions of BELIQs. Using the laws of
distributivity, we can convert it into an equivalent conjunction of
disjunctions of BELIQs $q_1 \wedge \cdots \wedge q_k$. To decide
whether $\Dmc^\approx_{\bar a},\Omc \models q^d$, it suffices to
decide whether $\Dmc^\approx_{\bar a},\Omc \models q_i$ for
$1 \leq i \leq k$. This can be done in time
$2^{O(||\Omc||+||q^d||)} \cdot O(||\Dmc||)$ by invoking Theorem~\ref{thm:eliqs}
with $S=\bar a$.

\smallskip

If $Q \in (\ALCI,\text{UCQ})$, then the above procedure runs in time
$2^{2^{O(||Q||^2)}} \cdot O(||\Dmc||)$, which yields both the 2\ExpTime
upper bound in combined complexity and the linear time upper bound in data
complexity in Point~1 of Theorem~\ref{thm:treeunravmain}. If
$Q \in (\ALCI,\text{bELIQ})$, then we can omit the initial
construction of $q^c$ and, as a consequence, obtain a running time
of $2^{O(||Q||)} \cdot O(||\Dmc||)$ and thus the \ExpTime upper bound
in combined complexity as well as the linear time uper bound in data
complexity in Point~2 of Theorem~\ref{thm:treeunravmain}.

\thmtwoexpwithouti*
\noindent
\begin{proof}\ In \cite{DBLP:conf/cade/Lutz08}, it is shown that
  evaluating OMQs from $(\ALCI,\text{CQ})$ is \TwoExpTime-hard on
  databases of the form $\{ A_0(a) \}$ and for Boolean OMQs that use
  only a single role name $r$. 
  The proof is by reduction from the word
  problem for exponentially space bounded alternating Turing machines
  (ATMs). It is not difficult to modify the reduction so that it uses
  exponentially \emph{time} bounded ATMs instead, thus only showing
  \ExpSpace-hardness.

  We thus know that the following problem is \ExpSpace-hard: given a
  Boolean OMQ $Q=(\Omc,\Sigma,q) \in (\ALCI,\text{CQ})$ with a single
  role name $r$ and a concept name $A_0 \in \Sigma$, decide whether
  there is a tree-shaped model \Imc of \Omc with $A_0$ true at the
  root and $\Imc \not\models q$. We call such an \Imc a \emph{tree
    witness} for $Q$ and~$A_0$'.  An inspection of the constructions
  in \cite{DBLP:conf/cade/Lutz08} shows that we can concentrate on
  witness trees that have depth at most $2^{n}-2$, $n$ the size of
  $Q$, and in which every node has at most 5 neighbors.\footnote{This
    is why we switch to exponentially time bounded ATMs; for
    exponentially space bounded ones, models can become double
    exponentially deep and it is not clear how to deal with that in
    the remainder of the reduction.} We refer to such an \Imc as a
  \emph{small} tree witness.
  
  Let $Q=(\Omc,\Sigma,q) \in (\ALCI,\text{CQ})$ be Boolean with single
  role name $r$ and \mbox{$A_0 \in \Sigma$}.  We construct in
  polynomial time a Boolean OMQ
  $Q'=(\Omc',\Sigma,q') \in (\ALC,\text{CQ})$ and $\Sigma$-database
  \Dmc such that there is a tree witness for $Q$ and $A_0$ if and only
  if $\Dmc^\approx,\Omc' \not\models q$. This gives the desired
  \ExpSpace lower bound.

  We start with the construction of the database \Dmc. The idea is to
  design $\Dmc^\approx$ so that small tree witnesses \Imc for $Q$ and
  $A_0$ can be embedded into models \Jmc of $\Dmc^\approx$ whose
  domain is $\mn{adom}(\Dmc^\approx)$, that is, when no new elements
  are introduced by existential quantification.  The constants used in \Dmc
  take the form
  $$
    a_{R,i,j} \text{ with } R \in \{r,r^-\}, \ 0 \leq i \leq 4, \text{
      and } 0 \leq j
  \leq 2.
  $$
  Informally, $R$ indicates that copies of $a_{R,i,j}$ in the
  unraveling serve as $R$-successors, $i$ is used to achieve that
  every constant in $\Dmc^\approx$ has 5 $r$-successors and 5
  $r^-$-successors (note that the 5 neighbors in small tree witnesses
  could be linked via $r$ or $r^-$, hence we prepare for both), and
  $j$ is used to implement a `directionality' in unravelings. Now,
  \Dmc contains the following facts, for all $R \in \{r,r^-\}$,
  $0 \leq i,i' \leq 4$, and $0 \leq j \leq 2$, and where $j \oplus 1$
  denotes $j+1$ modulo~5:
  \begin{itemize}

  \item $r(a_{R,i,j},a_{r,i',j \oplus 1})$ and $r(a_{r^-,i',j \oplus
      1},a_{R,i,j})$;

  \item $A_{R,i,j}(a_{R,i,j})$.

  \end{itemize}
  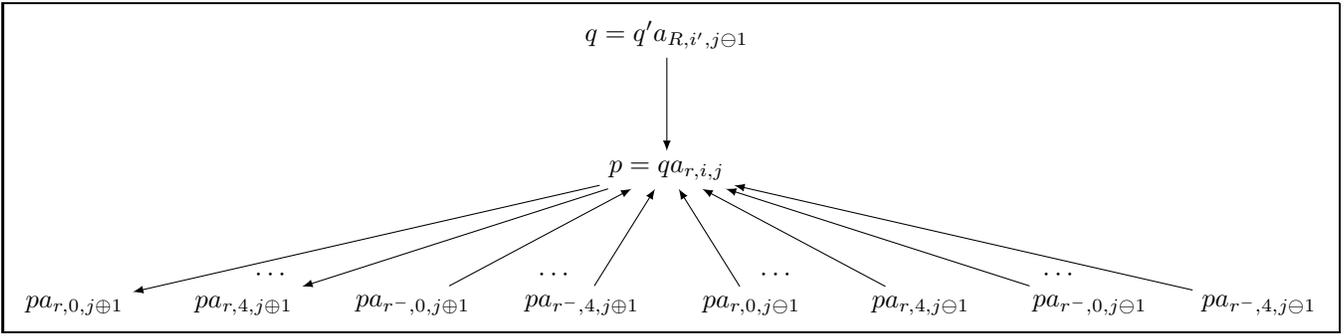
\begin{figure*}
      \begin{boxedminipage}{\textwidth}
    \centering 
\begin{tikzpicture}[scale=0.9,>=stealth',level/.style={sibling distance = 2.5cm, level distance = 2cm}, edge from parent/.style={draw,latex-}] 
\node {$q=q' a_{R,i',j\ominus 1}$}
    child{ node {$p=q a_{r,i,j}$} edge from parent [-latex]
            child{ node(1) {$p a_{r,0,j\oplus 1}$} edge from parent [-latex] }
            child{ node {$p a_{r,4,j\oplus 1}$} edge from parent [-latex] }
            child{ node {$p a_{r^-,0,j\oplus 1}$}}
            child{ node(2) {$p a_{r^-,4,j\oplus 1}$}}
            child{ node(3) {$p a_{r,0,j\ominus 1}$}}
            child{ node {$p a_{r,4,j\ominus 1}$}}
            child{ node {$p a_{r^-,0,j\ominus 1}$}}
            child{ node(4) {$p a_{r^-,4,j\ominus 1}$}}                                                
    };
\node [above right=of 1,yshift=-1.05cm,xshift=0.5cm] {$\cdots$};
\node [above=of 2,yshift=-1.05cm,xshift=-0.35cm] {$\cdots$};
\node [above=of 3,yshift=-1.05cm,xshift=0.35cm] {$\cdots$};
\node [above left=of 4,yshift=-1.05cm,xshift=-0.4cm] {$\cdots$};
\end{tikzpicture}
\end{boxedminipage}
\caption{Constant $qa_{r,i,j}$ and its neighborhood in $\Dmc^\approx$.}
\label{fig:neighbors}
\end{figure*}
  Recall that the constants of unraveled databases are paths. In
  Figure~\ref{fig:neighbors}, we show the neighborhood of a path in
  $\Dmc^\approx$ that ends with an constant of the form
  $a_{r,i,j}$. All edges are $r$-edges. The neighborhood of paths
  ending with constants of the form $a_{r^-,i,j}$ is similar, except
  that the direction of the edge from above and the right half of the
  edges from below is reversed. 

  We next assemble the ontology $\Omc'$.  We would like to say that
  the roots of $\Dmc^\approx$ are labeled with $A_0$ (or at least one
  of them is) while other `copies' of the same constant from $\Dmc$
  are not. However, there seems no way to do this since all such
  copies are bisimilar in $\Dmc^\approx$. This issue is in a sense the
  only obstacle in improving our \ExpSpace lower bound to a
  \TwoExpTime one.

  To address the issue, we install a binary counter that counts the depth of
  constants in $\Dmc^\approx$ modulo $2^n$. Every counter value is
  reached infinitely often and we make $A_0$ true at every constant
  with counter value~0. This allows us to embed infinitely many copies
  of a small tree witness into $\Dmc^\approx$. Note that the small
  tree witness has depth $2^{n}-2$, and thus the depth $2^{n}-1$ is
  present as a counter value in $\Dmc^\approx$, but not needed for
  representing nodes of the witness. We are instead going to use the
  constants with counter value $2^n-1$ to separate different copies
  of the tree witness from each other. This is done via a concept name
  $X$ that identifies those constants that `exist' in the sense that
  they correspond to a domain element of the small tree witness.

  We introduce concept names $L_0,\dots,L_{n-1}$ to implement the
  counter.  For every $R \in \{r, r^-\}$, $i \in \{0, 1, 2, 4\}$,
  $j \in \{0, 1, 2\}$ and $k \in \{0, \ldots, n-1\}$, $\Omc'$
  contains the following CIs:
\begin{align*}
  A_{R, i, j} \sqcap \bigsqcap_{0 \leq i < k} L_i \sqcap \neg L_{k} &\sqsubseteq \forall r.(A_{r, i', j \oplus 1} \rightarrow L_{k})\\
  A_{R, i, j} \sqcap \bigsqcap_{0 \leq i \leq k} L_i &\sqsubseteq
                                                       \forall
                                                       r.(A_{r, i', j
                                                       \oplus 1}
                                                       \rightarrow
                                                       \neg L_{k})\\
  A_{R, i, j} \sqcap \bigsqcup_{0 \leq i < k} \neg L_i \sqcap \neg L_{k} &\sqsubseteq \forall r.(A_{r, i', j \oplus 1} \rightarrow \neg L_{k}) \\
    A_{R, i, j} \sqcap \bigsqcup_{0 \leq i < k} \neg L_i \sqcap L_{k} &\sqsubseteq \forall r.(A_{r, i', j \oplus 1} \rightarrow L_{k}) \\
      \exists r.(A_{R, i, j} \sqcap \bigsqcap_{0 \leq i < k} L_i \sqcap \neg L_{k}) &\sqsubseteq A_{r^-, i', j \oplus 1} \rightarrow L_{k}\\
      \exists r.(A_{R, i, j} \sqcap \bigsqcap_{0 \leq i \leq k} L_i)
                                                                    &\sqsubseteq
                                                                      A_{r^-,
                                                                      i',
                                                                      j
                                                                      \oplus
                                                                      1}
                                                                      \rightarrow
                                                                      \neg
                                                                      L_{k}\\
  \exists r.(A_{R, i, j} \sqcap \bigsqcup_{0 \leq i < k} \neg L_i \sqcap \neg L_{k}) &\sqsubseteq A_{r^-, i', j \oplus 1} \rightarrow \neg L_{k} \\
    \exists r.(A_{R, i, j} \sqcap \bigsqcup_{0 \leq i < k} \neg L_i \sqcap L_{k}) &\sqsubseteq A_{r^-, i', j \oplus 1} \rightarrow L_{k}  
\end{align*}  
%
Note that the last four lines essentially serve the same purpose as
the first four ones. We have to use a different formulation, though,
as
we do not want to use quantification over inverse roles to stay
within \ALC.
We next say that depth~0 corresponds to the root of tree witnesses:
  $$
  \neg L_0 \sqcap \cdots \sqcap \neg L_{n-1} \sqsubseteq A_0 \sqcap X 
  $$
  The rest of the ontology $\Omc'$ is an adaptation of \Omc. In
  particular, we replace existential quantifiction with universal
  quantification, exploiting the fact that unraveling has already
  generated objects that can serve as witnesses for existential
  quantifiers. This also allows us to stay within \ALC despite the
  fact that the original ontology \Omc is formulated in \ALCI.

  We can assume w.l.o.g.\ that $\Omc$ has the form $ \{ \top
  \sqsubseteq C_\Omc \}$ with $C_\Omc$ in negation normal form, that
  is, negation is only applied to concept names, but not to compound
  concepts.
  Introducing fresh concept names, it is straightforward to further
  transform \Omc so that it consists only of 
  CIs of the form $\top \sqsubseteq A$, $A \sqsubseteq B$,
  $A \sqsubseteq \neg B$, $A \sqsubseteq B_1 \sqcup B_2$,
  $A \sqsubseteq \exists R . B$, and $A \sqsubseteq \forall R . B$
  where $A$, $B$, $B_1$, and $B_2$ are all concept names and
  $R \in \{r, r^-\}$.  The fresh concept names are of course not
  included in the database schema, so the resulting OMQ is equivalent
  to the original one. We reflect the CIs from \Omc in $\Omc'$ as follows:
  \begin{itemize}

  \item CIs $\top \sqsubseteq A,\ A \sqsubseteq B,\ A \sqsubseteq \neg
    B,\ A \sqsubseteq B_1 \sqcup B_2$ remain as they are except that
    we conjunctively add $X$ on the left-hand side;

  \item CIs $A \sqsubseteq \forall r . B$, become $X
  \sqcap A \sqsubseteq \forall r . (X \rightarrow B)$;
  
  \item CIs $A \sqsubseteq \forall r^- . B$, become $X \sqcap \exists r. (A \sqcap X)
  \sqsubseteq B$;
      
    \item CIs $A \sqsubseteq \exists R . B$, $R \in \{r,r^-\}$, are
    translated into the following CIs for all $i \in \{0,\dots,4\}$ 
    and $j \in \{0,1, 2\}$:
      $$
      \begin{array}{@{}r@{\;}c@{\;}l}
        X \sqcap A \sqcap A_{R,i,j} &\sqsubseteq& \bigsqcup_{0 \leq i' \leq 4}
        \forall r . (A_{R,i',j \oplus 1} \rightarrow (X \sqcap B)) \\[4mm]
        X \sqcap A \sqcap A_{R^-,i,j} &\sqsubseteq& \bigsqcup_{0 \leq i' \leq 4}
        \forall r . (A_{R,i',j \oplus 1} \rightarrow (X \sqcap B)). 
      \end{array}
    $$

  \end{itemize}
  This finishes the construction of the ontology $\Omc'$.  The CQ $q'$
  is obtained from $q$ by adding the atom $X(x)$ for every variable
  $x$ in $q$.  Let $Q'=(\Omc',\Sigma,q')$.  It remains to show the
  following.
  \\[2mm]
  {\bf Claim.} There is a small tree witness for $Q$ and $A_0$ if and
  only if $\Dmc^\approx,\Omc' \not\models q'$.
  \\[2mm]
  Let $\Imc$ be a small tree witness for $Q$ and $A_0$.  We construct
  a model $\Jmc$ of $\Dmc^\approx$ and $\Omc'$ that witnesses
  $\Dmc^\approx,\Omc' \not\models q'$. Intuitively, $\Jmc$ is obtained
  by representing infinitely many copies of $\Imc$ in $\Dmc^\approx$,
  where elements of $\mn{adom}(\Imc)$ are represented by constants
  that satisfy $X$.  To construct $\Jmc$, we start with $\Dmc^\approx$
  seen as an interpretation and then extend concept extensions, but we
  do neither extend role assertions nor add new constants. First, we
  assign counter values to all constants in $\Delta^\Jmc$.  This can
  be done by starting with the roots, that is, with all paths in
  $\Delta^\Jmc$ that consist of a single constant. Assing to all
  such $a \in \Delta^\Jmc$ the counter value zero, thus
  $a \notin L_i^\Jmc$ for $0 \leq i <n$. Then inductively assign
  incremented counter values (modulo $2^n$) to all neighbours of
  constants that have already been assigned a counter value.

Next, we embed $\Imc$ into $\Jmc$ infinitely often. Let $Z \subseteq \Delta^\Jmc$
be the set of elements that have been assigned counter value zero.
For every $a \in Z$,  inductively construct
an injective function $h_a$ from $\Delta^\Imc$ to $\Delta^\Jmc$ as follows:
\begin{itemize}
\item $h_a$ maps the root of $\Imc$ to $a$;
\item if $h_a(b) = c \in A_{R,i,j}^\Jmc$ and $(b,b') \in r^\Imc$ with
  $h_a(b')$ undefined, 
  then choose as $h_a(b')$ an $r$-successor of $h_a(b)$ in \Jmc
  that is in $A_{r,i',j \oplus 1}^\Jmc$ for some $i'$;
\item if $h_a(b) = c \in A_{R,i,j}^\Jmc$ and $(b',b) \in r^\Imc$,
then choose as $h_a(b')$ an $r$-successor of
$h_a(b)$ in \Jmc that is in $A_{r^-,i',j \oplus 1}^\Jmc$ for some $i'$.
\end{itemize}
Since every node in $\Imc$ has at most 5 neighbors, such an injective
function exists; and since the depth of
\Imc is at most $2^{n}-1$, $h_a$ never hits an element of \Jmc whose
counter value is $2^n-1$.  We
extend $\Jmc$ with concept names such that every $h_a$ becomes a
homomorphism, and additionally set
$X^\Jmc = \bigcup_{a \in Z} \mn{range}(h_a)$.  This finishes the
definition of $\Jmc$. It can be checked that $\Jmc$ is indeed a model
of $\Omc'$.

For the other direction, let $\Dmc^\approx, \Omc' \not \models q'$, so
there is a model $\Jmc$ of $\Dmc^\approx$ and $\Omc'$ with
$\Jmc \not \models q'$. By construction of $\Omc'$, we can assume
w.l.o.g.\ that $\Delta^\Jmc = \mn{adom}(\Dmc^\approx)$.  Since $\Omc'$
enforces a counter that counts modulo $2^{n}$, there are infinitely
many elements $a \in \Delta^\Jmc$ such that $a \notin L_i^\Jmc$ for
$0 \leq i < n$. From now on, let $a$ be a fixed such element.  We
construct a tree witness $\Imc$ for $Q$ and $A_0$.  Let $\Delta^\Imc$
be the smallest subset of $\Delta^\Jmc$ that contains $a$ and such
that if $b \in \Delta^\Imc$, $(b,b') \in r^\Jmc \cup (r^-)^\Jmc$, and
$b' \in X^\Jmc$, then $b' \in \Delta^\Imc$.  For every symbol
$\alpha \in \mn{sig}(\Omc) \cup \Sigma$, let $\alpha^\Imc$ be the
restriction of $\alpha^\Jmc$ to $\Delta^\Imc$. It can be checked that
$\Imc$ is indeed a tree witness for $Q$ and $A_0$.

\medskip

We only sketch the idea of the 2\ExpTime lower bound for
$(\ALC,\text{UCQ})$. We again start from the result established in
\cite{DBLP:conf/cade/Lutz08} that evaluating Boolean OMQs from
$(\ALCI,\text{CQ})$ is \TwoExpTime-hard on databases of the form
$\{ A_0(a) \}$. Unlike for the \ExpSpace version, however, tree
witnesses now have double exponential depth. This is a main obstacle
to re-using the above proof because we cannot
implement a double exponential counter using the ontology. The
solution
is to implement it using the query instead, which for this purpose
we need to be a UCQ.

Let us get into some more detail. We implement a binary counter that
counts modulo $2^{2^n}$ by attaching a binary subtree of depth $n$
below each of the already existing nodes of $\Dmc^\approx$.  The $2^n$
leaves of the tree serve as the bit positions of the counter and
concept names $T$ and $F$ are used to signal the value of each bit. We
can make room for the additional trees by letting $i$ range from $0$
to $6$ instead of from $0$ to $4$.

The ontology $\Omc'$ is constructed to ensure that exactly one of $T$
and $F$ is true at each leaf of a counting tree. It then suffices to
construct a Boolean CQ $p$ such that for any model \Jmc of $\Omc'$
and $\Dmc^\approx$ with $\Delta^\Jmc=\mn{adom}(\Dmc^\approx)$,
$\Jmc \models p$ iff somewhere in \Jmc the counter is not properly
incremented.  For if we have found such a $p$, we can use $p' \vee q$
in place of $q'$ in the reduction presented above.

Now, the (non-trivial) CQs constructed in the mentioned proof in
\cite{DBLP:conf/cade/Lutz08} are exactly what is needed for $p$. While
there are no counting trees in \cite{DBLP:conf/cade/Lutz08}, there are
configuration trees for exponentially space bounded ATMs, and
identifying incrementation defects in counting trees is essentially
the same task as identifying defects in ATM transitions (the former is
actually slightly simpler). We refrain from repeating the details of
the construction of $p$ and only remark that $p$ uses an alternation
of inverse $r$-edges and non-inverse $r$-edges, and thus the edges
of our counting trees need to consist of an inverse $r$-edge followed
by an $r$-edge (and thus these trees actually have depth $2n$ rather
than depth $n$, but branch only on every second level).
\end{proof}

\section{Proofs for Section~\ref{sect:tgd}}
\label{app:tgd}

\prophowwechoose*

\noindent
\begin{proof}\ {\sc coNP}-hardness is shown as in
  Example~\ref{ex:threecolagain}: consider the Boolean OMQ
  $Q (\Omc,\{e\},\exists x \, D(x))$ given there that expresses
  non-3-colorability. Then $\Omc^\approx_{\omega,\omega,1,1}$ contains
  the TGD $q_\Dmc\rightarrow \exists x \, D(x)$ for every undirected
  graph \Dmc that is not 3-colorable. Thus,
  $\Omc^\approx_{\omega,\omega,1,1}$ is equivalent to $Q$. 
  
  For \TwoExpTime-hardness, we recall the result from
  \cite{DBLP:conf/cade/Lutz08} that evaluating OMQs from
  $(\ALCI,\text{CQ})$ is \TwoExpTime-hard. It is easy to verify that
  the proof uses only Boolean OMQs, CQs of bounded treewidth, and
  databases of the simple form $\{ A(a) \}$. In other words, there are
  $\ell',k'$ such that given an OMQ $Q=(\Omc,\Sigma,q) \in
  (\ALCI,\text{CQ})$ with $q$ of treewidth $(\ell',k')$ and a
  database of the form $\Dmc = \{ A(a) \}$, it is \TwoExpTime-hard to
  decide whether $\Dmc \models Q$. This, however, is the case if and
  only if $\Dmc \models
  Q^\approx_{\textit{$\omega,\omega,1,1$}}=(\Omc^\approx_{\omega,\omega,1,1},\Sigma,q)$
  since $\Dmc \models Q$ implies that 
$\Omc^\approx_{\omega,\omega,1,1}$ contains the TGD $A(x) \rightarrow q()$.
\end{proof}

\medskip We now turn to Example~\ref{ex:leif}, providing the missing
proof that $\Dmc \not \models Q_{\omega, \omega, 1, m^2-1}^\approx$.

\smallskip

It is sufficient to argue that $\Umc \not \models q_m$, where for
brevity
$\Umc=\mn{ch}_{\Omc_{\omega, \omega, 1, m^2-1}^\approx}(\Dmc)$.
Assume the contrary that there is a homomorphism $h$ from $q_m$ to
$\Umc$. By construction, $\Umc$ has treewidth $(1, m^2-1)$.
The query $q_m$ has treewidth $(1, m^2)$, but not treewidth $(1, m^2-1)$, so
$h$ is not injective, say $h(x_{i_0,j_0}) = h(x_{i_1,j_1})$ for some
$i_0,j_0,i_1,j_1 \in \{1,\ldots,m\}$ with $(i_0,j_0) \neq
(i_1,j_1)$. Since $h$ is a homomorphism,
$h(x_{i_0,j_0}) \in A_{i_0,j_0}^\Umc \cap A_{i_1,j_1}^\Umc$. By
analysing the TGDs in $\Omc_{\omega, \omega, 1, m^2-1}^\approx$,
however, it is clear that $\Umc$ does not contain any element that
satisfies both $A_{i_0,j_0}$ and $A_{i_1,j_1}$.

\propmaxk*

\noindent
\begin{proof}\ The proposition can be proved by analyzing the careful
  chase algorithm presented below. In fact, it is a consequence of
  Lemma~\ref{lem:tgdcarefulchasecorr} and the fact that
  $\mn{trees}^{1,k'}=\mn{trees}^{1,|\mn{var}(q)|}$ whenever $k ' \geq
  |\mn{var}(q')|$ which implies that the constructed database $\Dmc'$
  is
  exactly identical for all such~$k'$.
\end{proof}

We now turn to giving a detailed description of the careful chase
algorithm and proving its correctness, with the aim of establishing
the upper bounds in Theorem~\ref{thm:tgdmain}. For doing so, we assume
that Theorem~\ref{thm:eliqsgen} is already available. A proof of the
latter is provided in the subsequent section. We start with the
observation that it suffices to treat CQs in place of UCQs.  The
following is a consequence of the fact that TGD-ontologies have
universal models, even if infinite.
\begin{lemma}
\label{lem:UCQtoCQTGDs}
Let $Q(\bar x)=(\Omc,\Sigma,q) \in (\ALCI,\text{UCQ})$,
$q=p_1 \vee \cdots \vee p_n$, \Dmc a $\Sigma$-database,
and $\ell,k,\ell',k' \geq 1$ with $\ell < k$ and $\ell' <k'$.
Then $Q^\approx_{\ell,k,\ell',k'}(\Dmc)=Q_1(\Dmc) \cup \cdots \cup
Q_n(\Dmc)$, $Q_i=(\Omc^\approx_{\ell,k,\ell',k'},\Sigma,p_i)$ for
$1 \leq i \leq p$.
\end{lemma}
Together with Theorem~\ref{thm:eliqsgen}, the following lemma 
allow us to check satisfiability of a database \Dmc w.r.t.\ the
infinite ontology $\Omc_{\ell,k,\ell',k'}^{\approx}$, given only \Dmc
and \Omc. 
\begin{lemma}\label{lem:sat5}
For any database $\Dmc$ and $\ALCI$-ontology $\Omc$, $\Dmc$ is satisfiable w.r.t.~$\Omc_{\ell,k,\ell',k'}^{\approx}$ iff $\Dmc_{\ell,k}^{\approx}$ is satisfiable w.r.t.~$\Omc$.
\end{lemma}
\noindent
\begin{proof} \
	Assume first that $\Dmc_{\ell,k}^{\approx}$ is satisfiable w.r.t.~$\Omc$.
	Then $\Dmc_{\ell,k}^{\approx}$ is satisfiable w.r.t.~$\Omc_{\ell,k,\ell',k'}^{\approx}$ since $\Omc\models \Omc_{\ell,k,\ell',k'}^{\approx}$. It follows that $\Dmc$ is satisfiable w.r.t.~$\Omc_{\ell,k,\ell',k'}^{\approx}$, by Lemma~\ref{lem:untol}.
	Conversely, for an indirect proof, assume that $\Dmc_{\ell,k}^{\approx}$
	is not satisfiable w.r.t.~$\Omc$. By compactness, there is a finite subset $\mathcal{F}$ of $\Dmc_{\ell,k}^{\approx}$ that is not satisfiable w.r.t.~$\Omc$. We view $\mathcal{F}$ as the head $F$ of a $\ell,k,1,1$-TGD
	$F \rightarrow \bot$. Then $\Omc\models F \rightarrow \bot$ and so
	$F\rightarrow \bot\in \Omc^{\approx}_{\ell,k,\ell',k'}$. It follows that  $\Dmc_{\ell,k}^{\approx}$ is not satsifiable w.r.t.~$\Omc^{\approx}_{\ell,k,\ell',k'}$. But then, by Lemma~\ref{lem:untol},
	$\Dmc$ is not satisfiable w.r.t.~$\Omc_{\ell,k,\ell',k'}^{\approx}$.
\end{proof}		
We next describe the careful chase algorithm.

\smallskip

Fix $\ell,k,k' \geq 1$ with $\ell < k$. Let
$Q(\bar x)=(\Omc,\Sigma,q) \in (\ALCI,\text{CQ})$, \Dmc a
$\Sigma$-database, and $\bar a \in \mn{adom}(\Dmc)^{|\bar x|}$.  We
want to decide whether $\bar a \in Q^\approx_{\ell,k,1,k'}(\Dmc)$.

We first check whether $\Dmc$ is satisfiable
w.r.t.~$\Omc^\approx_{\ell,k,1,k'}$.  By Lemma~\ref{lem:sat5}, this is
the case if $\Dmc_{\ell,k}^{\approx}$ is satisfiable w.r.t.~$\Omc$.
The latter problem can be reduced in polynomial time to the complement
of the problem considered in Theorem~\ref{thm:eliqsgen}, and thus
in \ExpTime and within the time requirements of FPT. Indeed,
as we have seen already in Appendix~\ref{app:eliu},
$\Dmc_{\ell,k}^{\approx}$ is satisfiable w.r.t.~$\Omc$ iff
$\Omc,\Dmc_{\ell,k}^{\approx}\not\models \exists x A(x)$, where $A$ is
a fresh concept name.

\smallskip

We may thus assume that $\Dmc$ is satisfiable
w.r.t.~$\Omc^\approx_{\ell,k,1,k'}$.  Let $\mn{trees}^{1,k'}(q)$
denote the set of all CQs of treewidth $(1,k')$ that can be obtained
from $q$ by first quantifying all variables, then taking a
contraction, then an induced subquery, and then selecting at most one
variable as the answer variable (thus dropping quantification from
it). In addition, $\mn{trees}^{1,k'}(q)$ contains all AQs $A(x)$ with
$A$ a concept name used in \Omc.  Note that all CQs in
$\mn{trees}^{1,k'}(q)$ are unary or Boolean.

Now extend \Dmc to a database $\Dmc'$ as follows. For every CQ $p(\bar
x) \in \mn{trees}^{1,k'}(q)$ and every $a \in \mn{adom}(\Dmc)$ with
$\langle \Dmc^\approx_{\ell,k}, a \rangle,\Omc \models p(\bar a)$,
take a disjoint copy of $p(x)$ viewed as a database and add it
to~\Dmc; if $p$ is unary, then glue the root to the single constant
$a$ in~$\bar a$. Note that $\langle \Dmc^\approx_{\ell,k}, a
\rangle,\Omc \models p(\bar a)$ is exactly what
Theorem~\ref{thm:eliqsgen} allows us to decide, in 
\ExpTime and within the time requirements of FPT.
Note that $\Dmc'$ is homomorphically equivalent to
a subdatabase of $\mn{ch}_{\Omc^\approx_{\ell,k,1,k'}}(\Dmc)$, which
is why we speak of a careful chase. 
The algorithm now checks whether $\bar a \in q(\Dmc')$ using brute
force (to attain \ExpTime combined complexity) or using as a blackbox
an algorithm that runs within the time requirements of fixed-parameter
tractability (to attain FPT) and returns the result. Correctness is
established by the following lemma.
\begin{lemma}
\label{lem:tgdcarefulchasecorr}
  $\bar a \in Q^\approx_{\ell,k,1,k'}(\Dmc)$ iff $\bar a \in q(\Dmc')$.
\end{lemma}
%
%
%
\noindent
\begin{proof}\
  We can show the following in exactly the same way as in the proof of
  Lemma~\ref{lem:horn}.
  \\[2mm]
  {\bf Claim~1.} The restriction of
  $\mn{ch}_{\Omc_{\ell,k,1,k'}^{\approx}}(\Dmc)$ to $\mn{adom(\Dmc)}$ is
  identical to the restriction of $\Dmc'$ to $\mn{adom(\Dmc)}$.  
  \\[2mm]
  Now for the proof of the lemma.  For the `if' direction, assume that
  $\bar a \in Q^\approx_{_{\ell,k,1,k'}}(\Dmc)$. By
  Lemma~\ref{lem:unimod1}, there is a homomorphism $h$ from $q$ to
  $\mn{ch}_{\Omc^\approx_{\ell,k,1,k'}}(\Dmc)$ with $h(\bar x)=\bar
    a$. It suffices to identify a homomorphism $g$ from $q$ to $\Dmc'$
    with $g(\bar x) = \bar a$.

    By Claim~1, we can start the definition of $g$ by setting
    $g(x)=h(x)$ whenever $h(x) \in \mn{adom}(\Dmc)$.  For completing
    the definition, let CQ $p$ be obtained from $q$ by quantifying all
    variables and then contracting by identifying variables $x,y$
    whenever $h(x)=h(y)$. Further, let $p^-$ be the restriction of $p$
    to those atoms that contain at least one variable $x$ with $h(x)
    \notin \mn{adom}(\Dmc)$, viewing the remaining variables $x$ with
    $h(x) \in \mn{adom}(\Dmc)$ as answer variables. It is clear that
    $p^-$ is a collection of CQs of treewidth $(1,k')$ with at
    most one answer variable each. It suffices to extend $g$ to all
    these CQs.

    Let $\widehat p(x)$ be a unary CQ in $p^-$. Then
    $\Dmc,\Omc^\approx_{\ell,k,1,k'} \models \widehat p(h(x))$ is
    witnessed by~$h$. Point~3 of Lemma~\ref{lem:untol} implies
    $\langle \Dmc^\approx_{\ell,k},
    h(x)\rangle,\Omc^\approx_{\ell,k,1,k'} \models \widehat p(h(x))$.
    Since $\Omc \models \Omc^\approx_{\ell,k,1,k'}$, this yields
    $\langle \Dmc^\approx_{\ell,k},h(x)\rangle,\Omc \models \widehat
    p(h(x))$.  By construction of $\Dmc'$, a copy of $\widehat p$ has
    been added to $\Dmc'$ with its root glued to $h(x)$ and thus we
    can extend $g$ to all variables in $\widehat p(x)$. The case of
    Boolean CQs from $p^-$ is analogous.  If variables $x_1$ and $x_2$
    have been identifyied during the construction of $p$ with the
    resulting variable called $x_1$, then additionally set
    $g(x_2)=g(x_1)$.

  It is readily verified that $g$ is indeed a homomorphism from 
  $q$ to $\Dmc'$ with $g(\bar x) = \bar a$.

\medskip

For the `only if' direction, assume that $\bar a \in q(\Dmc')$. Then
there is a homomorphism $h$ from $q$ to $\Dmc'$ with $h(\bar x) = \bar
a$. It suffices to show that there is a homomorphism $g$ from $\Dmc'$
to $\mn{ch}_{\Omc^\approx_{\ell,k,1,k'}}(\Dmc)$ with $g(\bar a)=\bar
a$ because then the composition of $h$ with $g$ yields a homomorphism
from $q$ to $\mn{ch}_{\Omc^\approx_{\ell,k,1,k'}}(\Dmc)$ that
witnesses that $\bar a \in Q^\approx_{\ell,k,1,k'}(\Dmc)$, as
required.

By the claim, we can start by setting $g(c)=h(c)$ for all $c \in
\mn{adom}(\Dmc)$.  Now assume that a copy of a unary CQ $p(x) \in
\mn{trees}(q)$ has been added during the construction of~$\Dmc'$,
glueing the root to some $a \in \mn{adom}(\Dmc)$. Then
$\langle \Dmc^\approx_{\ell,k},a\rangle,\Omc \models p(a)$.  Compactness yields a
finite subset $\Fmc$ of $\langle\Dmc^\approx_{\ell,k},a\rangle$ with $\Fmc,\Omc
\models p(a)$. We can view \Fmc as a unary CQ $\widehat p(x)$ of
treewidth $(\ell,k)$ where the free variable $x$ corresponds
to the constant $a$. Then $\Omc \models \widehat p(x) \rightarrow
p(x)$ and thus $\widehat p(x) \rightarrow p(x)$ is a CI in
$\Omc^\approx_{\ell,k,1,k'}$. Consequently,
$\langle\Dmc^\approx_{\ell,k},a\rangle,\Omc^\approx_{\ell,k,1,k'} \models p(a)$
and Point~1 of Lemma~\ref{lem:untol} yields
$\Dmc,\Omc^\approx_{\ell,k,1,k'} \models p(a)$. By
Lemma~\ref{lem:unimod1}, we find a homomorphism $h_p$ from $p(x)$ to
$\mn{ch}_{\Omc^\approx_{\ell,k,1,k'}}(\Dmc)$ with $h_p(x)=a$. We can
extend $g$ to the added copy of $p(x)$ at $a$ based on $h_p$.
The case of Boolean CQs is similar.
\end{proof}
It remains to analyze the running time of the careful chase
algorithm. The initial satisfiability check runs in \ExpTime and FPT
and the computation of $\Dmc'$ needs time $2^{O(||q||^2)} \cdot
|\mn{adom}(\Dmc)|$ entailment checks, thus overall also running
in \ExpTime and FPT. The size of $\Dmc'$ 
is bounded by $||\Dmc||+|\mn{adom}(\Dmc)| \cdot 2^{O(||q||^2)}$.
Checking $\bar a \in q(\Dmc')$ by brute force means to test
$||\Dmc'||^{||q||}$ candidate homomorphisms, that is, single
exponentially many in the case of combined complexity and polynomially
many in the case of data complexity, which gives the \ExpTime and
\PTime upper bounds in Point~1.  If $q \in \text{UCQ}^{\text{tw}}_k$,
we can check $\bar a \in q(\Dmc')$ using a blackbox procedure that
runs in \PTime combined
complexity~\cite{DBLP:journals/ai/DechterP89,DBLP:conf/aaai/Freuder90},
which yields the FPT upper bound in Point~2.

\section{Proof of Theorem~\ref{thm:eliqsgen}}

\thmeliqsgen*


To prove Theorem~\ref{thm:eliqsgen}, we use an approach based on
two-way alternating parity tree automata. We first introduce this
automata model.

Let \Nbbm\xspace denote the positive integers $\{1,2,\dots\}$
and let $\Nbbm^*$ denote the set of all words over \Nbbm\xspace viewed
as an infinite alphabet. A \emph{tree} is a non-empty (potentially
infinite) set $T \subseteq \Nbbm^*$ closed under prefixes.  The node
$\varepsilon \in T$ is the \emph{root} of $T$. An \emph{infinite path}
$P$ in $T$ is a prefix-closed set $P \subseteq T$ such that for every
$i \geq 0$, there is a unique word $x \in P$ that is of length $i$.
As a convention, we take $x \cdot 0 = x$ and $ (x \cdot c) \cdot -1 =
x$. Note that $\varepsilon \cdot -1$ is undefined.  We say that $T$ is
\emph{$m$-ary}, $m \geq1$, if for every $x \in T$, the set $\{ i \mid
x \cdot i \in T \}$ is of cardinality at most $m$.  W.l.o.g., we
assume that all nodes in an $m$-ary tree are from $\{1,\dots,m\}^*$.
For an alphabet $\Gamma$ and $m \geq 1$, a \emph{$\Gamma$-labeled
  $m$-ary tree} is a pair $(T,L)$ with $T$ a tree and $L:T \rightarrow
\Gamma$ a node labeling function.

For a
set $X$, we use $\Bmc^+(X)$ to denote the set of all positive Boolean
formulas over $X$, i.e., formulas built using conjunction and
disjunction over the elements of $X$ used as propositional variables,
and where the special formulas $\mn{true}$ and $\mn{false}$ are
admitted as well.
%
\newcommand{\dia}[1]{\langle #1 \rangle}
\begin{defn}[TWAPA]
  A \emph{two-way alternating parity automaton 
    (TWAPA) on $m$-ary trees} is a tuple
  $\Amf=(S,\Gamma,\delta,s_0,c)$ where $S$ is a finite set of
  \emph{states}, $\Gamma$ is a finite alphabet, $\delta: S \times
  \Gamma \rightarrow \Bmc^+(\mn{tran}(\Amf))$ is the \emph{transition
    function} with $\mn{tran}(\Amf) = \{ \dia{i} s, \ [i] s
  \mid -1 \leq
  i \leq m \text{ and } s \in S \}$ the set of
  \emph{transitions} of \Amf, $s_0 \in S$ is the \emph{initial state},
  and $c:S \rightarrow \Nbbm$ is the \emph{parity condition} that 
  assigns to each state a \emph{priority}.
\end{defn}
Intuitively, a transition $\dia{i} s$ with $i>0$ means that
a copy of the automaton in state $s$ is sent to the $i$-th successor
of the current node, which is then required to exist. Similarly,
$\dia0 s$ means that the automaton stays at the current
node and switches to state $s$, and $\dia{-1} s$ indicates
moving to the predecessor of the current node, which is then required
to exist. Transitions $[i] s$ mean that a copy of the automaton in
state $s$ is sent to the relevant node if that node exists
(which is not required).
%
\begin{defn}[Run, Acceptance]
  A \emph{run} of a TWAPA $\Amf = (S,\Gamma,\delta,s_0,c)$ on a 
  $\Gamma$-labeled tree $(T,L)$ is a $T \times S$-labeled tree
  $(T_r,r)$ such that: 
  \begin{enumerate}

  \item $r(\varepsilon) = ( \varepsilon, s_0)$;
    

  \item if $y \in T_r$, $r(y)=(x,s)$, and $\delta(s,L(x))=\vp$, then
    there is a (possibly empty) set $S \subseteq \mn{tran}(\Amf)$ such
    that $S$ (viewed as a propositional valuation) satisfies $\vp$ as
    well as the following conditions:
    \begin{enumerate}

    \item if $\dia{i} s' \in S$, then $x \cdot i$ is defined and 
      there is a node $y \cdot j \in T_r$ such that $r(y \cdot j)=(x 
      \cdot i,s')$;

    \item if $[i]s' \in S$ and $x \cdot i$ is defined and in $T$, then
      there is a
      node $y \cdot j \in T_r$ such that $r(y \cdot j)=(x \cdot
      i,s')$.

    \end{enumerate}

  \end{enumerate}
  We say that $(T_r,r)$ is \emph{accepting} if on all infinite paths
  in $T_r$, the maximum priority that appears infinitely often is
  even.  A $\Gamma$-labeled tree $(T,L)$ is \emph{accepted} by
  \Amf if there is an accepting run of \Amf on $(T,L)$. We use
  $L(\Amf)$ to denote the set of all $\Gamma$-labeled tree
  accepted by \Amf.
\end{defn}
It is known (and easy to see) that complement and intersection of
TWAPAs can be implemented with only a polynomial blowup.
%
It is also known
that 
their emptiness problem can be solved in time single exponential in
the number of states and the maximum priority and polynomial in all
other components of the automaton
\cite{DBLP:conf/lop/Vardi85}. 
In what follows, we shall generally only explicitly
analyze the number of states of a TWAPA, but implicitly also take care
that all other components are of the appropriate size for the complexity
result that we aim to obtain. In particular, the maximum priority will
always be 2.
%
%
%

\medskip

For the \ExpTime upper bound in Theorem~\ref{thm:eliqsgen}, let $\ell,
k, k', Q, \Dmc$ and $\bar a$ be as in the statement of the theorem
with $Q(\bar x)=(\Omc,\Sigma,q)$. We show how to constuct a TWAPA~\Amf
with polynomially many states such that $L(\Amf) = \emptyset$ iff
$\bar a \in Q(\langle \Dmc^\approx_{\ell,k},\bar a \rangle)$.  By what
was said above, this yields the desired $\ExpTime$ algorithm.

It is an immediate consequence of the semantics that $\bar a \notin
Q(\langle\Dmc^\approx_{\ell, k},\bar a \rangle)$ if and only if there
is model \Imc of \Omc and $\langle \Dmc^\approx_{\ell, k},\bar a \rangle$ such that
$\Imc \not \models q(\bar a)$. We call such a model a \emph{witness}
for $\bar a \notin Q(\langle\Dmc^\approx_{\ell, k},\bar a \rangle)$.  
The witness \Imc is \emph{forest like} if the interpretation obtained
from \Imc by replacing $r^\Imc$ with $r^\Imc \setminus
\mn{adom}(\langle \Dmc^\approx_{\ell, k},\bar a \rangle) \times
\mn{adom}(\langle \Dmc^\approx_{\ell, k},\bar a \rangle)$ for all role
names $r$ is a disjoint union of (potentially infinite) trees.  It is
well-known that if there is a witness for $\bar a \notin Q(\langle
\Dmc^\approx_{\ell, k},\bar a \rangle)$, then there is also one that
is forest like with the tree parts of outdegree at most $|\Omc|$, see
for instance \cite{DBLP:conf/dlog/Lutz08}. Witnesses of this kind can
be encoded as labeled trees and thus presented as an input to the
TWAPA~\Amf.

We define \Amf to run on $m$-ary $\Gamma$-labeled trees where
$m=|\Omc|+(|\mn{adom}(\Dmc)|^k \cdot 2^k)$ and for $N = \mn{adom}(\Dmc) \cup
\{1, 2, 3\}$, the input alphabet is 
$$
\begin{array}{rcl}
\Gamma &=& \{ \bot \} \cup \{\ (\Bmc,O) \mid
\Bmc \text{ is a } \Sigma \cup \mn{sig}(\Omc)\text{-database with}
\\[1mm]
&&\qquad\qquad \mn{adom}(B)
\subseteq N \text{ and }
|\mn{adom}(\Bmc)| \leq k, \\[1mm]
&&\qquad\qquad \text{and } O \subseteq \mn{adom}(\Bmc)\ \}.
\end{array}
$$
Informally, $\bot$ is a special label for the root node, the constants
from $\mn{adom}(\Dmc)$ are used to represent the copies of such
constants in $\langle \Dmc^\approx_{\ell, k} ,\bar a \rangle$, which
is a part of every forest like witness interpretation \Imc, and the
constants $\{1,2,3\}$ are used to represent the remaining, tree-shaped
parts of \Imc. If $L(t)=(\Bmc,O)$, then $O$ represents the overlap
between $\mn{adom}(\Bmc)$ and the domain of the database associated
with the predecessor of $t$ in the tree, c.f.\ the definition of
$\ell,k$-unravelings. For brevity, we then write $L_1(t)$ to denote
\Bmc and $L_2(t)$ for $O$. Whenever we write $L_1(t)$ or $L_2(t)$, we
silently assume that $L(t) \neq \bot$.  We say that $(T,L)$ is
\emph{proper} if it satisfies the following conditions for all $t,t'
\in T$:
\begin{enumerate}


\item $L(t)=\bot$ if and only if $t=\varepsilon$;

\item if $t$ has predecessor $t'$, then there is no other child $t''$
  of $t'$ such that $\mn{adom}(L_1(t))=\mn{adom}(L_1(t''))$ and
  $L_2(t)=L_2(t'')$; 

\item if $t$ has predecessor $t' \neq \varepsilon$, then $L_2(t') \subseteq \mn{adom}(L_1(t))$;

\item if $t$ has predecessor $t' \neq \varepsilon$, then $L_1(t)|_S = L_1(t')|_S$, $S =
  \mn{adom}(L_1(t)) \cap \mn{adom}(L_1(t'))$;

\item if $L_1(t) \cap \{1,2,3\} \neq \emptyset$,
then $|\mn{adom}(L_1(t))| \leq 2$ and for all children $t'$ of~$t$,
$\mn{adom}(L_1(t')) \cap \mn{adom}(\Dmc) = \emptyset$.

\end{enumerate}
Formally, every proper $\Gamma$-labeled tree $(T,L)$ encodes an
interpretation $\Imc_{(T,L)}$, as follows. First choose a function
$\mu$ that maps each pair $(t,a)$ with $t \in T \setminus
\{\varepsilon\}$ and $a \in \mn{adom}(L_1(t))$ to a constant $\mu(t,a)$
such that:
\begin{enumerate}

\item if $t$ has predecessor $\varepsilon$, then
  \begin{enumerate}

  \item if $q$ is unary, $\bar a = a$, $\mn{adom}(L_1(t))=\{a\}$, and
    $L_2(t)=\emptyset$,
    then $\mu(t,a)=a$;\footnote{Here we assume w.l.o.g.\ that the copy
      of $a \in \mn{adom}(\Dmc)$ in $\Dmc^\approx_{\ell,k}$ that the
      $\langle \cdot \rangle$ operation renames back to $a$ is the
      copy induced by the $\ell,k$-sequence $\{a\}$, c.f.\ the
      definition of $\ell,k$-unravelings.}

    \item otherwise, $\mu(t,a)$ is a fresh constant.

  \end{enumerate}

\item if $t$ has predecessor $t'$ and $a \in \mn{adom}(L_2(t))$, then
  $\mu(t,a)=\mu(t',a)$;

\item if $a \notin \mn{adom}(L_2(t))$,
  then $\mu(t,a)$ is a fresh constant.

\end{enumerate}
Then set
$$
\begin{array}{rcl}
  \Delta^{\Imc_{(T,L)}} &=& \{ \mu(t,a) \mid t \in T,\ a \in
  \mn{adom}(L_1(t)) \} \\[1mm]
  A^{\Imc_{(T,L)}} &=& \{ \mu(t,a) \mid t \in T,\ A(a) \in 
L_1(t) \} \\[1mm]
  r^{\Imc_{(T,L)}} &=& \{ (\mu(t,a),\mu(t,b)) \mid t \in T,\ r(a,b) \in 
L_1(t) \}
\end{array}
$$
for all concept names $A$ and role names $r$. The \emph{database part}
of $\Imc_{(T,L)}$ is the restriction of $\Imc_{(T,L)}$ to domain $\{
\mu(x,a) \mid x \in T \text{ and } a \in \mn{adom}(\Dmc) \}$ while the
\emph{existential part} is the restriction to domain $\{ \mu(x,a) \mid
x \in T \text{ and } a \in \{ 1,2,3 \}\}$. It is clear by construction
that the database part of $\Imc_{(T,L)}$ has treewidth 
$(\ell,k)$ while the existential part has treewidth 
$(1,2)$. Moreover, if $\Imc$ is a witness for $\bar a \notin
Q(\Dmc^\approx_{\bar a, \ell, k})$ in which the tree parts have
outdegree at most $|\Omc|$, then we find a $\Gamma$-labeled tree
$(T,L)$ such that $\Imc_{(T,L)}$ is isomorphic to \Imc and the
database part of $\Imc_{(T,L)}$ is identical to the restriction of
\Imc to domain $\mn{adom}(\langle\Dmc^\approx_{ \ell, k},\bar a
\rangle)$, up to renaming constants that are not in $\bar a$.

\smallskip

The TWAPA \Amf will be obtained as the intersection of four TWAPA,
$\Amf_\mn{proper}$, $\Amf_1$, $\Amf_2$ and $\Amf_3$. While
$\Amf_\mn{proper}$ makes sure that the input tree $(T,L)$ is proper,
$\Amf_1$ checks that, up to renaming constants that are not in $\bar
a$, $\Imc_{(T,L)}$ is a model of $\langle\Dmc^\approx_{ \ell, k},\bar
a \rangle$, $\Amf_2$ checks that $\Imc_{(T,L)}$ is a model of \Omc,
and $\Amf_3$ checks that $\Imc_{(T,L)} \not\models q(\bar a)$. In
summary, we thus have $L(\Amf) \neq \emptyset$ iff there is a witness
for $\bar a \notin Q(\langle \Dmc^\approx_{\ell, k},\bar a
\rangle)$. Regarding $\Amf_1$ note that the renaming of constants is
unproblematic because $\langle \Dmc^\approx_{\ell, k},\bar a \rangle
\models Q(\bar a)$ iff $\Emc \models Q(\bar a)$, for all databases
\Emc obtained from $\langle \Dmc^\approx_{\ell, k},\bar a \rangle$ by
renaming constants that are not in $\bar a$. The TWAPA
$\Amf_\mn{proper}$ is very easy to design, we leave details to the
reader. A sketch of the construction of the other TWAPAs is provided
below.

\smallskip
\noindent
\textbf{Details for $\Amf_1$.} This
TWAPA 
has states $s|_I$ and $s_{I,O}$ for every $I \subseteq
\mn{adom}(\Dmc)$ with $1 \leq |I| \leq k$ and every $O \subseteq
I$. When in state $s_I$ or $s_{I,O}$, the TWAPA verifies that the
current symbol encodes an interpretation that is a model of
$\Dmc|_{I}$. There is an additional starting state $s_0$ from which
the TWAPA transitions to every child, making sure that for every $I
\subseteq \mn{adom}(\Dmc)$ with $1 \leq |I| \leq k$, some child is
visited in state $s_I$. From state $s_I$ or $s_{I,O}$, the TWAPA again
transitions to every child, making sure that for every $J \subseteq
\mn{adom}(\Dmc)$ with $1 \leq |J| \leq k$ and every $O \subseteq J$,
some child is visited in state $s_{J,O}$. The number of states needed
for this TWAPA is $O(|\mn{adom}(\Dmc)|^k \cdot 2^k)$ and thus
polynomial since $k$ is fixed.

\smallskip
\noindent
\textbf{Details for
  $\Amf_2$.} 
We can assume $\Omc$ to be of the form $\{\top \sqsubseteq C_\Omc\}$,
where $C_\Omc$ is in negation normal form, that is, negation is
applied only to concept names, but not to compound concepts. The TWAPA
visits all nodes $x$ in the input tree, considers every constant $a$
in $L_1(x)$, and verifies that $\mu(x,a)$ satisfies $C_\Omc$ in
$\Imc_{(T,L)}$. This can be done using states of the form $s_{C,a}$,
with $C \in \mn{sub}(C_\Omc)$ and $a \in N$, where the purpose of
$s_{C,a}$ is to verify that $\mu(x,a) \in C^{\Imc_{(T,L)}}$ for the
current node $x$. When in state $s_{C_1 \sqcap C_2, a}$, for instance,
the automaton stays at the same node and transitions to both states
$s_{C_1, a}$ and $s_{C_2, a}$. The only non-trivial cases are concepts
of the form $\exists r.C$ and $\forall r . C$ since $r$-successors of
$\mu(x,a)$ in $\Imc_{(T,L)}$ can be generated by constants in some
$L_1(y)$ with $y \neq x$. In the $\forall r . C$ case, the automaton
considers all $r$-successors of $a$ in $L_1(x)$ and then moves to all
children $y$ of $x$ with $a \in L_2(y)$ and to the parent $y$ of $x$
if $a \in L_2(x)$ and considers all $r$-successors of $a$ in
$L_1(y)$. Concepts $\exists r . C$ are dealt with in a dual
way. Instead of sending copies of itself to all neighbors, the TWAPA
guesses a neighbor to send a copy of itself to. The parity condition
is used to ensure that the TWAPA eventually arives at a node in
which $\exists r . C$ is satisfied for $a$. The number of states is
$O(|\Omc|\cdot k)$.

\smallskip
\noindent
\textbf{Details for $\Amf_3$.} Since TWAPAs can be complemented
without increasing the number of states, it suffices to construct a
TWAPA that checks that $\Imc_{(T,L)} \models q(\bar a)$. 
We characterize the existence of a homomorphism $h$ from $q$ to
$\Imc_{(T,L)}$ with $h(\bar x)=\bar a$ by the existence of a match
witness, as defined below.

Let $(T,L)$ be a proper $\Gamma$-labeled tree. We say that nodes
$t_1,t_2 \in T$ are \emph{$a$-connected}, $a$ a constant from $N$, if
there is a path $s_1,\dots,s_n$ in $T$ with $t_1=s_1$, $t_2=s_n$, and
such that for $1 \leq i < n$, either $s_{i+1}$ is a successor of $s_i$
and $a \in L_2(s_{i+1})$ or $s_{i+1}$ is the predecessor of $s_i$
and $a \in L_2(s_i)$. Fix a $(1,k')$-tree
decomposition $T_q = (V_q, E_q, B_q)$ of $q$. A \emph{match witness for
  $q$ in $(T,L)$} is a triple $(D,P,H)$ where
\begin{itemize}

\item $D$ assigns to each $v \in V_q$ a tree decomposition
  $D(v)=(V_v,E_v,B_v)$ of $q|_{B_q(v)}$;

\item $P$ assigns to each $v \in V_q$ and $\widehat v \in V_v$
  a node $P(v,\widehat v)$ from the input tree $T$;


\item $H$ assigns to each $v \in V_q$ and $\widehat v \in V_v$
  a  homomorphism $H(v,\widehat v)$ from $q|_{B_v(\widehat v)}$ to 
  $L_1(P(v,\widehat v))$. 


\end{itemize}
such that the following conditions are satisfied:
\begin{enumerate}

\item for all $u,v \in V_q$, $\widehat u \in V_u$, $\widehat v \in
  V_v$, and $x \in B_u(\widehat u) \cap B_v(\widehat v)$,
$H(u,\widehat u)(x) =
   H(v,\widehat v)(x)$ and 
$P(u,\widehat u)$ and 
$P(v,\widehat v)$ are $H(u,\widehat u)(x)$-connected in~$T$;



\item if $q$ is unary with answer variable $x$ and $\bar a = a$, then
  for all $v \in V_q$ and $\widehat v \in V_v$ such that $x \in
  B_v(\widehat v)$, $H(v,\widehat v)(x) = a$ and $P(v,\widehat
  v)$ is $a$-connected to the unique successor $t$ of $\varepsilon$
  with $\mn{adom}(L_1(t))=\{a\}$ and $L_2(t)=\emptyset$ (please see Condition 1(a) in the
  definition of $\mu(t,a)$).

\end{enumerate}
\begin{lemma}
\label{lem:matchwitness}
Let $(T,L)$ be a proper $\Gamma$-labeled tree.
There is a match witness for $q$ in $(T,L)$
if and only if $\Imc_{(T,L)} \models q(\bar a)$.
\end{lemma}
\noindent
\begin{proof}\ ``$\Rightarrow$''. Let $(D, P, H)$ be a
  match witness for $q$ in $(T, L)$.  To show
  that $\Imc_{(T,L)} \models q(\bar a)$, we construct a homomorphism
  $h$ from $q$ to $\Imc_{(T,L)}$ with $h(\bar x)=\bar a$. Let $x$ be a
  variable in $q$. Then there is a $v \in V_q$ and a $\widehat v \in V_v$
  such that $x \in B_v(\widehat v)$. Set $h(x)$ to be the element of
  $\Delta^{\Imc_{(T,L)}}$ generated by $H(v, \widehat v)(x)$, that is,
  $\mu(P(v, \widehat v),H(v, \widehat v)(x))$.  
  Due to the connectedness
  condition of tree decompositions and by Condition~1, the choices of
  $v$ and $\widehat v$ do not matter and $h$ is well-defined.
  Moreover, $h$ is clearly
  a homomorphism from $q$ to $\Imc_{(T,L)}$ since each $H(v, \widehat v)$
  is a homomorphism from $q|_{B_v(\widehat v)}$ to 
  $L_1(P(v,\widehat v))$. Finally $h(\bar x)=\bar a$ due to Condition~2.

  \smallskip

  ``$\Leftarrow$''. Let $\Imc_{(T,L)} \models q(\bar a)$ and let $h$
  be a homomorphism from $q$ to $\Imc_{(T,L)}$ with $h(\bar x)=\bar
  a$. We define the match witness $(D, P, H)$ as follows.
  Let $v \in V_q$. For every node $t$ in the input tree, let $W_{v,t}$
  denote the set of all variables $x \in B_q(v)$ such that for some
  constant $a$ that occurs in $L_1(t)$, $h(x)$ is the element of
  $\Delta^{\Imc_{(T,L)}}$ generated by $a$ in~$t$, that is,
  $h(x)=\mu(t,a)$ where $\mu$ is the function from the definition
  of $\Imc_{(T,L)}$.
  \begin{itemize}

  \item $V_v$ is the $\subseteq$-minimal prefix-closed subset of $T$
    such that for all $x,y \in B_q(v)$ that cooccur in an atom in $q$,
    there is a $t \in V_v$ with $\{x,y\} \subseteq W_{v,t}$.

  \item $E_v = \{ (t, t \cdot i) \mid t \cdot i \in V_v, i \in \{1, \ldots, m\} \}$;

  \item $B_v(\widehat v)=W_{v,\widehat v}$;

  \item $P(v, \widehat v) := \widehat v$, and

  \item $H(v, \widehat v)(x)$ is defined to be the constant that
    generates $h(x)$, that is, $\mu(\widehat v,x)$, for all $x \in
    W_{v,\widehat v}$.

  \end{itemize}
  It follows from the definition of $\Imc_{(T,L)}$ that $(V_v, E_v,
  B_v)$ is a tree decomposition of $q|_{B_q(v)}$.  Using that $h$ is a
  homomorphism and Condition~3 of proper input trees, it can further
  be shown that $H(v,\widehat v)$ is a homomorphism from
  $q|_{B_v(\widehat v)}$ to $L_1(P(v,\widehat v))$, as
  required. Moreover, Conditions~1 and~2 of match witnesses are
  clearly satisfied.
\end{proof}
Note that the proof of Lemma~\ref{lem:matchwitness} does not rely on the
fact that $q$ has treewidth $(1,k')$, so the lemma holds even
if $T_q$ is an $(\ell',k')$-tree decomposition with $\ell'>1$.
However, we aim to show that the existence of a match tree can be
checked using a TWAPA $\Amf_3$ with polynomially many states, 
and this is where the assumption $\ell'=1$ is needed.

The general idea is that $\Amf_3$ processes the bags of $T_q$ in a
top-down manner, always storing the current node $v \in V_q$ in its
state.  For every $v \in V_q$, it verifies the existence of a tree
decomposition $D(v)$ of $q|_{B_q(v)}$, also in a top-down
manner. Note, however, that $D(v)$ is not fixed like $T_q$, but rather
the automaton needs to construct it, that is, we must be able to
extract the decompositions $D(v)$ from a successful run. While
verifying that the required $D(v)$ exists, the automaton also verifies
the existence of suitable $P$ and $H$ components of the match
witness. When considering a node $\widehat v$ of decomposition $D(v)$,
it moves to the node in the input tree that is $P(v,\widehat v)$ and 
then checks that the required homomorphism $H(v,\widehat v)$ exists.

We use states of the form $s_{v, M,h}$, where 
\begin{itemize}

\item $v \in V_q$ is the current node of $T_q$, 

\item $M : B_q(v) \rightarrow \{\mn{here}, \mn{below}, \mn{else} \}$
  records which variables from $B_q(v)$ are part of the bag of $D(v)$
  that is currently being treated (`\mn{here}'), which are not in the
  current bag, but in a bag strictly below (`\mn{below}'), and which
  are only part of bags that are neither the current one nor below it
  (`\mn{else}'), and

\item $h$ is a homomorphism from $q|_{\{x \in B_q(v) \mid
    M(x)=\mn{here}\}}$ to the database $L_1(t)$ associated with the current node
  $t$ of the input tree.

\end{itemize}
Note that the number of variables in $B_q(v)$ is bounded by $k'$, and
thus the number of possible functions $M$ is polynomial.  Including
some book keeping states, in fact, the number of required states is
$O(|q|^2\cdot 3^{k'} \cdot k^{k'})$.

$\Amf_3$ send a copy of itself to some node of the input tree and then
starts in state $s_{v,M,h}$ where $v$ is the root of $T_q$, $h$ the
empty homomorphism, and $M$ maps every variable to $\mn{here}$ or
$\mn{below}$.  Now assume it is in some state $s_{v,M,h}$ which
(implicitly) represents some node $\widehat v$ of $D(v)$. It then
spawns $n \leq |B_q(v)|$ copies of itself, corresponding to children
$\widehat v_1, \ldots, \widehat v_n$ of $\widehat v$ in $D(v)$, in
states $s_{v,M_1,h_1},\dots,s_{v,M_n,h_n}$. In the $M_i$, each
variable that was \mn{here} before is now \mn{here} or \mn{else} and
each variable that was \mn{else} before is still \mn{else}. Moreover,
each variable $x$ that was \mn{below} before is now \mn{here} or
\mn{below} or \mn{else}, but is \mn{here} or \mn{below} in exactly one
$M_i$. What is more, $h_i$ agrees with $h$ on the intersection of
their domains. The latter partly achieves Condition~1 from the
definition of match witnesses. The automaton then moves each of the
$n$ copies to a node in the input tree, corresponding to the choice of
$P(\widehat v_i)$. To make sure that it eventually reaches that node,
we use the parity condition. Moreover, we make sure that for every
variable $x$ that is in the domain of both $h$ and $h_i$, the path is
$h(x)$-connected. This fully achieves Condition~1.

The TWAPA might need to spawn additional copies to transition to
successors of $v$ in the `outer' tree decomposition $T_q$. For
successors $u$ of $v$ such that $B_q(u) \cap B_q(v) = \emptyset$, this
is done as soon as the processing of $v$ starts. For successors $u$ of
$v$ such that $B_q(u) \cap B_q(v) = \{ x \}$, this is done the first
time that $M(x)$ is set to \mn{here}.  If the current state is
$s_{v,h,M}$, we then spawn the new copy in a state $s_{u,h',M'}$,
where $h'$ is the homomorphism that maps only $x$ to $h(x)$ and $M'$
sets $x$ to $\mn{here}$ and all other variables from $B_q(u)$ to
$\mn{below}$. We thus start the top-down verification of $D(u)$ at a
bag that contains the node $x$ that $B_q(u)$ shares with its
predecessor bag $B_q(v)$ in $T_q$. Note that this part fails for 
$\ell'>1$ since then $B_q(u)$ and $B_q(v)$ can share two variables
$x_1,x_2$ and there is no guarantee that there is a bag of $D(v)$
in which both $x_1$ and $x_2$ are mapped to \mn{here}.

Finally, we also need to verify Condition~2 from the definition of
match witnesses. This is easy. If $q$ has answer variable $x$ and
$M(x) = \mn{here}$, then $\Amf_3$ makes sure that $h(x)= a$ (with
$\bar a =a$) and it spawns a copy that verifies that the current node
is $\bar a$-connected to the (unique) successor $t$ of $\varepsilon$ with
$\mn{adom}(L_1(t))=\{a\}$ and $L_2(t)=\emptyset$.


\section{Proofs for Section~\ref{sect:btw}}

\lemunravfundtwo*

\noindent
\begin{proof} \
  We generalize the proof of Lemma~\ref{lem:unravfund}.
  For the ``$\supseteq$'' direction, observe that for every finite subset $\Dmc'$ of $\Dmc^\approx_{\bar a,\ell,k}$ containing $\bar a$ the pointed database $(\Dmc',\bar a)$ is a member of $\Dmf_{\ell,k}$. We also have a homomorphism from $\Dmc'$ to $\Dmc$ mapping $\bar a$ to $\bar a$. Thus, this direction is again a consequence of compactness. 
  
  For the ``$\subseteq$'' direction, assume that there is
  a pointed database $(\Dmc',\bar b)$ of treewidth 
  $(\ell,k)$, $\bar b \in Q(\Dmc')$, and there is a homomorphism
  $h$ from $\Dmc'$ to \Dmc such that $h(\bar b)=\bar a$. To show that
  $\bar a \in Q(\Dmc_{\bar a,\ell,k}^\approx)$, it suffices to construct a
  homomorphism $g$ from $\Dmc'$ to $\Dmc_{\bar a,\ell, k}^\approx$ with 
  $g(\bar b)=\bar a$ and to recall that answers to OMQs are preserved under taking
  homomorphic images of databases. But the existence of such a homomorphism 
  is a consequence of Lemma~\ref{lem:homintree}.
  \end{proof}

\propeliqbtw*

\noindent
\begin{proof}\ 
	For the proof of Point~1, assume that $Q$ is an OMQ in
	$(\ALCI,\text{UCQ}^{\mn{tw}}_{\ell,k})$. Assume that 
	$\bar a \in Q^\approx_{\ell,k,\ell',k'}(\Dmc)$. Then, by Lemma~\ref{lem:untol2}, 	$\bar a \in Q^\approx_{\ell,k,\ell',k'}(\Dmc)$ iff $\bar a \in Q^\approx_{\ell,k,\ell',k'}(\Dmc_{\bar a,\ell,k}^{\approx})$. Thus, 
	$\bar a \in Q(\Dmc_{\bar a, \ell,k}^{\approx})$ since $\Omc \models \Omc^{\approx}_{\ell,k,\ell',k'}$.
	
	For the proof of Point~2, assume that $Q=(\Omc,\Sigma,q(\bar x))$ is an OMQ in $(\ALCI,\text{CQ}^{\mn{tw}}_{\ell',k'})$ of arity at most $r\leq 1$. 
	Let $\bar a \in Q(\Dmc_{\bar a, \ell,k}^{\approx})$.
	Compactness yields a
	finite subset $\Fmc$ of $\Dmc^\approx_{\bar a,\ell,k}$ with $\Fmc,\Omc
	\models q(\bar a)$. We can view \Fmc as a CQ $p(\bar x)$ of
	treewidth $(\ell+r,k+r)$ where $\bar x$ corresponds to $\bar a$. Then $\Omc \models p(\bar x) \rightarrow q(\bar x)$ and thus $p(\bar x) \rightarrow q(\bar x)$ is a TGD in
	$\Omc^\approx_{\ell+r,k+r,l',k'}$. Consequently, $\bar a \in Q^\approx_{\ell+r,k+r,\ell',k'}(\Dmc)$, as required.
\end{proof}

We prove the upper bounds in Theorem~\ref{thm:btwunravmain}.
\thmbtwunravmain*

%
We prove the upper bounds in Theorem~\ref{thm:btwunravmain} using an
elimination algorithm that is inspired by algorithms for achieving
$k$-consistency of constraint satisfaction problems. It also bears
similarity to type elimination in modal logic. 
We start with Point~2 and then proceed to Point~1. For Point~2, we
consider a slight generalization in preparation of the proof of
Point~1 later on. By Lemma~\ref{lem:unravfundtwo}, deciding
$\bar a \in \mn{app}_{\Dmf_{\ell,k}}(\Dmc)$ amounts to deciding
$\bar a \in Q(\Dmc^\approx_{\bar a,\ell,k})$.  Based on this, the
announced generalization of Point~2 is
as follows.
\begin{theorem}
\label{thm:newkidontheblock}
  Fix $\ell,k \geq 1$ with $\ell < k$.  Given an OMQ
  $Q(\bar x)=(\Omc,\Sigma,q) \in (\ALCI,\text{bELIQ})$, a
  $\Sigma$-database \Dmc, $S \subseteq \mn{adom}(\Dmc)$, and
  $\bar a \in \mn{adom}(\Dmc)^{|\bar x|}$, it is fixed-parameter
  tractable with single exponential running time to decide whether
  $\bar a \in Q(\Dmc^\approx_{S,\ell,k})$.
\end{theorem}
To prove Theorem~\ref{thm:newkidontheblock}, fix $\ell,k \geq 1$ with
$\ell < k$.  Let $Q(\bar x)=(\Omc,\Sigma,q) \in (\ALCI,\text{bELIQ})$,
\Dmc be a $\Sigma$-database, $S \subseteq \mn{adom}(\Dmc)$, and
$\bar a \in \mn{adom}(\Dmc)^{|\bar x|}$. Before we formulate the
algorithm, we introduce several notions that it uses. We use
$\Cl(\Omc,q)$ to denote the smallest set that contains all concepts in
\Omc and the bELIQ $q$, viewed as an $\ELI^u$-concept, and that is
closed under subconcepts and single negation.


For a set $L \subseteq \mn{adom}(\Dmc)$ with $S \subseteq L$, an
\emph{$L$-assignment} $\mu$ is a set that contains exactly one of
$C(a), \neg C(a)$ for all $\neg C \in \cl(\Omc,q)$ and $a \in L$,
subject to the following conditions:
\begin{enumerate}

\item if $q$ is an ELIQ and $\bar a = a_0$, then $\neg q(a_0) \in
  L$;

\item if $q$ is a BELIQ, then $\neg q(a) \in L$ for all $a \in L$.
  
\end{enumerate}
We shall be interested in $L$-assignments with $L \subseteq
\mn{adom}(\Dmc)$, $S \subseteq L$, and $|L\setminus S| \leq
\ell$. When speaking of sets $L \subseteq \mn{adom}(\Dmc)$, we from now
on silently assume that $L$ satisfies the latter two  conditions.

The $L$-assignment $\mn{as}_\Imc(L)$ \emph{realized} in a model \Imc
of \Omc with $L \subseteq \Delta^\Imc$ is the set that contains $C(a)$
for all $C \in \cl(\Omc,q)$ and $a \in L$ with $a \in C^\Imc$.

A \emph{$k$-subset} of $\mn{adom}(\Dmc)$ is a set
$K \subseteq \mn{adom}(\Dmc)$ such that $S \subseteq K$ and
$|K \setminus S| \leq k$. 
Let $\Gamma$ be a set of $L$-assignments and let $K$ be a $k$-subset.
A \emph{$\Gamma$-choice} for $K$ is a subset
$\Gamma' \subseteq \Gamma$ that contains exactly one $L$-assignment
for each $L \subseteq K$. We say that $\Gamma'$ is \emph{consistent}
if the extended database $$\Dmc|_K \cup \{ C(a) \mid \exists \mu \in \Gamma': C(a) \in \mu
  \}$$ is satisfiable with the $\ALCI$-ontology
  $ \Omc \cup \{ \top \sqsubseteq C \}$ where $C=D$ if
  $q=\exists u . D(x)$ is a BELIQ and $C=\top$ otherwise.
  
  We are now ready to describe the algorithm. It computes a sequence
  $\Gamma_0,\Gamma_1,\dots$ of sets of $L$-assignments, starting with
  the set $\Gamma_0$ of all $L$-assignments for any $L \subseteq
  K$. When constructing $\Gamma_{i+1}$ from $\Gamma_i$, it considers
  all $L$-assignments $\mu \in \Gamma_i$ and checks whether for all
  $k$-subsets~$K \supseteq L$, there is a consistent $\Gamma_i$-choice
  $\Gamma$ for $K$ such that $\mu \in \Gamma$. If this is the case,
  then $\mu$ remains in $\Gamma_{i+1}$. Otherwise, it is removed. Let
  $f \geq 0$ be smallest such that $\Gamma_f=\Gamma_{f+1}$. The
  algorithm answers `yes' if $\Gamma_f$ is empty and `no' otherwise.

Before we prove correctness, let us analyze the running time of the
algorithm. We start with the number of iterations. The number of sets
$L \subseteq \mn{adom}(\Dmc)$ with $|L| \leq \ell$ is bounded by
$(||\Dmc||+1)^\ell$, thus polynomial in $||\Dmc||$. For each such~$L$,
the number of $L$-assignments is in $2^{O(||Q||^2)}$. Since at least
one $L$-assignment is eliminated in each iteration until we reach
$\Gamma_f$, the number of iterations is bounded by
$2^{O(||Q||^2)} \cdot (||\Dmc||+1)^\ell$. It remains to consider a
single iteration. The algorithm goes through all sets $L$ and all
$L$-assignments $\mu$, thus considers
$2^{O(||Q||^2)} \cdot (||\Dmc||+1)^\ell$ assignments.  It then
considers all $k$-subsets $K \supseteq L$ of which there are at most
$(||\Dmc||-1)^k$ many. Next, it goes through all $\Gamma_i$-choices
for $K$. As there are only $(k+1)^\ell$ sets $L \subseteq K$ with
$|L| \leq \ell$, the number of such choices is also bounded single
exponentially in $||\Omc||+||q||$ (with $(k+1)^\ell$ appearing as a
constant in the exponent). Overall, we achieve fixed-parameter
tractability with single exponential running time, as desired.
\begin{lemma}
  The algorithm answers 'yes' iff 
  $\Dmc^\approx_{S,\ell,k} \models Q(\bar a)$.
\end{lemma}
\noindent
\begin{proof}\
  For the `only if' direction, assume $\Dmc_{S, \ell, k}^\approx \not\models Q(\bar a)$.
  We need to show that the algorithm answers `no'.
  Since $\Dmc_{S, \ell, k}^\approx \not \models Q(\bar a)$, there is a model $\Imc$
  of $\Omc$ and $\Dmc_{S, \ell, k}^\approx$ with $\Imc \not \models q(\bar a)$.
  For every $b \in \mn{adom}(\Dmc_{S, \ell, k}^\approx)$,
  define $\mn{tp}(b) = \{C \in \mn{cl}(\Omc, q) \mid b \in C^\Imc\}$.
  For every $\ell,k$-sequence $v=S_0, O_0, \ldots, O_{n-1}, S_n$ with $n \geq 1$
  we define $L_v \subseteq \mn{adom}(\Dmc)$ as $h(O_{n-1})$, where
  $h : \mn{adom}(\Dmc_{S, \ell, k}^\approx) \rightarrow \mn{adom}(\Dmc)$
  is the uncopying map, and we define the $L_v$-assignment
  $\mu_v = \{C(a) \mid C \in \mn{tp}(a_v), a \in L_v\}$.
  Since $\Imc$ is a model
  of $\Omc$ and $\Dmc_{S, \ell, k}^\approx$ with $\Imc \not \models q(\bar a)$,
  $\mu_v$ is indeed satisfies Conditions~1 and~2 of $L_v$-assignments.
  Let $\Psi = \{\mu_v \mid v \text{ is an } \ell,k\text{-sequence of length } n \geq 1\}$.
  We argue by induction on $i$ that $\Psi \subseteq \Gamma_i$
  for every $i \geq 0$. 
  Since $\Gamma_0$ is the set of all $L$-assignments, $\Psi \subseteq \Gamma_0$.
  For the induction step, assume $\Psi \subseteq \Gamma_i$ for some $i \geq 0$.
  Let $\mu_v \in \Psi$ be an $L_v$-assignment for some
  $v=S_0, O_0, \ldots, O_{n-1}, S_n$.
  We need to show that for every $k$-subset $K$ with $K \supseteq L_v$
  there is a consistent $\Gamma_i$-choice $\Gamma$ for $K$ such that $\mu_v \in \Gamma$.
  Let $K$ be a $k$-subset.
  For every $L \subseteq K$, choose a set $S_{n+1} \supseteq L$ with
  $|S_{n+1} \setminus S| \leq k$ and consider the sequence
  $v_{K,L}=S_0, O_0, \ldots, O_{n-1}, K, L, S_{n+1}$. This is
  an $\ell,k$-sequence that yields an $L$-assignment $\mu_{v_{K,L}} \in \Psi$.
  The set of all such $L$-assignments forms a $\Gamma_i$-choice $\Gamma$.
  Since $\Imc$ is a model that realizes all $L$-assignments from $\Gamma$,
  $\Gamma$ is consistent, so $\mu_v$ remains in $\Gamma_{i+1}$.
  Since $\Psi$ is not empty, this implies that the algorithm answers `no'.
  
  For the `if' direction, assume the algorithm answers `no'. We need to show that
  $\Dmc_{S, \ell, k}^\approx \not\models Q(\bar a)$.
  Since the algorithm answers `no', the set $\Gamma := \Gamma_f$ is not empty.
  This implies that $\Gamma$ contains at least one $L$-assignment for every set
  $L \subseteq \mn{adom}(\Dmc)$ with $S \subseteq L$ and $|L \setminus S| \leq \ell$,
  since as soon as there is an $L_0$ such that some $\Gamma_i$
  does not contain an $L_0$-assignment, the algorithm eventually
  eliminates all $L$-assignments and thus returns `yes'.
  Thus, for every $L$-assignment $\mu \in \Gamma$ and every $k$-subset $K$,
  there is a consistent $\Gamma$-choice $\Gamma_{\mu,K}$ for $K$
  such that $\mu \in \Gamma_{\mu,K}$.
  We construct a model $\Imc$ of $\Dmc_{S, \ell, k}^\approx$ and $\Omc$
  with $\Imc \not \models q(\bar a)$.
  First, we define a function
  $t : \mn{adom}(\Dmc_{S, \ell, k}^\approx) \rightarrow 2^{\mn{cl}(\Omc, q)}$
  that assigns a type $t(a)$ to every $a \in \mn{adom}(\Dmc_{S, \ell, k}^\approx)$
  and in a second step, we construct the model $\Imc$ using $t$.
  
  We start by defining $t$ on $S$. Choose any $S$-assignment $\mu_0 \in \Gamma$
  and define $t(a) = \{C \in \mn{cl}(\Omc, q) \mid C(a) \in \mu_0\}$
  for every $a \in S$. We proceed to define $t$ inductively
  on the length
  of the $\ell,k$-sequence in which the element appears first.
  For sequences of length $0$, let $v=S_0$ be an $\ell,k$-sequence.
  For every $a_v \in \mn{adom}(\Dmc_{S, \ell, k}^\approx)$ that is a copy of
  $a \in S_0 \setminus S$, define $t(a_v)$ according to  
  $\Gamma_{\mu_0, S_0}$, i.e. choose any $\mu \in \Gamma_{\mu_0, S_0}$
  that is defined on $a$
  and set $t(a_v) = \{C \in \mn{cl}(\Omc, q) \mid C(a) \in \mu\}$.
  If $t$ has already been defined for all elements that appear in $\ell,k$-sequences
  of length at most $i$, let $v=S_0, O_0, S_1, \ldots, S_i, O_i, S_{i+1}$.
  Since $O_i \subseteq S_i$, $t$ has already been defined
  (by means of an $O_i$-assignment $\mu$)  
  on elements of
  $\mn{adom}(\Dmc_{S, \ell, k}^\approx)$ that are copies of elements of
  $O_i$.
  For every $a_v \in \mn{adom}(\Dmc_{S, \ell, k}^\approx)$ that is a copy of an
  $a \in S_{i+1} \setminus S_i$ define $t(a_v)$ according to  
  $\Gamma_{\mu, S_{i+1}}$, i.e. choose any $\mu' \in \Gamma_{\mu, S_{i+1}}$
  that is defined on $a$
  and set $t(a_v) = \{C \in \mn{cl}(\Omc, q) \mid C(a) \in \mu'\}$.
  This finishes the definition of $t$.
  
  Now we construct the model $\Imc$ of $\Dmc_{S, \ell, k}^\approx$ and $\Omc$
  with $\Imc \not \models q(\bar a)$. For every $a \in \mn{adom}(\Dmc_{S, \ell, k})$,
  choose a tree-shaped model $\Imc_a$ of $t(a)$. The interpretation $\Imc$ is defined
  as follows: Start from $\Dmc_{S, \ell, k}^\approx$ seen as an interpretation
  and for every $a \in \mn{adom}(\Dmc_{S, \ell, k})$, add a copy of $\Imc_a$ and
  identify $a$ with the root of $\Imc_a$. It can be verified that $\Imc$
  is indeed a model of $\Omc$
  with $\Imc \not \models q(\bar a)$.
  
\end{proof}
We now turn to Point~1 of Theorem~\ref{thm:btwunravmain}.
Fix $\ell,k \geq 1$ with $\ell < k$. Let
$Q(\bar x)=(\Omc,\Sigma,q) \in (\ALCI,\text{UCQ})$, \Dmc be a
$\Sigma$-database, and $\bar a \in \mn{adom}(\Dmc)^{|\bar x|}$. By
Lemma~\ref{lem:unravfundtwo}, it suffices to decide whether
$\bar a \in Q(\Dmc^\approx_{\bar a,\ell,k})$.  To reduce complexity,
we first replace $Q$ with a set of Boolean OMQs, similar to what was
done in the proof of Theorem~\ref{thm:treeunravmain}.

We decorate \Dmc and decorate as well as simplify $q$.  Let
$\bar x = x_0 \cdots x_{n-1}$ and $\bar a = a_0 \cdots a_{n-1}$.  For
each (possibly inverted) role name $r$ used in $q$ and all $i < n$,
introduce a fresh concept name $A_{r,a_i}$ that, informally, expresses
the existence of an $r$-edge to $a_i$. Extend \Dmc to a database
$\Dmc^d$ by adding $A_{r,a_i}(b)$ for all $r(b,a_i) \in \Dmc$. This is
clearly possible in time $O(||\Dmc|| \cdot ||q||^2)$. Further, let $q^d$ be
obtained from $q^c$ by doing the following for each CQ $p$ in $q^c$:
\begin{enumerate}

\item replace each atom $r(y,a_i)$, $y$ a quantified variable, with
  $A_{r,a_i}(y)$;

\item for each atom $r(x_{i_1},x_{i_2})$, $0 \leq i_1,i_2 < n$, check
  whether $r(a_{i_1},a_{i_2}) \in \Dmc$; if the check fails, remove
  $p$ from the UCQ; if it succeeds, remove atom $r(x_{i_1},x_{i_2})$
  from $p$;

\item for each atom $A(x_i)$, $0 \leq i < n$, check whether
  $\Dmc^\approx_{\bar a,\ell,k},\Omc\models A(a_i)$ using
  Theorem~\ref{thm:newkidontheblock}
  with $S=\bar a$; if the check fails, remove $p$ from the
  UCQ; if it succeeds, remove atom $A(x_{i})$ from $p$.
    
\end{enumerate}
If some of the CQs in $q^c$ becomes empty in the process (that is, all
of its atoms have been dropped), then we know that
$\bar a \in Q^c(\Dmc^\approx_{\bar a,\ell,k})$ and return `yes'. If all CQs
have been dropped, then we know that
$\bar a \notin Q^c(\Dmc^\approx_{\bar a,\ell,k})$ and return `no'. Clearly
all of the above can be done in time $2^{O(||Q||^2)} \cdot O(||\Dmc||)$.
Note that $q^d$ contains no answer variables as all atoms that
mention them have been dropped.
It is not hard to show the following using some basic manipulations of
homomorphisms that witness query answers.
\begin{lemma}
  $\bar a \in Q^c(\Dmc^\approx_{\bar a,\ell,k})$ iff
  $(\Dmc^d)^\approx_{\bar a,\ell,k},\Omc \models q^d$.
\end{lemma}
It thus remains to decide whether
$(\Dmc^d)^\approx_{\bar a,\ell,k},\Omc \models q^d$. 
We may view $q^d$ as a Boolean UCQ in which each CQ is a disjoint
union of connected CQs, that is, a
disjunction of conjunctions of such CQs. Using the laws of
distributivity, we can convert $q^d$ into an equivalent conjunction of
disjunctions of CQs 
$q_1 \wedge \cdots \wedge q_k$. To decide whether
$(\Dmc^d)^\approx_{\bar a,\ell,k},\Omc \models q^d$, it suffices to decide
whether $(\Dmc^d)^\approx_{\bar a,\ell,k},\Omc \models q_i$ for
$1 \leq i \leq k$. In the following, we concentrate on this task.
\smallskip

For the sake of readability, let us summarize the remaining task and
reorganize notation. We still have fixed $\ell,k \geq 1$ with
$\ell < k$. We further have a Boolean OMQ
$Q=(\Omc,\Sigma,q) \in (\ALCI, \text{UCQ})$ with each CQ in $q$
connected and a $\Sigma$-database \Dmc. 
We want to decide whether $\Dmc^\approx_{\bar a,\ell,k} \models Q$.

We reuse and suitable adapt the notions and algorithm from the proof
of Theorem~\ref{thm:newkidontheblock}.
With $\mn{cl}(q)$, we denote the set of all CQs that can be obtained
from a CQ in $q$  as follows:
\begin{enumerate}

\item take a contraction of treewidth $(\ell,k)$;
  
\item take a subquery, that is, the atoms in the new CQ are
  a (non-empty) subset of those in the given one;

\item choose zero or more quantified variables and make them answer
  variables.

\end{enumerate}
We shall consider CQs that we obtain from a CQ $p(\bar y)$ in
$\mn{cl}(q)$ by replacing the answer variables $\bar y$ with a tuple
of constants $\bar c$. We denote the result of such a replacement with
$p(\bar c)$ and call it an \emph{instantiation} of $p$.

We use $\Cl(\Omc,q)$ to denote the smallest set that contains all
concepts in \Omc and all ELIQs in $\Cl(q)$, viewed as
\ELI-concepts, and that
is closed under subconcepts and single negation. 


For a set $L \subseteq \mn{adom}(\Dmc)$ with $S \subseteq L$,
an \emph{$L$-assignment} $\mu$ is a set that contains 
\begin{itemize}
\item exactly one of $C(a), \neg C(a)$ for all $\neg C \in \cl(\Omc,q)$ and $a \in L$, and

\item exactly one of $p(\bar c), \neg p(\bar c)$ for all 
instantiations $p(\bar c)$ of a CQ from $\mn{cl}(q)$ with $\bar c
\subseteq L$.
\end{itemize}
We require that $\mu$ does not contain any contractions of
$q$ (thus it contains the negation of all such contractions of
treewidth $(\ell,k)$).
%
%
%
%
%
%
We shall be interested in $L$-assignments with $L \subseteq
\mn{adom}(\Dmc)$, $S \subseteq L$, and $|L\setminus S| \leq
\ell$. When speaking of sets $L \subseteq \mn{adom}(\Dmc)$, we from now
on silently assume that $L$ satisfies the latter two  conditions.

The $L$-assignment
$\mn{as}_\Imc(L)$ \emph{realized} in a model \Imc of \Omc with $L
\subseteq \Delta^\Imc$ is the set that contains the following:
\begin{itemize}

\item $C(a)$ for all $C \in \cl(\Omc,q)$ and $a \in L$ with $a \in C^\Imc$;

\item $p(\bar c)$ for all CQs $p(\bar y) \in \mn{cl}(q)$ with
  $\bar c \in p(\Imc)$ and $\bar c \in L$.
  
\end{itemize}
A \emph{$k$-subset} of $\mn{adom}(\Dmc)$ is a set
$K \subseteq \mn{adom}(\Dmc)$ such that $S \subseteq K$ and
$|K \setminus S| \leq k$. 

Let $\Gamma$ be a set of
$L$-assignments and let $K$ be a $k$-subset.  A \emph{$\Gamma$-choice} for $K$ is a subset
$\Gamma' \subseteq \Gamma$ that contains exactly one $L$-assignment
for each $L \subseteq K$.  Every $\Gamma$-choice $\Gamma'$ gives rise
to a \emph{canonical database} $\Dmc_{\Gamma'}$ that is constructed by
starting with $\Dmc|_K$ and then doing the following for all
$\mu \in \Gamma'$ and $p(\bar c) \in \mu$: rename all constants in
$\Dmc_p{(\bar c)}$ except those in $\bar c$ to fresh constants and add
the resulting database.  We say that $\Gamma'$ is \emph{consistent} if
the following conditions are satisfied:
\begin{enumerate}

\item the extended database $$\Dmc|_K \cup \{ C(a) \mid \exists \mu \in \Gamma': C(a) \in \mu
  \}$$ is satisfiable with the $\ALCI^u$-ontology
  $$
    \Omc \cup \{ \top \sqsubseteq \neg p \mid p \text{ contraction of
      $q$
      that is a BELIQ} \}.
  $$

\item If $\Dmc_{\Gamma'} \models p(\bar c)$ with $p(\bar y) \in \mn{cl}(q)$
  and $\bar c \in K$, and $\mu \in \Gamma'$ is an $L$-assignment with
  $\bar c \subseteq L$, then $p(\bar c) \in \mu$.

\end{enumerate}
The algorithm is now exactly identical to the one used in the proof of
Theorem~\ref{thm:newkidontheblock}. The running time can also be
analyzed in a similar way, the only difference being that the number
of $L$-assignments, for an $L \subseteq \mn{adom}(\Dmc)$ with
$|L| \leq \ell$, is now bounded single exponentially in $||\Omc||$ and
\emph{double} exponentially in $||q||$. The running time of the
algorithm changes accordingly and yields fixed-parameter tractability
with double exponentially running time. It remains to prove
correctness.
\begin{lemma}
  The algorithm answers 'yes' iff 
  $\Dmc^\approx_{S,\ell,k} \models Q$.
\end{lemma}
\noindent
\begin{proof}\
  For the `only if' direction, assume $\Dmc_{S, \ell, k}^\approx \not\models Q$.
  We need to show that the algorithm answers `no'.
  Since $\Dmc_{S, \ell, k}^\approx \not \models Q$, there is a model $\Imc$
  of $\Omc$ and $\Dmc_{S, \ell, k}^\approx$ with $\Imc \not \models p$
  for every disjunct $p$ in $q$.
  For every $a \in \mn{adom}(\Dmc_{S, \ell, k}^\approx)$,
  define \[\mn{tp}(a) = \{C \in \mn{cl}(\Omc, q) \mid a \in C^\Imc\}\]
  and for every $\bar c \subseteq \mn{adom}(\Dmc_{S, \ell, k}^\approx)$,
  define \[\mn{inst}(\bar c) = \{p(\bar y) \in \mn{cl}(q) \mid \bar c \in p(\Imc)\}.\]
  For every $\ell,k$-sequence $v=S_0, O_0, \ldots, O_{n-1}, S_n$ with $n \geq 1$
  we define $L_v \subseteq \mn{adom}(\Dmc)$ as $h(O_{n-1})$, where
  $h : \mn{adom}(\Dmc_{S, \ell, k}^\approx) \rightarrow \mn{adom}(\Dmc)$
  is the uncopying map, and we define the $L_v$-assignment
  \begin{align*}
  \mu_v \,=\, &\{C(h(a)) \mid C \in a \in L_v, \mn{tp}(a)\} \,\cup\\
  &\,\{p(h(\bar c)) \mid \bar c \subseteq L_v, p(\bar y) \in \mn{inst}(\bar c)\}.
  \end{align*}
  Since $\Imc$ is a model
  of $\Dmc_{S, \ell, k}^\approx$ with $\Imc \not \models q$,
  $\mu_v$ indeed satisfies the conditions of $L_v$-assignments.
  Let $\Psi = \{\mu_v \mid v \text{ is an } \ell,k\text{-sequence of length } n \geq 1\}$.
  We argue by induction on $i$ that $\Psi \subseteq \Gamma_i$
  for every $i \geq 0$. 
  Since $\Gamma_0$ is the set of all $L$-assignments, $\Psi \subseteq \Gamma_0$.
  For the induction step, assume $\Psi \subseteq \Gamma_i$ for some $i \geq 0$.
  Let $\mu_v \in \Psi$ be an $L_v$-assignment for some
  $v=S_0, O_0, \ldots, O_{n-1}, S_n$.
  We show that for every $k$-subset $K$ with $K \supseteq L_v$
  there is a consistent $\Gamma_i$-choice $\Gamma$ for $K$ such that $\mu_v \in \Gamma$.
  Let $K$ be a $k$-subset.
  For every $L \subseteq K$, choose a set $S_{n+1} \supseteq L$ with
  $|S_{n+1} \setminus S| \leq k$ and consider the sequence
  $v_{K,L}=S_0, O_0, \ldots, O_{n-1}, K, L, S_{n+1}$. This is
  an $\ell,k$-sequence that yields an $L$-assignment $\mu_{v_{K,L}} \in \Psi$.
  The set of all such $L$-assignments forms a $\Gamma_i$-choice $\Gamma$ for $K$.
  Since $\Imc$ is a model that realizes all $L$-assignments from $\Gamma$
  and such that $\Imc \not \models q$,
  $\Gamma$ satisfies the two conditions of a consistent $\Gamma_i$-choice,
  so $\mu_v$ remains in $\Gamma_{i+1}$.
  Since $\Psi$ is not empty, this implies that the algorithm answers `no'.
  
  For the `if' direction, assume the algorithm answers `no'. We need to show that
  $\Dmc_{S, \ell, k}^\approx \not\models Q$, so we construct a model
  $\Imc$ of $\Omc$ and $\Dmc_{S, \ell, k}^\approx$ with $\Imc \not \models q$.
  Since the algorithm answers `no', the set $\Gamma := \Gamma_f$ is not empty.
  This implies that $\Gamma$ contains at least one $L$-assignment for every set
  $L \subseteq \mn{adom}(\Dmc)$ with $S \subseteq L$ and $|L \setminus S| \leq \ell$,
  since as soon as there is an $L_0$ such that some $\Gamma_i$
  does not contain an $L_0$-assignment, the algorithm eventually
  eliminates all $L$-assignments and thus returns `yes'.
  Thus, for every $L$-assignment $\mu \in \Gamma$ and every $k$-subset $K$,
  there is a consistent $\Gamma$-choice $\Gamma_{\mu,K}$ for $K$
  such that $\mu \in \Gamma_{\mu,K}$.
  We construct a model $\Imc$ of $\Dmc_{S, \ell, k}^\approx$ and $\Omc$
  with $\Imc \not \models q$.
  
  For every $\ell,k$-sequence $v = S_0, O_0, S_1, \ldots, S_n$, we
  choose a consistent $\Gamma$-choice $\Gamma_v$, using the following inductive procedure.
  First, choose an arbitrary $S$-assignment $\mu_0 \in \Gamma$.
  For every sequence of the form $v=S_0$, choose a consistent $\Gamma$-choice
  $\Gamma_v \subseteq \Gamma$ with $\mu_0 \in \Gamma_v$. Since the algorithm
  did not remove $\mu_0$ from $\Gamma$, there is such a $\Gamma$-choice $\Gamma_v$.  
  For the induction step, assume that $\Gamma_v$ has already been defined
  for all $\ell,k$-sequences of length $n$ and consider $\ell,k$-sequence
  $v = S_0, O_0, \ldots, S_n, O_n, S_{n+1}$ of length $n+1$.
  Let $v' = S_0, O_0, \ldots, S_n$. Since $O_n \subseteq S_n$,
  $\Gamma_{v'}$ contains a (unique) $O_n$-assignment $\mu$.
  Choose a consistent $\Gamma$-choice $\Gamma_v \subseteq \Gamma$ for $S_{n+1}$.
  Such a $\Gamma_v$ must exists, since the algorithm did not remove
  $\mu$ from $\Gamma$.
  
  For every $\ell,k$-sequence $v=S_0,O_0,\ldots,S_n$, let $\Dmc_v$ be obtained from
  the canonical database $\Dmc_{\Gamma_v}$ by renaming every element
  $a \in \mn{adom}(\Dmc_{\Gamma_v}) \cap S_n$ to $a_v$, which is the
  name of the copy of $a$ used in the bag for $v$ in $\Dmc_{S, \ell, k}^\approx$.
  For every element $a_v \in S_n$,
  let $\mn{tp}(a) = \{C \in \mn{cl}(\Omc,q) \mid C(a) \in \mu_v\}$.
  To construct $\Imc$, we start with the union of $\Dmc_v$ over all
  $\ell,k$-sequences $v$.
  Additionally, for every $a_v \in \mn{adom}(\Dmc_{S, \ell, k})$,
  $v$ an $\ell,k$-sequence,
  choose a tree-shaped model $\Imc_a$ of $\mn{tp}(a)$ and add a copy of $\Imc_a$
  to $\Imc$ and identify $a$ with the root of $\Imc_a$. It can be verified that $\Imc$
  is indeed a model of $\Omc$. We refer to elements of
  $\Delta^\Imc \setminus \mn{adom}(\Dmc_{S, \ell, k})$ as \emph{anonymous elements}.
  
  To see that $\Imc \not \models q$, assume the opposite, so assume that there is
  a disjunct $p$ of $q$ and a homomorphism $h : p \rightarrow \Imc$.
  The range of $h$ cannot only consist of anonymous elements, since $p$ is
  connected and this would imply that the range of $h$ is completely contained
  in one of the $\Imc_a$, which in turn would imply $\mn{tp}(a) \models p$,
  but contradicts the first condition of consistent $\Gamma$-choices.
  Thus, the range of $h$ contains at least one element of $\mn{adom}(\Dmc_{S, \ell, k})$.
  
  Since $\Imc$ has treewidth $(l,k)$, there is a contraction $p'$ of
  $p$ of treewidth $(l,k)$ such that there is an injective
  homomorphism $h' : p' \rightarrow \Imc$.
  For every $\ell,k$-sequence $v=S_0,O_0,\ldots,S_n$,
  let $B_v \subseteq \mn{var}(p')$ be
  the set of all variables $x$ such that $u(h'(x)) \in S_n$, where $u$
  is the uncopying map, or such that $h'(x)$ is an anonymous element in
  $\Imc_a$, where $u(a) \in S_n$. It is possible to choose a finite
  set $V$ of $\ell,k$-sequences such that $\bigcup_{v \in V} B_v = \mn{var}(p')$
  and such that $V$ is connected, i.e.
  \begin{itemize}
  \item for each two $v,v' \in V$, $V$ also contains
    the longest common prefix of $v$ and $v'$ and
  \item for each two $v,v' \in V$ with $v'$ a prefix of $v$,
    $V$ contains every $\ell,k$-sequence $v''$ such that $v'$ is
    a prefix of $v''$ and $v''$ is a prefix of $v$.
  \end{itemize}
  This yields an $(\ell,k)$-tree decomposition $(V,E,B_v)$ of $p'$, where
  $(v,v') \in E$ if and only if $v'$ is an immediate successor of $v$.  
  For every $v \in V$, let $p'_v$ be $p'$ restricted to variables that appear
  in $B_v$ or in some $B_{v'}$ for some $v' \in V$ that is part of the subtree
  rooted at $v$. In particular, if $v$ is the root of $(V,E,B_v)$,
  then $p'_v = p'$.
  For every $p'_v$ where $v=S_0,O_0,\ldots,S_{n-1},O_{n-1},S_n$
  is not the root of $(V,E,B_v)$, make all variables $x$ where $u(h'(x)) \in O_{n-1}$
  answer variables and denote the resulting query by $p'_v(\bar y_v)$.
  Let $\bar c_v = u(h'(\bar y_v))$. If $v$ is the root of $(V,E,B_v)$, we do not
  add any answer variables to $p'_v$.
  
  \smallskip
  \noindent
  \textbf{Claim:} For every $v \in V$,  $p'_v(\bar c_v) \in \Gamma_v$.
  
  \smallskip
  \noindent
  The claim can be proven inductively on the position of $v$ in $(V, E, B_v)$, starting
  with the leaves. If $v$ is a leaf, then $p'_v(\bar c_v) \in \Gamma_v$ follows
  from the existence of the homomorphism $h'$ restricted to $p'_v(\bar y_v)$.
  If $v = S_0,O_0,\ldots,O_{n-1},S_n$ is not a leaf, then
  let $v_1,\ldots,v_n$ be the children of $v$ in $(V, E, B_v)$.
  By induction, the claim holds for $v_1,\ldots,v_n$, so we have
  $p'_{v_i}(\bar c_{v_i}) \in \Gamma_{v_i}$ for all $i$ with $1\leq i \leq n$.
  Since every $c_{v_i} \subseteq S_n$, $p'_v(\bar c_v) \in \Gamma_v$ for
  for all $i$ with $1\leq i \leq n$. Together with the homomorphism $h'$ restricted
  to $B_v$, this yields $p'_v(\bar c_v) \in \Gamma_v$, which finishes the proof
  of the claim.
  
  For $v$ the root of $V$, the claim yields $p'() \in \Gamma_v$, which contradicts
  the requirement that every $L$-assignment $\mu \in \Gamma$ contains the negation
  of $p'()$ for every contraction $p'$ of $p$.
  Thus, $\Imc$ is a model of $\Omc$ and
  $\Dmc_{S, \ell, k}^\approx$ with $\Imc \not \models q$.

\end{proof}

\section{Proofs for Section~\ref{sect:fromabove}}

\subsection{Ontology strengthening approximation}

Our aim is to prove Theorem~\ref{thm:ontabove}. We split it here
into two theorems, one for the upper bound and one for the lower
bound. Let $\ELIU^-_\bot$ be the restriction of $\ELIU_\bot$ where
no disjunction is admitted on the left-hand side of concept
inclusions. Note that every $\ELIU_\bot$-ontology can easily
be transformed into an equivalent $\ELIU^-_\bot$-ontology, but
at the cost of an exponential blowup.
\begin{theorem}
  \label{thm:ontaboveupper}
  $\ELI_\bot$-ontology strengthening OMQ evaluation in
  $(\ELIU_\bot,\text{UCQ})$ is in $\TwoExpTime$ in combined
  complexity and FPT with double exponential running time.
  In $(\ELIU^-_\bot,\text{UCQ})$, it is in $\ExpTime$ in combined
  complexity and FPT with single exponential running time.
\end{theorem}
To prove Theorem~\ref{thm:ontaboveupper}, we start with some
preliminaries.
Let $\Imc_1,\ldots,\Imc_n$ be interpretations. The \emph{direct product}
of $\Imc_1,\ldots,\Imc_n$ is the interpretation $\Imc$ with
$$
\begin{array}{rcl}
\Delta^\Imc &=& \Delta^{\Imc_1} \times \ldots \times \Delta^{\Imc_n} \\[1mm]
A^\Imc &=& \{(a_1,\ldots,a_n) \mid a_i \in A^{\Imc_i}
\text{ for } 1 \leq  i \leq n\} \\[1mm]
r^\Imc &=& \{ ((a_1,\ldots,a_n),(b_1,\ldots,b_2))
           \mid (a_i,b_i) \in r^{\Imc_i} \\[1mm]
  &&\hspace*{4.5cm}
\text{for } 1 \leq i \leq n\}
\end{array}
$$
for all role names $r$ and concept names $A$.
We denote the direct product of $\Imc_1,\ldots,\Imc_n$ by
$\prod_{i \in \{1,\ldots,n\}} \Imc_i$, and we denote the direct product of
two interpretations $\Imc_1$ and $\Imc_2$ by $\Imc_1 \otimes \Imc_2$.
It is well known that being a model of an $\ELI_\bot$ ontology is
preserved under direct
products \cite{DBLP:journals/tocl/HernichLPW20}, i.e.\ if $\Imc_1,\ldots,\Imc_n$ are models of an $\ELI_\bot$-ontology~$\Omc$,
then $\prod_{i \in \{1,\ldots,n\}} \Imc_i$ is also a model of $\Omc$.

The proof of Theorem~\ref{thm:ontaboveupper} is based on exhaustive
$\ELI_\bot$-approximation sets as introduced in the main body of the
paper and crucially uses direct products. The following example gives
a first idea.
\begin{exmp}
\label{ex:above}
Let $\Omc=\{\top \sqsubseteq A_1 \sqcup A_2\} \cup \Omc'$, with
$\Omc'$ an \ELI-ontology. Further let
$\Omc_1 = \{\top \sqsubseteq A_1\} \cup \Omc'$ and
$\Omc_2 = \{\top \sqsubseteq A_2\} \cup \Omc'$. We show below that
$\Mmc = \{\Omc_1,\Omc_2\}$ is an exhaustive $\ELI_\bot$-approximation
set. Consequently for all OMQs $Q(\bar x)=(\Omc,\Sigma,q)$ with $q$ a
UCQ,
$\mn{app}_{\ELI_\bot}^\uparrow (Q, \Dmc)=Q_1(\Dmc) \cap Q_2(\Dmc)$
where $Q_i=(\Omc_i,\Sigma,q)$.

Assume to the contrary that \Mmc is not an exhaustive
$\ELI_\bot$-approximation set. Then there is an $\ELI_\bot$-ontology $\Omc'$ such that $\Omc' \models \Omc$,
$\Omc' \not \models \Omc_1$, and $\Omc' \not \models \Omc_2$. This
means that there are models
$\Imc_1$ and $\Imc_2$ of $\Omc'$, such that $\Imc_i$ is not a model of
$\Omc_i$
for $i \in \{1,2\}$. In particular, there is an element $a \in \Delta^{\Imc_i}$ such that
$a \notin A^{\Imc_i}$.
Then $\Imc_1 \otimes \Imc_2$
is a model of $\Omc'$, but $(a,b) \in (\neg A_1 \sqcap \neg A_2)^{\Imc_1 \otimes \Imc_2}$.
Thus, $\Imc_1 \otimes \Imc_1$ is not a model of $\Omc$, contradicting $\Omc' \models \Omc$.
\end{exmp}

\medskip

For a given $\ELIU_\bot$-concept $C$, define the set $f(C)$ inductively as follows:
\begin{itemize}
\item If $C \in \{\top, \bot\} \cup \NC$, then $f(C) = \{C\}$.
\item If $C = \exists r. D$, then $f(C) = \{\exists r. D' \mid D' \in f(D)\}$.
\item If $C = D_1 \sqcap D_2$, then $f(C) = \{D_1' \sqcap D_2' \mid D_1' \in f(D_1), D_2' \in f(D_2)\}$.
\item If $C = D_1 \sqcup D_2$, then $f(C) = f(D_1) \cup f(D_2)$.
\end{itemize}
It is easy to check that every $\ELIU_\bot$-concept $C$ is equivalent
to $\bigsqcup_{C' \in f(C)} C'$.

Given an $\ELIU_\bot$-ontology $\Omc$, we first construct an
equivalent ontology $\Omc'$ obtained from $\Omc$ by replacing every
$C \sqsubseteq D \in \Omc$ by the set of CIs
$\{C' \sqsubseteq D \mid C' \in f(C)\}$. If $\Omc$ is formulated in
$\ELIU^-_\bot$, then $\Omc' = \Omc$. In general though, the size of
$\Omc'$ is exponential in that of $\Omc$. Now we define $\Emc_\Omc$ to
be the set of ontologies $\Omc''$ that can be obtained by choosing for
every CI $C \sqsubseteq D \in \Omc'$, a CI $C \sqsubseteq D'$ with 
$D' \in f(D)$ and including it in $\Omc''$.

\begin{lemma}
\label{lem:eliubot-exhaustive}
For every $\ELIU_\bot$-ontology $\Omc$, the set $\Emc_\Omc$ is an exhaustive $\ELI_\bot$-approximation set for $\Omc$.
\end{lemma}

\noindent
\begin{proof} \
Assume $\Emc_\Omc$ is not an exhaustive $\ELI_\bot$-approximation set for $\Omc$. Then
there exists an $\ELI_\bot$-ontology $\Omc'$ such that $\Omc' \models \Omc$ and
$\Omc' \not \models \widehat \Omc$ for every $\widehat \Omc \in \Emc_\Omc$.
Thus, for every $\widehat \Omc \in \Emc_\Omc$ there is a model $\Imc_{\widehat \Omc}$
of $\Omc'$ that is not a model of $\widehat \Omc$.
Thus, for every $\widehat \Omc$ there is a CI
$C_{\widehat \Omc} \sqsubseteq D_{\widehat \Omc} \in \widehat \Omc$
that is violated by $\Imc_{\widehat \Omc}$.
Let $S$ be the set of all such CIs that are violated by some $\Imc_{\widehat \Omc}$.
We claim that there is a $C \sqsubseteq D \in \Omc$ such that
$S$ contains $C \sqsubseteq D'$ for all $D' \in f(D)$. If this was not the case, then
for every $C \sqsubseteq D \in \Omc$, there is a $D' \in f(D)$
with $C \sqsubseteq D' \notin S$.
But then the ontology $\Omc_0 \in \Emc_\Omc$ obtained as the product
of these choices contains only CIs that are never violated in any of the $\Imc_\Omc$,
contradicting that $\Imc_{\Omc_0}$ violates some CI from $\Omc_0$.
From now on, let $C \sqsubseteq D \in \Omc$ a fixed CI that fulfils 
$\{C \sqsubseteq D' \mid D' \in f(D)\} \subseteq S$.
Let $\Mmc = \{\widehat \Omc \in \Emc_\Omc \mid \Imc_{\widehat \Omc}$
violates a CI of the form $C \sqsubseteq D'$ for some $D' \in f(D)\}$.

Define $\Imc = \prod_{\widehat \Omc \in \Mmc} \Imc_{\widehat \Omc}$.
Since models of $\ELI_\bot$-ontologies are closed under products, $\Imc$ is a model
of $\Omc'$. We aim to show that $\Imc$ is not a model of $\Omc$, contradicting
$\Omc' \models \Omc$. For every $\widehat \Omc \in \Mmc$,
$\Imc_{\widehat \Omc}$ violates a CI of the form $C \sqsubseteq D'$ for some $D' \in f(D)$.
Let $a_{\widehat \Omc} \in \Delta^{\Imc_{\widehat \Omc}}$ such that
$a_{\widehat \Omc} \in \Imc^{C_{\widehat \Omc}} \setminus \Imc^{D_{\widehat \Omc}}$.
Since models of $\ELI_\bot$-ontologies are closed under products, the element
$a_0 := \prod_{\widehat \Omc \in \Mmc} a_\Omc$ fulfils $a_0 \in \Imc^C$, but
$a_0 \notin \Imc^{D'}$ for all $D' \in f(D)$. Since $D \equiv \bigsqcup_{D' \in f(D)} D'$,
$a_0 \notin \Imc^D$, showing that $\Imc$ is not a model of $\Omc$.
\end{proof}

Lemma~\ref{lem:eliubot-exhaustive} allows us to compute
$\mn{app}_{\ELI_\bot}^\uparrow(Q,\Dmc)$ for any given
$Q \in (\ELIU_\bot, \text{UCQ})$ and database $\Dmc$ by computing $\Emc_\Omc$
and evaluating the query under every ontology $\widehat \Omc \in \Emc_\Omc$.
More precisely, we use the following algorithm:
Given $Q = (\Omc, \Sigma, q(\bar x)) \in (\ELIU_\bot, \text{UCQ})$,
a $\Sigma$-database \Dmc, and $\bar a \in \mn{dom}(\Dmc)^{|\bar x|}$,
iterate over all ontologies $\Omc' \in \Emc_\Omc$ and check whether
$\Omc', \Dmc \models Q(\bar a)$. If this is the case for every $\Omc' \in \Emc_\Omc$,
answer \emph{yes}, otherwise \emph{no}. Correctness follows from 
Lemma~\ref{lem:eliubot-exhaustive} and the definition of
$\ELI_\bot$-approximation sets.
For the running time, note that the number of ontologies in $\Emc_\Omc$
is at most double exponential in $||\Omc||$ and
every $\Omc' \in \Emc_\Omc$ is of size at most single exponential in $||\Omc||$.
Since answering OMQs from $(\ELI_\bot,\text{UCQ})$ is in $\ExpTime$
\cite{DBLP:conf/jelia/EiterGOS08} and FPT with single exponential running time,
this is yields a $\TwoExpTime$ algorithm and FPT with double
exponential running time. If the original ontology \Omc is formulated
in $\ELIU^-_\bot$, then the ontologies in $\Emc_\Omc$ are only of
polynomial size and thus we obtain an \ExpTime algorithm.

\begin{theorem}
\label{thm:above-2explower}
$\ELI_\bot$-ontology strengthening OMQ evaluation in $(\ELIU_\bot, \text{AQ})$
is $\TwoExpTime$-hard in combined complexity.
\end{theorem}

We prove Theorem~\ref{thm:above-2explower} by a reduction from the word problem
of alternating exponentially space bounded Turing machines, which is
$\TwoExpTime$-complete~\cite{DBLP:journals/jacm/ChandraKS81}. 
An \emph{alternating Turing Machine (ATM)} is a tuple
$(Q, \Sigma, \Gamma, q_0, q_a, q_r, \Delta)$ where
\begin{itemize}
\item $Q = Q_\exists \uplus Q_\forall \uplus \{q_a, q_r\}$ is the set of states,
\item $\Sigma$ is the input alphabet,
\item $\Gamma \supseteq \Sigma$ is the tape alphabet that contains a \emph{blank symbol} $\square \in \Gamma \setminus \Sigma$,
\item $q_0 \in Q$ the starting state,
\item $q_a \in Q$ the accepting state,
\item $q_r \in Q$ the rejecting state and
\item $\Delta \subseteq (Q_\exists \cup Q_\forall) \times \Gamma \times \Gamma \times \{l,r\} \times Q$ the transition relation.
\end{itemize}

A \emph{configuration} is a word $wqw'$ with $w,w' \in \Gamma^*$ and $q \in Q$.
The intended meaning is that the tape contains the word $ww'$
(with only blanks before and behind it),
the machine is in state $q$ and the head is on the leftmost
symbol of $w'$. A configuration $vpv'$ is a \emph{successor configuration} of a
configuration $wq\gamma_1w'$ if there is a tuple $(q,\gamma_1, \gamma_2, d, p) \in \Delta$
and $vpv'$ is obtained from $wq\gamma_1w'$ by replacing the $\gamma_1$ under the head by $\gamma_2$,
change the state from $q$ to $p$ and move the head one step into direction $d$.
A \emph{halting configuration} is of the form $vqv'$ with
$q \in \{q_a, q_r\}$.
We inductively define accepting configurations.
Let $c$ be a configuration.
\begin{itemize}
\item If $c$ is of the form $wq_aw'$, then $c$ is \emph{accepting}.
\item If $c$ is of the form $wqw'$ with $q \in Q_\exists$ and there is a successor
configuration $c'$ of $c$ such that $c'$ is accepting, then $c$ is \emph{accepting}.
\item If $c$ is of the form $wqw'$ with $q \in Q_\forall$ and every successor configuration $c'$
of $c$ is accepting, then $c$ is \emph{accepting}.
\end{itemize}
The \emph{starting configuration} for a word $w \in \Sigma^*$
is the configuration $q_0w$.
A word $w$ is accepted by $M$ if the starting configuration for $w$ is accepting.
We use $L(M)$ to denote the set $\{w \in \Sigma^* \mid M \text{ accepts } w\}$.

Let $M$ be an exponentially space bounded ATM that decides a $\TwoExpTime$-complete
problem.
Given an input word $w \in \Sigma^*$,
we construct in polynomial time an OMQ $Q \in (\ELIU_\bot, \text{AQ})$ and
a database $\Dmc = \{A(a)\}$ such that
$w \in L(M)$ if and only if $a \in \mn{app}_{\ELI_\bot}^\uparrow(Q,\Dmc)$.
W.l.o.g.\ we make the following assumptions about $M$:
\begin{itemize}
\item $q_0 \in Q_\exists$.
\item For every $q \in Q_\exists \cup Q_\forall$ and $\gamma_1 \in \Gamma$, there are precisely
two transitions of the form $(q,\gamma_1, \gamma_2, d, p) \in \Delta$. We consider
$\Delta$ as two functions $\delta_1, \delta_2 : (Q_\exists \cup Q_\forall) \times \Gamma \rightarrow \Gamma \times \{l,r\} \times Q$ where each $\delta_i$ yields one of the two transitions.
\item For every configuration with a state from $Q_\exists$, both successor
configurations have a state from $Q_\forall$, and vice versa.
\item On an input $w$, $M$ uses at most $2^{|w|}$ cells of the tape, and it does
not use cells to the left of the starting position of the head.
\end{itemize}

The general idea for the reduction is to construct $\Omc$ in such a way
that of all the ontologies in $\Emc_\Omc$, only a single ontology $\Omc_0$
is satisfiable with $\Dmc$. By Lemma~\ref{lem:eliubot-exhaustive},
this implies that $a \in \mn{app}_{\ELI_\bot}^\uparrow(Q,\Dmc)$ if and only
if $\Omc_0, \Dmc \models Q(a)$. The ontology $\Omc_0$ has the effect that
starting from the single assertion $A(a)$ in the database, it creates
the tree of all relevant configurations of $M$ on input $w$ in the anonymous part,
and if $M$ accepts $w$, it propagates the query concept name $B$ back to
the element $a$.

We describe the reduction in two steps. First, we describe $\Omc_0$ and argue
that $\Omc_0, \Dmc \models Q(a)$ if and only if $w \in L(M)$.
Secondly, we construct $\Omc$ such that $\Omc_0 \in \Emc_\Omc$ and
such that all other ontologies in $\Emc_\Omc$ are unsatisfiable with $\Dmc$.

Let $w=w_1w_2\ldots w_n \in \Sigma^*$. Starting from $A$, we generate
an infinite binary tree where each node stands for a configuration.
The concept name $T$ represents a configuration, the concept name $T_\exists$
represents configurations where the state is from $Q_\exists$, and
the concept name $T_\forall$
represents configurations where the state is from $Q_\forall$.
The role name $s$ serves as the successor relation, and we use 
concept names $S_1$ and $S_2$ to distinguish the two successors from each other.
\begin{align*}
A &\sqsubseteq T \sqcap T_\exists\\
T &\sqsubseteq \exists s. (T \sqcap S_1)\\
T &\sqsubseteq \exists s. (T \sqcap S_2)\\
\exists s^-.T_\exists &\sqsubseteq T_\forall\\
\exists s^-.T_\forall &\sqsubseteq T_\exists
\end{align*}
To store a configuration of $M$ in every node, we let $T$ generate a binary
tree of depth $n$, so that the $2^n$ leaves of this tree can be used as tape cells.
We use concept names of the form $B_i^j$,
$0\leq i < n$, $j \in \{0,1\}$, to encode which leaf encodes which position
of the tape, where $B_i^j$ means that the $i$th bit of the position (in binary) is $j$.
For convenience, we label the two children of every inner node of the tree with $L$ and $R$
to indicate whether the subtree contains the left or the right half of the tape.
\begin{align*}
T &\sqsubseteq L_0\\
L_i &\sqsubseteq \exists r.(L_{i+1} \sqcap B_i^0 \sqcap L) \qquad \text{for } 0 \leq i < n\\
L_i &\sqsubseteq \exists r.(L_{i+1} \sqcap B_i^1 \sqcap R) \qquad \text{for } 0 \leq i < n\\
\exists r^-.B_i^j &\sqsubseteq B_i^j \qquad\text{for } 0 \leq i < n, j \in \{0,1\}
\end{align*}
Next, we populate the tree below $A$ with the starting configuration.
We first label every node in the tree of the starting configuration with a
concept name $T_0$.
\begin{align*}
A &\sqsubseteq T_0\\
\exists r^-.T_0 &\sqsubseteq T_0
\end{align*}
The content of a tape cell is encoded in a concept name of the form $M_{\gamma, q}$ or $M_\gamma$, 
where $\gamma \in \Gamma$ and $q \in Q$.
A concept name of the form $M_{\gamma, q}$ means that the tape cell contains the symbol $\gamma$
the head is currently in this tape cell, and the ATM is in state $q$.
A concept name $M_\gamma$ means that the tape cell contains the symbol $\gamma$
and the head is currently not in this tape cell.
For every $x \in \{0, \ldots, 2^n-1\}$ with binary representation $x_0x_1\ldots x_{n-1}$,
we write $B_x$ for the concept $B_0^{x_0} \sqcap B_1^{x_1} \sqcap \ldots \sqcap B_{n-1}^{x_{n-1}}$.
We introduce the following CIs:
\begin{align*}
T_0 \sqcap L_n \sqcap B_0 \ &\sqsubseteq M_{w_0,q_0}\\
T_0 \sqcap L_n \sqcap B_i \ &\sqsubseteq M_{w_i} \qquad \text{ for } 1 \leq i \leq n
\end{align*}
The next CIs write blanks into all of the remaining positions.
\begin{align*}
T_0 \sqcap L_n \sqcap B_n \ &\sqsubseteq E_0\\
\exists r.E_0 &\sqsubseteq E_0\\
R \sqcap \exists r^-.\exists r.(L \sqcap E_0) &\sqsubseteq E_1\\
\exists r^-.E_1 &\sqsubseteq E_1\\
E_1 \sqcap L_n &\sqsubseteq M_{\square}
\end{align*}
This finishes the starting configuration. Next, we implement the transitions
of $M$. Let
\[\Mmc = \{M_{\gamma,q}, M_\gamma \mid \gamma \in \Gamma, q \in Q\}\,.\]
For every concept name $M \in \Mmc$, we introduce a concept name $M'$.
To implement the transitions of the ATM, we first
copy every configuration to its two successor configurations, but using the
primed concept names in the successor.
For every number $i \in \{0, \ldots, 2^{n-1}\}$ and every $M \in \Mmc$,
use the following CIs:
\begin{align}
\label{eq:CIs}
B_i \sqcap \exists (r^-)^n.\exists s^-.\exists r^n.(B_i \sqcap M) \sqsubseteq M' \qquad
\end{align}
A short remark regarding the second part of the proof:
Note that the number of CIs introduced in (\ref{eq:CIs}) is exponential in $n$.
In the second part of the proof, we will show how to produce these CIs in $\Omc_0$
using only a single CI in $\Omc$ that can be replaced in exponentially many ways.
The key problem of the second part of the proof will then be to construct $\Omc$ such
that all other ontologies in $\Emc_\Omc$ that are obtained by replacing the CI in
$\Omc$ in an unintended way become unsatisfiable with~$\Dmc$.

Now we calculate the two successor configurations that are determined by $\delta_1$
and $\delta_2$. We propagate the concept names $S_1$ and $S_2$ to the leaves of the
trees by introducing for $i \in \{1, 2\}$ the following CI:
\begin{align*}
\exists r^-.S_i \sqsubseteq S_i
\end{align*}

When computing the $i$th successor configuration for $i \in \{1, 2\}$,
the content of the cell that contained the head in the previous configuration
depends on $\delta_i$. So for every $\gamma \in \Gamma$ and $q \in Q$ and $i \in \{1, 2\}$,
let $\delta_i(q,\gamma) = (\alpha, d, p)$ for some
$\alpha \in \Gamma$, $d \in \{l,r\}$ and $p \in Q$.
We introduce the following CI:
\begin{align*}
M_{\gamma,q}' \sqcap S_i \sqsubseteq M_{\alpha} \sqcap F_{p,d}
\end{align*}
Here, $F_{p,d}$ is a fresh concept name that indicates that a transition needs to be done, namely
into state $p$ while moving the head one step in direction $d$.
We propagate $F_{p,d}$ to the leaf that represents the tape cell one step in direction $d$.
For the case $d=l$, we introduce the following CIs for every $\beta \in \Gamma$:
\begin{align*}
\exists r.(L \sqcap F_{p,l}) &\sqsubseteq F_{p,l}\\
L \sqcap \exists r^-.\exists r.(R \sqcap F_{p,l}) &\sqsubseteq F'_{p,l}\\
R \sqcap \exists r^-.F'_{p,l} &\sqsubseteq F'_{p,l}\\
F'_{p,l} \sqcap M_{\beta}' \sqcap L_n &\sqsubseteq M_{\beta, p}
\end{align*}
For the case $d=r$, we introduce the following CIs for every $\beta \in \Gamma$:
\begin{align*}
\exists r.(R \sqcap F_{p,r}) &\sqsubseteq F_{p,r}\\
R \sqcap \exists r^-.\exists r.(L \sqcap F_{p,r}) &\sqsubseteq F'_{p,r}\\
L \sqcap \exists r^-.F'_{p,r} &\sqsubseteq F'_{p,r}\\
F'_{p,r} \sqcap M_{\beta}' \sqcap L_n &\sqsubseteq M_{\beta, p}
\end{align*}
Finally, we send a marker $H$ to every other tape cell, to notify the tape cell that it
does not contain the head of the ATM in this configuration.
For every $\gamma \in \Gamma$ and every $q \in Q$, we introduce the CI
\begin{align*}
M_{\gamma,q} \sqsubseteq H
\end{align*}
and propagate it to all other tape cells using the following CIs:
\begin{align*}
\exists r.H &\sqsubseteq H\\
R \sqcap \exists r^-.\exists r. (L \sqcap H) &\sqsubseteq H'\\
L \sqcap \exists r^-.\exists r. (R \sqcap H) &\sqsubseteq H'\\
\exists r^-.H' &\sqsubseteq H'
\end{align*}
After the marker arrived, we can create the symbol that encodes the content of the
tape cell. For every $\beta \in \Gamma$, we introduce the following CIs:
\begin{align*}
H' \sqcap M_{\beta}' \sqcap L_n &\sqsubseteq M_{\beta}
\end{align*}
To check acceptance, we mark the root of every tree that encodes an accepting configuration
with the concept name $B$. For every $\gamma \in \Gamma$, introduce the following CIs:
\begin{align*}
M_{\gamma,q_a} &\sqsubseteq B\\
\exists r.B &\sqsubseteq B\\
T_\forall \sqcap \exists s.(S_1 \sqcap B) \sqcap \exists s.(S_2 \sqcap B) &\sqsubseteq B\\
T_\exists \sqcap \exists s. B &\sqsubseteq B
\end{align*}
This finishes the definition of $\Omc_0$.
It can be verified that $w \in L(M)$ if and only if $\Omc_0, \Dmc \models B(a)$.

For the second part of the proof, we would like to construct the ontology $\Omc$ such that
$\Omc$ is of polynomial size,
$\Omc_0 \in \Emc_\Omc$ and
such that $\Dmc$ is unsatisfiable with every $\Omc' \in \Emc_\Omc$, $\Omc' \neq \Omc_0$.
In fact, our construction of $\Omc$ will not achieve this, but instead produces
an ontology $\Omc_1 \in \Emc_\Omc$ which satisfies $\Omc_1 \supseteq \Omc_0$,
and such that $\Omc_0$ and $\Omc_1$ yield the same certains answers to OMQs on $\Dmc$,
which is sufficient to make the reduction work.
The ontology $\Omc$ consists of all CIs in $\Omc_0$, besides the CIs~(\ref{eq:CIs}).
Instead of the exponentially many CIs~(\ref{eq:CIs}),
we include in $\Omc$ the following single CI.
\begin{align*}
&(B_0^0 \sqcup B_0^1) \sqcap \ldots \sqcap (B_{n-1}^0 \sqcup B_{n-1}^1) \sqcap\\
&(\exists r^-.)^n \exists s^-. (\exists r.)^n(\\
&(B_0^0 \sqcup B_0^1) \sqcap \ldots \sqcap (B_{n-1}^0 \sqcup B_{n-1}^1) \sqcap \bigsqcup_{M \in \Mmc} M)\\
&\sqsubseteq (\bigsqcup_{M \in \Mmc} M') \sqcup D \hspace{5cm} (2)
\end{align*}
Here, $D$ is a fresh dummy concept name.
Recall the procedure how $\Emc_\Omc$ is obtained from an ontology $\Omc$
and note that applying the construction of $\Emc_\Omc$ to the CI~(2) indeed yields
an ontology that contains (among others) all the CIs~(\ref{eq:CIs}).
In fact, the construction of $\Emc_\Omc$ forces us to include a CI for every
concept from $f(\textnormal{LHS~of~(2)})$. This also includes concepts where
the two encoded tape positions in the left hand side are different, i.e.\ where
there exists an $i \in \{0, \ldots, n-1\}$ such that once $B_i^0$ was chosen and
once $B_i^1$ was chosen to appear in the CI. For all CIs of this form, we would
like to choose the dummy concept name $D$ on the right hand side.

To summarize, there are two kinds unintended CIs that we want to `disable', meaning that we want
every ontology from $\Emc_\Omc$ containing such a CI to become unsatisfiable with $\Dmc$.
The first kind are CIs of the form
\[B_i \sqcap (\exists r^-.)^n\exists s^-.(\exists r.)^n(B_i \sqcap M)
\sqsubseteq N'\]
with $M, N \in \Mmc$ and $M \neq N$. The second kind are of the form
\[B_i \sqcap (\exists r^-.)^n\exists s^-.(\exists r.)^n(B_j \sqcap M)
\sqsubseteq N'\]
with $M, N \in \Mmc$ and $i \neq j$
To achieve this, for every uninteded CI, we generate a substructure that contradicts this CI.

For every concept from $f(\textnormal{LHS~of~(2)})$,
we generate an element that satisfies that concept.
We construct a tree using a fresh role name $t$
such that for every $i \in \{0, \ldots, n-1\}$,
every leaf is labelled with precisely one concept name
from $X_i^0$ and $X_i^1$,
precisely one concept name from $Y_i^0$ and $Y_i^1$,
and precisely one concept name of the form $\widehat M$ for some $M \in \Mmc$.
We use concept names $V_i$, $i \in \{0, \ldots, 2n+1\}$ to label the layers of the tree.
For all $i \in \{0, \ldots, n-1\}$ and $j \in \{0, 1\}$, introduce the following CIs:
\begin{align*}
A &\sqsubseteq V_0\\
V_i &\sqsubseteq \exists t.(V_{i+1} \sqcap X_i^0) \sqcap
\exists t.(V_{i+1} \sqcap X_i^1)\\
V_{n+i} &\sqsubseteq \exists t.(V_{n+i+1} \sqcap Y_{n+i}^0) \sqcap
\exists t.(V_{n+i+1} \sqcap Y_{n+i}^1)\\
\exists t^-.X_i^j &\sqsubseteq X_i^j\\
\exists t^-.Y_i^j &\sqsubseteq Y_i^j
\end{align*}
The last layer of the tree is generated by the following CIs for every $M \in \Mmc$.
\[V_{2n} \sqsubseteq \exists t.(V_{2n+1} \sqcap \widehat M)\]

At every leaf, we figure out whether the two numbers encoded by the $X_i^j$ and the
$Y_i^j$ are equal. If so, we derive a marker $K_=$, otherwise a
marker $K_{\neq}$.
For every $i \in \{0, \ldots, n-1\}$, introduce the following CIs:
\begin{align*}
X_i^0 \sqcap Y_i^1 &\sqsubseteq K_{\neq}\\
X_i^1 \sqcap Y_i^0 &\sqsubseteq K_{\neq}\\
X_i^0 \sqcap Y_i^0 &\sqsubseteq Z_i\\
Z_0 \sqcap \ldots \sqcap Z_{n-1} &\sqsubseteq K_=
\end{align*}
Below every leaf of the tree, we genereate a substructure used to contradict one specific CI.
For every $i \in \{0, \ldots, n-1\}$ and $j \in \{0,1\}$, introduce the following CIs:
\begin{align*}
V_{2n+1} &\sqsubseteq \exists (r^-)^n. \exists s^-. \exists r^n.V\\
\exists (r^-)^n.\exists s.\exists r^n.X_i^j &\sqsubseteq X_i^j\\
\exists (r^-)^n.\exists s.\exists r^n.\widehat M &\sqsubseteq \widehat M\\
Y_i^j \sqcap V_{2n+1} &\sqsubseteq B_i^j\\
X_i^j \sqcap V &\sqsubseteq B_i^j\\
\widehat M \sqcap V &\sqsubseteq M
\end{align*}
Here, the element satisfying $V_ {2n+1}$ plays the role of a tape cell in a successor configuration,
while the element satisfying $V$ plays the role of a tape cell in the previous configuration.
To disable all CIs of the form
\[B_i \sqcap \exists (r^-)^n.\exists s^-.\exists r^n.(B_i \sqcap M)
\sqsubseteq N'\]
with $M \neq N$, we introduce for every $M \in \Mmc$ the CI
\begin{align*}
K_= \sqcap \widehat M \sqsubseteq M'
\end{align*}
and for each two $M_1, M_2 \in \Mmc$ with $M_1 \neq M_2$, we introduce the CIs
\begin{align*}
K_= \sqcap M_1' \sqcap M_2' &\sqsubseteq \bot\\
K_= \sqcap M_1' \sqcap D &\sqsubseteq \bot
\end{align*}
To disable all CIs of the form
\[B_i \sqcap \exists (r^-)^n.\exists s^-.\exists r^n.(B_j \sqcap M)
\sqsubseteq N'\]
with $i \neq j$, we have introduced the dummy concept name $D$ in the right hand side of CI~(2).
We introduce for every $M \in \Mmc$ the following CI:
\[ K_{\neq} \sqcap M' \sqsubseteq \bot \]
This finishes the definition of $\Omc$.
If can be verified that in $\Emc_\Omc$, the only
ontology that is satisfiable with $\Dmc$ is the unique ontology $\Omc_1$ that contains
every CI from $\Omc_0$. This yields the following result.
\begin{lemma}
For the OMQ $Q = (\Omc, \{A\}, B(x))$ and database $\Dmc = \{A(a)\}$ constructed from $M$ and $w$,
it holds that $w \in L(M)$ if and only if $a \in \mn{app}^\uparrow_{\ELI_\bot}(Q,\Dmc)$.
\end{lemma}

We discuss Example~\ref{ex:infiniteapproxsets} in which
$\Omc = \{\exists r.\top \sqcap \forall r . A \sqsubseteq B_1 \sqcup B_2 \}$. Then for
each $n \geq 1$, the $\ELI_\bot$-ontology
$$
\Omc_n = \{ \exists r.A \sqsubseteq \exists r^n . X, \ \exists r . (A
\sqcap \exists r^{n-1} .X) \sqsubseteq B_1\}
$$
is such that $\Omc_n \models \Omc$. We show
that every $\ELI_\bot$-ontology $\Omc'_n$
with $\Omc_n \models \Omc'_n \models \Omc$ is equivalent 
to $\Omc_{n}$. Assume $\Omc_{n}'$ with $\Omc_{n} \models \Omc_{n}'\models \Omc$ is given and assume for a proof by contradiction that $\Omc_{n}'\not\models \Omc_{n}$. Assume first that $\Omc_{n}'\not\models \exists r.A \sqsubseteq \exists r^n . X$. Then one can show that the interpretation  $\Imc$ with $\Delta^{\Imc}=\{0,\ldots,n\}$ and $r^{\Imc}= \{(i,i+1) \mid 0\leq i <n\}$, $A^{\Imc}= \{1\}$ is a model of $\Omc'_{n}$ and we have derived a contradiction to $\Omc'_{n}\models \Omc$. Now assume that  $\Omc_{n}'\not\models \exists r . (A
\sqcap \exists r^{n-1} .X) \sqsubseteq B_1$. Then one can show that the interpretation $\Imc'$ obtained from $\Imc$ by adding $n$ to $X^{\Imc}$
is a model of $\Omc'_{n}$ and we have again derived a contradiction to $\Omc'_{n}\models \Omc$.

The following example shows that transforming an \ALCI-ontology into an $\ELIU_{\bot}$-ontology in a straightforward way does not preserve $\ELI_{\bot}$-ontology strengthening OMQ evaluation.	
\begin{exmp}
	Consider the ontology
	\begin{eqnarray*}
		\Omc & = & \{ \top \sqsubseteq \exists r.\neg A \sqcup \exists s.\neg B,\\
		&   & \;\;\; \top \sqsubseteq \exists r.\top \sqcap \exists s.\top,\\
		&   & \;\;\; \exists r^{-}.E \sqsubseteq A, \exists s^{-}.F \sqsubseteq B\}
	\end{eqnarray*}
	and the database $\Dmc= \{E(a),F(b)\}$. A typical procedure to obtain an $\ELIU_\bot$-ontology $\Omc'$ from $\Omc$ introduces fresh concept names $A'$ and $B'$, adds the CIs $A' \sqsubseteq \neg A$,
	and $B' \sqsubseteq \neg B$ to $\Omc$, and then replaces $\neg A$ by $A'$ and $\neg B$ by $B'$ in $\Omc$.
	%
	%
	As $\Dmc$ is not satisfiable w.r.t.~the ontologies obtained from $\Omc'$ by replacing 
	$\top \sqsubseteq \exists r.A' \sqcup \exists s.B'$ by 
	$\top \sqsubseteq \exists r.A'$ or by $\top \sqsubseteq \exists s.B'$, 
	we obtain that $a \in \mn{app}_{\ELI_\bot}^\uparrow(Q,\Dmc)$ for any $Q(x)=(\Omc',\Sigma,q(x))$.
	
	This is not the case for $\Omc$ itself. For the ontology $\Omc^{\ast}$
	obtained from $\Omc$ by adding the CIs $A' \sqsubseteq \neg A$,
	and $B' \sqsubseteq \neg B$ and replacing
	$\top \sqsubseteq \exists r.\neg A \sqcup \exists s.\neg B$
	by $\exists r.(M \sqcap A) \sqsubseteq \exists s.B'$ and $\top \sqsubseteq \exists r.M$ we have that $\Omc^{\ast}\models \Omc$ but $\Dmc$ is satisfiable w.r.t.~$\Omc^{\ast}$.
\end{exmp}

\medskip

\subsection{Database strengthening approximation}

For classes of pointed databases \Dmf that are closed under substructures,
\Dmf-database strengthening approximations can be characterized
in a natural way.  For $\langle \Dmc, \bar a \rangle \in \Dmf$,
let $\mn{id}_\Dmf(\Dmc, \bar a)$ be
the set of ponted databases $\langle \Dmc', \bar a' \rangle$
that can be obtained from $\langle \Dmc, \bar a \rangle \in \Dmf$ by
identifying constants. Every such identification gives rise to an obvious
\emph{provenance homomorphism} $h$ from $\Dmc$ to~$\Dmc'$ with $h(\bar a) = \bar a'$.
\begin{restatable}{lemma}{lemidentifications}
\label{lem:identifications}
Let $\Dmf$ be a class of pointed databases that is closed under substructures,
$Q(\bar x)=(\Omc, \Sigma, q) \in (\text{FO},\text{UCQ})$ an OMQ,
$\Dmc$ a $\Sigma$-database and $\bar a \in \mn{adom}(\Dmc)^{|\bar x|}$.
Then $\bar a \in \mn{app}^\uparrow_\Dmf(Q,\Dmc)$ iff
for all $\langle \Dmc', \bar a' \rangle \in \mn{id}_\Dmf(\Dmc, \bar a)$,
$\bar a' \in Q(\Dmc')$.
\end{restatable}


\noindent
\begin{proof}\
Only the `if' needs to be proved.
Assume that for all $\langle \Dmc', \bar a' \rangle \in \mn{id}_\Dmf(\Dmc, \bar a)$,
$\bar a' \in Q(\Dmc')$. Let $\langle \Dmc', \bar b \rangle \in \Dmf$
and $h$ be a homomorphism from $\Dmc$ to $\Dmc'$ with $h(\bar a) = \bar b$.
Let $\Dmc'' \subseteq \Dmc'$
be the set $\{R(h(\bar c)) \mid R(\bar c) \in \Dmc\}$. Clearly,
$\langle \Dmc'', \bar b \rangle$ is isomorphic to a
database that can be obtained from $\langle \Dmc, \bar a \rangle$ by identifying constants,
i.e. isomorphic to a pointed database from
$\mn{id}(\Dmc, \bar a)$. Since $\Dmf$ is
closed under substructures, $\langle \Dmc'', \bar b \rangle \in \Dmf$.
By assumption, $h(\bar a) \in Q(\Dmc'')$.
Since $\Dmc'' \rightarrow \Dmc'$ and answers to queries from $(\text{FO}, \text{UCQ})$
are preserved under homomorphisms, $h(\bar a) \in Q(\Dmc')$.
Thus, $\bar a \in \mn{app}^\uparrow_\Dmf(Q,\Dmc)$.
\end{proof}

In the rest of this section, we only deal with Boolean (U)CQs, so just write
$\Dmc$ instead of $\langle \Dmc, () \rangle$ and $\mn{id}_\Dmf(\Dmc)$ instead
of $\mn{id}_\Dmf(\Dmc, ())$.
To prove Theorem~\ref{thm:dbfromabovesucks}, we show the
following.
\begin{lemma}
\label{lem:tw1overapprox}
There is a boolean UCQ $q$ such that the following problem is $\coNP$-hard: Given
a database $\Dmc$, is it true that $\Dmc' \models q$ for all
$\Dmc' \in \Dmf_1$ with $\Dmc \rightarrow \Dmc'$?
\end{lemma}

Before giving a definition of the UCQ $q$,
we describe the idea of the $\coNP$-hardness proof.
We reduce from the validity problem of propositional formulas in 3-DNF.
Given a 3-DNF formula $\varphi$, we construct a database $\Dmc_\varphi$
such that $\Dmc_\varphi \models q$ if and only if $\varphi$ is valid.
$\Dmc_\varphi$ consists of several variable gadgets $\Dmc_x$,
one for every variable $x$ in $\varphi$.
The variable gadget $\Dmc_x$ is a database that
does not have treewidth 1, but there are precisely two intended ways
to identify constants of $\Dmc_x$ to obtain a database of
treewith 1, where one way corresponds to assigning \emph{true} to $x$,
the other way to assigning \emph{false}. There will be more than
the two intended ways to identify constants of $\Dmc_x$, but
these unintended ways are detected using a UCQ $q_\mn{unintended}$,
which is entailed in all databases that are obtained by an unintended way of identifying
constants. However, if all variable gadgets are identified in an intended
way, this corresponds to an assignment of the variables of $\varphi$.
We then use a second UCQ $q_\mn{valid}$ to check whether there
is a conjunct where all literals evaluate to \emph{true}.
The UCQ $q$ is then defined as $q = q_\mn{unintended} \vee q_\mn{valid}$.

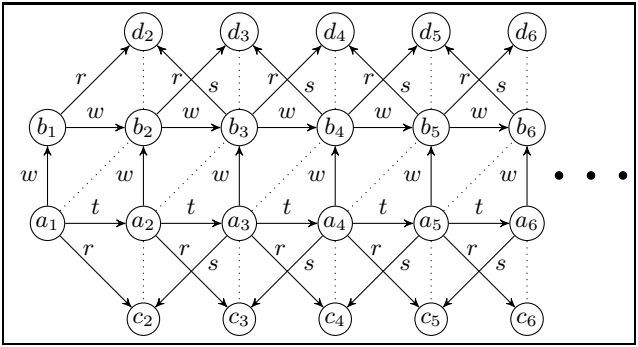
\begin{figure}
\begin{boxedminipage}{\columnwidth}
      \centering
\begin{tikzpicture}[->,>=stealth',level distance = 0.8cm,font=\sffamily\small,scale=0.85]
\tikzstyle{node}=[shape=circle, draw,inner sep=1.0pt]
\node [node] (A1) at (0,0) {$a_1$};
\node [node] (A2) at (1.5,0) {$a_2$};
\node [node] (A3) at (3,0) {$a_3$};
\node [node] (A4) at (4.5,0) {$a_4$};
\node [node] (A5) at (6,0) {$a_5$};
\node [node] (A6) at (7.5,0) {$a_6$};

\node [node] (B1) at (0,1.5) {$b_1$};
\node [node] (B2) at (1.5,1.5) {$b_2$};
\node [node] (B3) at (3,1.5) {$b_3$};
\node [node] (B4) at (4.5,1.5) {$b_4$};
\node [node] (B5) at (6,1.5) {$b_5$};
\node [node] (B6) at (7.5,1.5) {$b_6$};

\node [node] (C2) at (1.5,-1.5) {$c_2$};
\node [node] (C3) at (3,-1.5) {$c_3$};
\node [node] (C4) at (4.5,-1.5) {$c_4$};
\node [node] (C5) at (6,-1.5) {$c_5$};
\node [node] (C6) at (7.5,-1.5) {$c_6$};

\node [node] (D2) at (1.5,3) {$d_2$};
\node [node] (D3) at (3,3) {$d_3$};
\node [node] (D4) at (4.5,3) {$d_4$};
\node [node] (D5) at (6,3) {$d_5$};
\node [node] (D6) at (7.5,3) {$d_6$};

\draw [->] (A1) -- (B1) node[midway, left] {$w$};
\draw [->] (A2) -- (B2) node[midway, left] {$w$};
\draw [->] (A3) -- (B3) node[midway, left] {$w$};
\draw [->] (A4) -- (B4) node[midway, left] {$w$};
\draw [->] (A5) -- (B5) node[midway, left] {$w$};
\draw [->] (A6) -- (B6) node[midway, left] {$w$};

\draw [->] (A1) -- (A2) node[midway, above] {$t$};
\draw [->] (A2) -- (A3) node[midway, above] {$t$};
\draw [->] (A3) -- (A4) node[midway, above] {$t$};
\draw [->] (A4) -- (A5) node[midway, above] {$t$};
\draw [->] (A5) -- (A6) node[midway, above] {$t$};

\draw [->] (B1) -- (B2) node[midway, above] {$w$};
\draw [->] (B2) -- (B3) node[midway, above] {$w$};
\draw [->] (B3) -- (B4) node[midway, above] {$w$};
\draw [->] (B4) -- (B5) node[midway, above] {$w$};
\draw [->] (B5) -- (B6) node[midway, above] {$w$};

\draw [->] (A1) -- (C2) node[pos=0.4, above] {$r$};
\draw [->] (A2) -- (C3) node[pos=0.4, above] {$r$};
\draw [->] (A3) -- (C4) node[pos=0.4, above] {$r$};
\draw [->] (A4) -- (C5) node[pos=0.4, above] {$r$};
\draw [->] (A5) -- (C6) node[pos=0.4, above] {$r$};

\draw [->] (A3) -- (C2) node[pos=0.4, right] {$s$};
\draw [->] (A4) -- (C3) node[pos=0.4, right] {$s$};
\draw [->] (A5) -- (C4) node[pos=0.4, right] {$s$};
\draw [->] (A6) -- (C5) node[pos=0.4, right] {$s$};

\draw [->] (B1) -- (D2) node[pos=0.3, above] {$r$};
\draw [->] (B2) -- (D3) node[pos=0.3, above] {$r$};
\draw [->] (B3) -- (D4) node[pos=0.3, above] {$r$};
\draw [->] (B4) -- (D5) node[pos=0.3, above] {$r$};
\draw [->] (B5) -- (D6) node[pos=0.3, above] {$r$};

\draw [->] (B3) -- (D2) node[pos=0.4, right] {$s$};
\draw [->] (B4) -- (D3) node[pos=0.4, right] {$s$};
\draw [->] (B5) -- (D4) node[pos=0.4, right] {$s$};
\draw [->] (B6) -- (D5) node[pos=0.4, right] {$s$};

\draw [-,dotted] (A1) -- (B2) node[] {};
\draw [-,dotted] (A2) -- (B3) node[] {};
\draw [-,dotted] (A3) -- (B4) node[] {};
\draw [-,dotted] (A4) -- (B5) node[] {};
\draw [-,dotted] (A5) -- (B6) node[] {};

\draw [-,dotted] (A2) -- (C2) node[] {};
\draw [-,dotted] (A3) -- (C3) node[] {};
\draw [-,dotted] (A4) -- (C4) node[] {};
\draw [-,dotted] (A5) -- (C5) node[] {};
\draw [-,dotted] (A6) -- (C6) node[] {};

\draw [-,dotted] (B2) -- (D2) node[] {};
\draw [-,dotted] (B3) -- (D3) node[] {};
\draw [-,dotted] (B4) -- (D4) node[] {};
\draw [-,dotted] (B5) -- (D5) node[] {};
\draw [-,dotted] (B6) -- (D6) node[] {};

\node [node,fill=black] at (8,0.75) {}; 
\node [node,fill=black] at (8.5,0.75) {};
\node [node,fill=black] at (9,0.75) {};
\end{tikzpicture}
  \end{boxedminipage}
\caption{The figure shows a part of a variable gadget $\Dmc_x$. The dotted lines correspond to
the first of the two intended identifications, which stands for assigning \emph{true} to $x$.}
\label{fig:variable-gadget}
\vspace*{-3mm}
\end{figure}

Let $\varphi = (\ell_1^1 \wedge \ell_1^2 \wedge \ell_1^3)
\vee \ldots \vee (\ell_n^1 \wedge \ell_n^2 \wedge \ell_n^3)$
be a propositional formula in 3-DNF, where every $\ell_i^j$ is 
either a variable or a negated variable.
We start by constructing the variable gadgets.
Let $x$ be a variable of $\varphi$. The database $\Dmc_x$ takes the form
of a grid of height 1 and width $N:=9 \cdot (2n-1)$,
plus some additional constants and facts that are used for technical reasons.
We use constants of the form $a^x_i$ and $b^x_i$, where $1 \leq i \leq N$, and of the form $c^x_i$ and $d^x_i$ where $1 < i < N$.
\begin{align*}
\Dmc_x = &\{t(a^x_i, a^x_{i+1}) \mid 1 \leq i < N\} \cup \\
&\{w(b^x_i, b^x_{i+1}) \mid 1 \leq i < N\} \cup \\
&\{w(a^x_i, b^x_{i}) \mid 1 \leq i \leq N\} \cup \\
&\{r(a^x_i, c^x_{i+1}) \mid 1 \leq i \leq N-2\} \cup \\
&\{s(a^x_{i+1}, c^x_{i}) \mid 2 < i \leq N-1\} \cup \\
&\{r(b^x_i, d^x_{i+1}) \mid 1 \leq i \leq N-2\} \cup \\
&\{s(b^x_{i+1}, d^x_{i}) \mid 2 < i \leq N-1\}
\end{align*}
Figure~\ref{fig:variable-gadget} shows $\Dmc_x$.
So far, all $\Dmc_x$ are isomorphic. They will later be distinguished by adding certain unary
facts.
The two \emph{intended identifications} in $\Dmc_x$ are the following:
\begin{enumerate}
\item Identify $b_i^x$ with $a_{i-1}^x$ for every $i$ with $1 < i \leq N$.
Furthermore, identify $a_i^x$ with $c_i^x$ and identify $b_i^x$ with $d_i^x$
for every $i$ with $1 < i < N$.
This identification corresponds to assigning \emph{true} to $x$.
\item Identify $b_i^x$ with $a_{i+1}^x$ for every $i$ with $1 \leq i < N$.
Furthermore, identify $a_i^x$ with $c_i^x$ and identify $b_i^x$ with $d_i^x$
for every $i$ with $1 < i < N$.
This identification corresponds to assigning \emph{false} to $x$.
\end{enumerate}
We claim that the following UCQ detects precisely the unintended identifications:
\[q_\mn{unintended} = (\exists x \exists y r(x,y) \wedge s(x,y)) \vee \exists x w(x,x) \vee \exists x t(x,x)\]

\begin{lemma}
\label{lem:vargadget}
Let $\Dmc_x$ be a variable gadget and $\Dmc' \in \mn{id}_{\Dmf_1}(\Dmc_x)$. Then $\Dmc' \models q_\mn{unintended}$ if and only if the identification that was used to obtain $\Dmc'$ from $\Dmc_x$ is none of the two intended identifications.
\end{lemma}

\noindent
\begin{proof}\
One direction is easy to check: If $\Dmc'$ is obtained by one of the two intended
identifications of constants, then $\Dmc' \not \models q_\mn{unintended}$.
For the other direction, let $\Dmc' \not \models q_\mn{unintended}$.
Since $\Dmc'$ has treewidth 1, it does not contains any cycle.
For every $i$ with $1 \leq i < N$, there is the cycle
$w(a_i^x,b_i^x), w(b_i^x,b_{i+1}^x), w^-(b_{i+1}^x, a_{i+1}^x), t^-(a_{i+1}^x, a_i^x) \in \Dmc_x$.
In $\Dmc'$, some of the four affected constants are identified.
Since $\Dmc' \not \models q_\mn{unintended}$, no two constants
that were adjacent in $\Dmc_x$ are identified, because this would have created
a self-loop with the role $w$ or $t$. The only two remaining possibilities are that
(1) $a_i^x$ was identified with $b_{i+1}^x$ or that
(2) $a_{i+1}^x$ was identified with $b_i^x$.
If for some $i$, $a_i^x$ was identified with $b_{i+1}^x$,
then $b_{i+1}^x$ cannot be identified with $a_{i+2}^x$,
since this would identify $a_i^x$ with $a_{i+2}^x$,
allowing a match for the first disjunct of $q_\mn{unintended}$,
mapping $x$ to $a_i^x$ and $y$ to $c_{i+1}^x$.
Thus, if $a_i^x$ was identified with $b_{i+1}^x$ for some $i$,
then $a_{i+1}^x$ was identified with $b_{i+2}^x$.
Similarly, if $b_i^x$ was identified with $a_{i+1}^x$ for some $i$,
then $b_{i+1}^x$ was identified with $a_{i+2}^x$.
By induction, this shows that either $a_i^x$ is identified with $b_{i+1}^x$
for all $i$ with $1 \leq i < N$ or $b_i^x$ is identified with $a_{i+1}^x$
for all $i$ with $1 \leq i < N$, just as in the two intended identifications.

It remains to show that for every $i$ with $1<i<N$, $a_i^x$ is identified with $c_i^x$
and $b_i^x$ with $d_i^x$.
Consider the cycle
$t(a_{i-1}^x, a_{i}^x), t(a_i^x, a_{i+1}^x), s(a_{i+1}^x, c_{i}^x), r^-(c_{i}^x, a_{i-1}^x) \in \Dmc_x$,
that was collapsed in $\Dmc'$. If $a_{i-1}^x$ was identified with $a_i^x$ or
$a_i^x$ with $a_{i+1}^x$, this would have created a match for $\exists x t(x,x)$.
The only remaining options are identifying $a_{i-1}^x$ with $a_{i+1}^x$
or $a_i^x$ with $c_i^x$. The first option would create a match for the first disjunct of
$q_\mn{unintended}$, so it follows that $a_i^x$ was identified with $c_i^x$.
A similar argument show that $b_i^x$ is identified with $d_i^x$, considering
the cycle $w(b_i^x, b_{i+1}^x), w(b_{i+1}^x, b_{i+2}^x),
s(b_{i+2}^x, d_{i+1}^x), r^-(d_{i+1}^x, b_i^x) \in \Dmc_x$.
Thus, the identification of constants used to obtain $\Dmc'$ from $\Dmc_x$
is one of the two intended identifications.
\end{proof}

Next, we construct $\Dmc_\varphi$. We start by taking the disjoint union of all $\Dmc_x$,
for all variables $x$ that appear in $\varphi$. Now we glue these gadgets together along
their $t$-paths, i.e.~for every $i$ with $1 \leq i \leq N$ we do the following identification:
identify all constants of the form
$a_i^x$ for some $x \in \mn{var}(\varphi)$ to a single constant, which we just call $a_i$.
Finally, we add some unary facts to encode $\varphi$ in $\Dmc$.
For all $x \in \mn{var}(\varphi)$ and all $i,j$ such that $\ell_i^j \in \{x, \neg x\}$,
\begin{itemize}
\item add $T(b^x_{k})$, where $k=18 (i-1) + 3j - 1$;
\item if  $\ell_i^j = x$, add $A(a_k)$, where $k=18 (i-1) + 3j - 2$;
\item if  $\ell_i^j = \neg x$, add $A(a_k)$, where $k=18 (i-1) + 3j$.
\end{itemize}
It is clear that $\Dmc_\varphi$ can be constructed from $\varphi$ in polynomial time.
Note the effect that the two intended identifications have on $\Dmc$: The identification
that corresponds to an assignment that makes a literal $\ell_i^j$ become true creates
a constant that satisfies both $A$ and $T$.

Now we finish the definition of $q$ by defining $q_\mn{valid}$.
Our aim is to build $q_\mn{valid}$ in such a way that it detects whether the
chosen assignment makes $\varphi$ true, which is the case if and only if there
is an $i$ such that all $\ell_i^1$, $\ell_i^2$, and $\ell_i^3$ evaluate to true under
the assignment. This can be checked using a UCQ asking for certain $t$-paths
that contain $3$ different nodes that satisfy both $A$ and $T$.
For $S \in \{1, \ldots, 9\}^3$,
let $q_S = \exists x_1 \ldots x_9 \bigwedge_{i = 1}^8 t(x_i,x_{i+1})
\wedge \bigwedge_{i \in S} A(x_i) \wedge T(x_i)$, so $q_S$ asks for a $t$-path
of length $8$, where all positions that appear in $S$ are labelled with both $A$ and $T$.
Define \[q_\mn{valid} = \bigvee_{S \in \{1,3\} \times \{4, 6\} \times \{7, 9\}} q_S\]
and let $q = q_\mn{unintended} \vee q_\mn{valid}$.
To finish the proof of Lemma~\ref{lem:tw1overapprox}, we need to argue the following:

\begin{lemma}
$\varphi$ is valid if and only if for all $\Dmc' \in \Dmf_1$ with $\Dmc_\varphi \rightarrow \Dmc'$, $\Dmc' \models q$.
\end{lemma}

\noindent
\begin{proof}\
Let $\varphi$ be valid. By Lemma~\ref{lem:identifications}, it is sufficient to show that
$\Dmc' \models q$ for all $\Dmc' \in \mn{id}_{\Dmf_1}(\Dmc_\varphi)$.
Let $\Dmc' \in \mn{id}_{\Dmf_1}(\Dmc_\varphi)$ and assume, for the sake of contradiction,
that $\Dmc' \not \models q$.
For every $x \in \mn{var}(\varphi)$, $\Dmc_x$ is a substructure of $\Dmc_\varphi$, so
by Lemma~\ref{lem:vargadget}, $\Dmc'$ was obtained from $\Dmc_\varphi$
by using one of the two intended identifications for every variable gadget $\Dmc_x$.
Recall that each of the intended identifications of $\Dmc_x$ corresponds to a truth assignment
of $x$, so let $V: \mn{var}(\varphi) \rightarrow \{0, 1\}$ be the corresponding assignment.
Since $\varphi$ is valid, there is one conjunct that evaluates to \emph{true} under the
assignment $V$, so there is an $i \in \{1, \ldots, n\}$ such that all three literals $\ell_i^1$,
$\ell_i^2$ and $\ell_i^3$ evaluate to true. Then $q_\mn{valid}$ has a match in $\Dmc'$
via the homomorphism $h(x_j) = a_{18(i-1)+j}$, a contradiction.

For the other direction, let $\varphi$ be invalid, so let $V: \mn{var}(\varphi) \rightarrow \{0, 1\}$
be an assignment that makes $\varphi$ false. Let $\Dmc'$ be obtained from $\Dmc_\varphi$
by choosing for every substructure $\Dmc_x$ the identification corresponding to $V(x)$.
It is easy to check that $\Dmc' \in \Dmf_1$, in fact, $\Dmc'$ is a path with multi-edges.
Since $\Dmc'$ was obtained by only using intended identifications and by
Lemma~\ref{lem:vargadget},
$\Dmc' \not \models q_\mn{unintended}$.
For the sake of contradiction, assume that $\Dmc' \models q_\mn{valid}$, so there
is a $(s_1, s_2, s_3) \in \{1, 3\} \times \{4, 6\} \times \{7, 9\}$
such that $\Dmc' \models q_{(s_1, s_2, s_3)}$ via a
homomorphism $h$.
Since $\Dmc'$ contains only a single long $t$-path $t(a_1,a_2),\ldots,t(a_{N-1},a_N)$,
the homomorphism $h$ is of the form $h(x_i) = a_{k+1}$ for some $k \in \{1, \ldots, N\}$.
In particular, $A(a_{s_1}), A(a_{s_2}), A(a_{s_3}) \subseteq \Dmc'$. However,
$\Dmc_\varphi$ is constructed in such a way that whenever there are three occurrences
$A(a_{k+s_1}), A(a_{k+s_2}), A(a_{k+s_3})$ within a radius of at most 8, then these were introduced
for the three different literals $\ell_i^1$, $\ell_i^2$ and $\ell_i^3$ of the $i$th conjunct for 
some $i \in \{1, \ldots, n\}$. Furthermore, the existence of $h$ also yields the three facts
$T(a_{k+s_1}), T(a_{k+s_2}), T(a_{k+s_3}) \in \Dmc'$. Since $\Dmc'$ was obtained
from $\Dmc_\varphi$ using only intended identifications, and by construction of
the $\Dmc_x$, the fact $T(a_{s_j})$ can only be created by identifying $a_{k+s_j}$ with
$b_{k+s_j+1}^x$ or with $b_{k+s_j-1}^x$, where $x$ is the variable of $\ell_i^j$. But these
identifications correspond to an assignment, that makes all three literals $\ell_i^1$,
$\ell_i^2$ and $\ell_i^3$ evaluate to true, contradicting the assumption that $V$ makes
$\varphi$ false.
\end{proof}

We describe how the construction can be adapted slightly to show that
$\Dmf_1$-database strengthening approximation in
$(\EL,\text{CQ})$ is $\coNP$-hard as well.
Let $q_0 = \exists x \exists y r(x,y) \wedge s(x,y) \wedge M(y)$.
The idea is to design an ontology $\Omc$ and database $\Dmc$ in such a way that for every
$\Dmc' \in \mn{id}_{\Dmf_1}(\Dmc)$ we have $\Dmc' \models q$ if and only if $\Omc, \Dmc' \models q_0$.

In the construction of $\Dmc_x$, we add the facts $M(c_i^x)$ and $M(d_i^x)$ for all $i$
with $1 < i < N$. Furthermore, we introduce constants $e_i$ for $1 \leq i \leq N$ and add
the facts $r(a_i,e_i), s(a_i, e_i)$ and $v(e_i,a_i)$.
We also introduce new unary symbols $A_1, A_2, A_3$ and $B_1, B_2, B_3$ and
add $A_{i~\mn{mod}~3}(a_i^x)$ and $B_{i~\mn{mod}~3}(b_i)$ for all $x \in \mn{var}(\varphi)$
and $i$ with $1 \leq i \leq N$.

The ontology is used to translate every match of $q$ into a match of $q_0$.
Recall that
\[q_\mn{unintended} = (\exists x \exists y r(x,y) \wedge s(x,y)) \vee \exists x w(x,x) \vee \exists x t(x,x) \,.\]
For the first disjunct of $q_\mn{unintended}$, there is nothing to do, since we already added the facts
$M(c_i^x)$ and $M(d_i^x)$ to $\Dmc$. For the second and third disjunct, we introduce the concept
inclusions $\exists v . C \sqsubseteq M$ for every $C \in \{A_1 \sqcap A_2, A_1 \sqcap A_3, A_2 \sqcap A_3, B_1 \sqcap B_2, B_1 \sqcap B_3, B_2 \sqcap B_3, A_1 \sqcap B_1, A_2 \sqcap B_2, A_3 \sqcap B_3\}$. Every disjunct $q_S$ of $q_\mn{valid}$ can be seen as an $\EL$-concept $C_S$. We
introduce a concept inclusion $\exists v. C_S \sqsubseteq M$ for each such $C_S$. It can then be
verified that $\Omc, \Dmc \models q_0$ if and only if $\varphi$ is valid.

It is an interesting question whether the lower bound can already be established
for $(\EL, \text{AQ})$, or for the empty ontology with CQs or AQs.
We leave this as an open problem.

\end{document}